\documentclass[11pt]{llncs}
\usepackage{helvet}

\usepackage[top=2.5cm, bottom=2.5cm, left=2.5cm, right=2.5cm]{geometry} 
\usepackage{multirow}
\usepackage{amsmath,graphicx}
\usepackage{bm}
\usepackage[colorlinks=true,linkcolor=blue,citecolor=blue]{hyperref}
\usepackage{caption}
\usepackage{array} 
\usepackage{url} 
\usepackage{tikz}
\usepackage[noadjust]{cite}
\usetikzlibrary{calc, arrows, positioning, shapes, shadows, spy, snakes, plotmarks, matrix, fit, backgrounds}

\title{Identifying the Best Machine Learning Algorithms for Brain Tumor Segmentation, Progression Assessment, and Overall Survival Prediction in the BRATS Challenge}
\author{Spyridon Bakas \inst{1,2,3,\dag,\ddag,*} \and
Mauricio Reyes \inst{4,\dag} \and
Andras Jakab \inst{5,\dag,\ddag} \and
Stefan Bauer \inst{4,6,169,\dag} \and
Markus Rempfler \inst{9,65,127,\dag} \and
Alessandro Crimi \inst{7,\dag} \and
Russell Takeshi Shinohara \inst{1,8,\dag} \and
Christoph Berger \inst{9,\dag} \and
Sung Min Ha \inst{1,2,\dag} \and
Martin Rozycki \inst{1,2,\dag}
% ...
}

\institute{Center for Biomedical Image Computing and Analytics, University of Pennsylvania, Philadelphia, PA, USA \and
            Department of Radiology, Perelman School of Medicine, University of Pennsylvania, Philadelphia, PA, USA \and
            Department of Pathology and Laboratory Medicine, Perelman School of Medicine, University of Pennsylvania, Philadelphia, PA, USA \and
            Institute for Surgical Technology and Biomechanics, University of Bern, Bern, Switzerland \and
            Center for MR-Research, University Children's Hospital Zurich, Zurich, Switzerland \and
            Support Centre for Advanced Neuroimaging Inselspital, Institute for Diagnostic and Interventional Neuroradiology, Bern University Hospital, Bern, Switzerland \and
            University Hospital of Zurich, Zurich, Switzerland \and
            Center for Clinical Epidemiology and Biostatistics, University of Pennsylvania, Philadelphia, USA \and
            Image-Based Biomedical Modeling Group, Technical University of Munich, Munich, Germany \and
            Icahn School of Medicine, Mount Sinai Health System, New York, NY, USA \and
            Leidos Biomedical Research, Inc., Frederick National Laboratory for Cancer Research, Frederick, MD 21701, USA \and
            Cancer Imaging Program, National Cancer Institute, National Institutes of Health, Bethesda, MD 20814, USA \and
            Department of Neurology, The University of Alabama at Birmingham, Birmingham, AL, USA \and
            Department of Diagnostic Radiology, University of Texas MD Anderson Cancer Center, Houston, TX, USA \and
            Department of Psychology, Washington University, St. Louis, MO, USA \and
            Neuroimaging Informatics and Analysis Center, Washington University, St. Louis, MO, USA \and
            Department of Radiology, Washington University, St. Louis, MO, USA \and
            Institute of Diagnostic and Interventional Radiology, Pediatric Radiology and Neuroradiology, University Medical Center Rostock, Ernst-Heydemann-Str. 6, 18057 Rostock, Germany \and
            Tata Memorial Centre, Homi Bhabha National Institute, Mumbai, India \and
            Shri Guru Gobind Singhji Institute of Engineering and Technology, Nanded, India \and
            NVIDIA, Santa Clara, USA \and
            Division of Medical Image Computing, German Cancer Research Center (DKFZ), Heidelberg, Germany \and
            Department of Neuroradiology, University of Heidelberg Medical Center, Heidelberg, Germany \and
            Neurology Clinic, University of Heidelberg Medical Center, Heidelberg, Germany \and
            School of Electronic and Information Engineering, South China University of Technology, Guangzhou, China \and
            UBTECH Sydney AI Centre, SIT, FEIT, University of Sydney, Sydney, Australia \and
            Biomedical Engineering, University of Virginia, Charlottesville, VA 22903, USA \and
            Radiology and Medical Imaging, University of Virginia, Charlottesville, VA 22903, USA \and
            EPITA Research and Development Laboratory (LRDE), France \and
            Southern University of Science and Technology, Shenzhen, China \and
            Peking University, China \and
            Institute of Imaging and Computer Vision, RWTH Aachen University, Aachen, Germany \and
            Vision Lab, Electrical and Computer Engineering, Old Dominion University, Norfolk, VA, USA \and
            iTeam, Universitat Politecnica Valencia, Spain \and
            Instituto de Fisica Corpuscular, Universitat de Valencia, Consejo Superior de Investigaciones Cientificas, Spain \and
            Machine Intelligence Unit, Indian Statistical Institute, Kolkata, 700108, India \and
            Department of CSE, University of Calcutta, Kolkata, India \and
            Computer Vision Lab, School of Computer Science, University of Nottingham, Nottingham, UK \and
            School of Biosciences, University of Nottingham, Nottingham, UK \and
            Research Institute of Computer Vision and Robotics, University of Girona, Spain \and
            Henry Ford Health System, Detroit, MI, USA \and
            CVN, CentraleSuplec, Universit Paris-Saclay, France \and
            Gustave Roussy Institute, France \and
            TheraPanacea, France \and
            Department of Electrical Engineering, National Taiwan University of Science and Technology, Taipei, Taiwan \and
            Dept. of Biomedical Engineering, School of Control Science and Engineering, Shandong University, China \and
            Department of Computer Science, Florida State University, Tallahassee, FL, USA \and
            Department of Radiology, Columbia University Medical Center, New York, NY, USA \and
            Data Science Institute, Columbia University, New York, NY, USA \and
            Faculty of Business and Economics, The University of Hong Kong, Hong Kong \and
            Department of Biostatistics, University of North Carolina, Chapel Hill, NC, USA \and
            Department of Biostatistics, University of Texas, MD Anderson Cancer Center, Houston, TX, USA \and
            Southern Medical University, Guangzhou, China \and
            Research Center for Brain-inspired Intelligence, Institute of Automation, Chinese Academy of Sciences, China \and
            University of Chinese Academy of Sciences, China \and
            Center for Excellence in Brain Science and Intelligence Technology, Chinese Academy of Sciences, China \and
            Afeka Academic College of Engineering, Tel Aviv, Israel \and
            Department of Imaging Physics, University of Texas MD Anderson Cancer Center, Houston, TX, USA \and
            Department of Cancer Systems Imaging and Diagnostic Radiology, University of Texas MD Anderson Cancer Center, Houston, TX, USA \and
            HealthMyne, Madison, WI, USA \and
            University of California, Berkeley, CA, USA \and
            Institute for Computational Engineering and Science, University of Texas, Austin, TX, USA \and
            Odiga London, London, UK \and
            Malong Tehnologies, China \and
            Department of Computer Science, Technical University of Munich, Munich, Germany \and
            School of Computer Science and Engineering, Northwestern Polytechnical University, University of Xi'an, Xi'an 710072, China \and
            The First Affiliated Hospital of Xi'an Jiao Tong University, People's Republic of China, China \and
            Shanghai United Imaging Intelligence Co., Ltd, Shanghai, China \and
            School of Biomedical Engineering, Southeast University, Nanjing, China \and
            Harvard University, Cambidge, MA, USA \and
            Future Processing, Poland \and
            Silesian University of Technology, Gliwice, Poland \and
            Department of Mathematics, Nanjing University of Science and Technology, China \and
            LMCS Laboratory, National Higher School of Computer Sciences (ESI), Algeria \and
            LabGed Laboratory, Dept. of Computer Sciences, University Badji-Mokhtar of Annaba, Algeria \and
            School of Biomedical Engineering and Imaging Sciences, King's College London, London, UK \and
            Wellcome / EPSRC Centre for Interventional and Surgical Sciences, University College London, London, UK \and
            Stryker Corporation, Navigation. Freiburg im Breisgau, Germany. \and
            Vision Research Lab, University of California, Santa Barbara, CA, USA \and
            UC Irvine Health, University of California, Irvine, CA, USA \and
            Graduate School for Integrative Sciences and Engineering, National University of Singapore, Singapore \and
            Department of Biomedical Engineering, National University of Singapore, Singapore \and
            Dept. of Instrumentation and Control Engineering, National Institute of Technology, Tiruchirappalli, India \and
            Intel, Nizhny Novgorod, Russian Federation \and
            Lobachevsky State University, Russian Federation \and
            Technological Institute of Irapuato, Information Technologies Laboratory, Mexico \and
            College of Engineering, Mathematics and Physical Sciences, University of Exeter, UK \and
            Department of Biomedical Informatics, University of Pittsburgh, PA, USA \and
            Philosophy Department, Carnegie Mellon University, Pittsburgh, PA, USA \and
            UBTECH Sydney AI Centre, SIT, FEIT, The University of Sydney, Australia \and
            University of Washington, Seattle, WA, USA \and
            Centre for Intelligent Machines (CIM), McGill University, Montreal, QC, Canada \and
            Yale University, New Haven, CT, USA \and
            Cura Cloud Cooperation, Seattle, WA, USA \and
            Hasso-Plattner Institute for Digital Engineering, Prof.-Dr.-Helmert-Strae 2-3, 14482 Potsdam, Germany \and
            Department of Computer Science, University of Medicine, Pharmacy, Sciences and Technology, Romania \and
            Department of Electrical Engineering, Sapientia Hungarian University of Transylvania, Romania \and
            School of Information Science and Engineering, Yunnan University, P.R.China \and
            QMENTA, Boston, MA, USA \and
            Computer Vision Center, Universitat Autonoma de Barcelona, Spain \and
            SIMBIOsys, Universitat Pompeu Fabra, Spain \and
            University Clinic for Radio-oncology, Bern University Hospital, Switzerland \and
            Faculty of Math and Computer Science, University of Science, Vietnam National University, Vietnam \and
            Department of Computer Science, Sai Gon University, Vietnam \and
            INESC-ID, Portugal \and
            Instituto Superior Tecnico, Portugal \and
            Indian Institute of Technology Madras, Chennai, India \and
            University of Michigan, Ann Arbor, MI, USA \and
            Indian Institute of Technology Roorkee, India \and
            Tianjin University, China \and
            National Taiwan University, Taiwan \and
            School of Biomedical Engineering, Shanghai Jiao Tong University, China \and
            School of Biomedical Engineering, Shanghai Jiao Tong University, China \and
            Department of biological science and medical engineering, Beihang University, China \and
            School of Information Science and Engineering, Shandong University, China \and
            Institute of Brain and Brain-Inspired Science, Shandong University, China \and
            Tencent Youtu Lab, China \and
            QED Technologies. Co., Ltd, China \and
            Beihang University, China \and
            Beijing University of Posts and Telecommunications University, China \and
            Deepnoid, South Korea \and
            Department of Neurosurgery, Mount Sinai Health System, New York, NY, USA \and
            Athinoula A Martinos Center for Biomedical Imaging, Massachusetts General Hospital, Boston, MA, USA \and
            Department of Radiology, Harvard Medical School, Harvard University, Boston, MA, USA \and
            Department of Applied Mathematics and Computer Science, Technical University of Denmark, Denmark \and
            Department of Neuroradiology, Technical University of Munich, Munich, Germany \and
            Institute for Biomedical Engineering, Technical University of Munich, Munich, Germany \and
            Biomedical Image Analysis Group, Imperial College London, London, UK \and
            Case Western Reserve University, Cleveland OH 44106, USA \and
            Monash Biomedical Imaging, Monash University, Melbourne, Australia \and
            Radiation Oncology Branch, National Cancer Institute, National Institutes of Health, Bethesda, MD 20814, USA \and
            Department of Computer Science, University of Colorado, Colorado Springs, CO, USA \and
            College of Computer Science and Software Engineering, Computer Vision Institute, Shenzhen University, Shenzhen, China \and
            Department of Biomedical Engineering, Universidad de los Andes, Bogota, Colombia \and
            College of Computer and Information Sciences, Regis University, Denver, CO, USA \and
            Indian Institute of Information Technology, Vadodara, Gandhinagar Campus, Vadodara 382028, Gujarat, India \and
            Kharkevich Institute for Information Transmission Problems, Moscow, Russia \and
            Signal Theory and Communications Department, Universitat Politecnica de Catalunya - BarcelonaTech, Barcelona, Spain \and
            Independent research, No Institutional Affiliation \and
            Radiation Oncology Department, American University of Beirut Medical Center, Beirut 1107 2020, Riad El-Solh, Lebanon \and
            Department of Electrical Engineering (ESAT), STADIUS Center for Dynamical Systems, Signal Processing and Data Analytics, KU Leuven, Leuven, Belgium \and
            Mathematical Oncology Laboratory, Universidad de Castilla-La Mancha, Ciudad Real, Spain \and
            University of Lincoln, Lincoln, LN6 7TS, UK \and
            National Laboratory of Pattern Recognition, Institute of Automation, Chinese Academy of Sciences, Beijing, China \and
            Beijing Neurosurgical Institute, Capital Medical University, Beijing, China \and
            Department of Neurosurgery, Beijing Tiantan Hospital, Capital Medical University, Beijing, China \and
            Beijing Institute for Brain Disorders Brain Tumor Center, Beijing, China \and
            China National Clinical Research Center for Neurological Diseases, Beijing, China \and
            Universidad Tecnologica Nacional, Buenos Aires, Argentina \and
            Comision Nacional de Energia Atomica, Buenos Aires, Argentina \and
            GAMMA3 (UTT-INRIA), ICD-UMR CNRS 6281, University of Technology of Troyes, Troyes, France \and
            Department of Biomedical Engineering, University of Basel, Switzerland \and
            Computer Laboratory, University of Cambridge, Cambridge, CB3 0FD, UK \and
            Department of Radiology, University of Cambridge, Cambridge, CB2 0QQ, UK \and
            Computer Science and Engineer College, Chongqing University of Technology, China, 40054 \and
            IBM T. J. Watson Research, Yorktown Height, NY 10598, USA \and
            Computing, School of Science and Engineering, University of Dundee, Dundee, UK \and
            Department of Electronic Engineering, Fudan University, Shanghai, China \and
            Hospital Israelita Albert Einstein, São Paulo, Brazil \and
            Massachusetts General Hospital, Boston, MA, USA \and
            NVIDIA, India \and
            National University of Defense Technology, Changsha 410073, China \and
            Siemens, USA \and
            State University of New York at Buffalo, Buffalo, NY, USA \and
            Department of Mathematics, College of Natural Sciences and Mathematics, University of Houston, Houston, TX, USA \and
            Department of Computer Science, University of Stuttgart \and
            School for Technology and Health (STH), KTH Royal institute of technology, SE-14152 Huddinge, Stockholm, Sweden \and
            General Electric, USA \and
            Helbling Technik AG, Bern, Switzerland
% ...
}

\begin{document}
	\maketitle
    \centerline{\dag: People involved in the organization of the challenge}

    \centerline{\ddag: People contributing data from their institutions}

    \centerline{* Corresponding authors: s.bakas@uphs.upenn.edu, bjoern.menze@tum.de}
	
    \begin{abstract}
        Gliomas are the most common primary brain malignancies, with different degrees of aggressiveness, variable prognosis and various heterogeneous histologic sub-regions, i.e., peritumoral edematous/invaded tissue, necrotic core, active and non-enhancing core. This intrinsic heterogeneity is also portrayed in their radio-phenotype, as their sub-regions are depicted by varying intensity profiles disseminated across multi-parametric magnetic resonance imaging (mpMRI) scans, reflecting varying biological properties. Their heterogeneous shape, extent, and location are some of the factors that make these tumors difficult to resect, and in some cases inoperable. The amount of resected tumor is a factor also considered in longitudinal scans, when evaluating the apparent tumor for potential diagnosis of progression. Furthermore, there is mounting evidence that accurate segmentation of the various tumor sub-regions can offer the basis for quantitative image analysis towards prediction of patient overall survival. This study assesses the state-of-the-art machine learning (ML) methods used for brain tumor image analysis in mpMRI scans, during the last seven instances of the International Brain Tumor Segmentation (BraTS) challenge, i.e., 2012-2018. Specifically, we focus on i) evaluating segmentations of the various glioma sub-regions in pre-operative mpMRI scans, ii) assessing potential tumor progression by virtue of longitudinal growth of tumor sub-regions, beyond use of the RECIST/RANO criteria, and iii) predicting the overall survival from pre-operative mpMRI scans of patients that underwent gross total resection. Finally, we investigate the challenge of identifying the best ML algorithms for each of these tasks, considering that apart from being diverse on each instance of the challenge, the multi-institutional mpMRI BraTS dataset has also been a continuously evolving/growing dataset.
	\end{abstract}

	
	\section{Introduction}
	\label{section:introduction}
        \subsection{Scope}
            The Brain Tumor segmentation (BraTS) challenge focuses on the evaluation of state-of-the-art methods for the segmentation of brain tumors in multi-parametric magnetic resonance imaging (mpMRI) scans. Its primary role since its inception has been two-fold: a) a publicly available dataset and b) a community benchmark \cite{bratsTmiPaper,NatSciDataPaper,TciaGbmPaper,TciaLggPaper}. BraTS utilizes multi-institutional pre-operative mpMRI scans and focuses on the segmentation of intrinsically heterogeneous (in appearance, shape, and histology) brain tumors, namely gliomas. Furthermore, to pinpoint the clinical relevance of this segmentation task, BraTS 2018 also focuses on the prediction of patient overall survival, via integrative analyses of radiomic features and machine learning (ML) algorithms.

        \subsection{Clinical Relevance}
            Gliomas are the most common primary brain malignancies, with different degrees of aggressiveness, variable prognosis and various heterogeneous histological sub-regions, i.e., peritumoral edematous/invaded tissue, necrotic core, active and non-enhancing core. This intrinsic heterogeneity of gliomas is also portrayed in their imaging phenotype (appearance and shape), as their sub-regions are described by varying intensity profiles disseminated across mpMRI scans, reflecting varying tumor biological properties. Due to this highly heterogeneous appearance and shape, segmentation of brain tumors in multimodal MRI scans is one of the most challenging tasks in medical image analysis.

        \subsection{Before the BraTS era}
\begin{figure}[t!]
                \begin{centering}
                    \includegraphics[width=0.7\columnwidth]{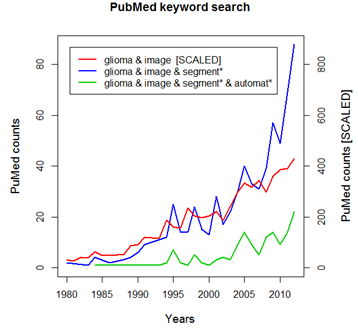}
                    \caption{Search on PubMed in 2012 showing related growing body of literature. Figure taken from \cite{bratsTmiPaper}.}
                    \label{fig:B18_stats}
                \end{centering}
        	\end{figure}

\begin{figure}[t!]
                \begin{centering}
                    \includegraphics[width=1\columnwidth]{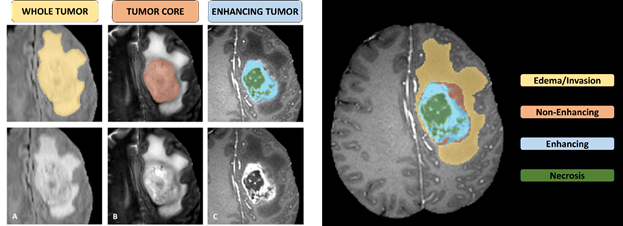}
                    \caption{Glioma sub-regions. The image patches show from left to right: the whole tumor (yellow) visible in T2-FLAIR (A), the tumor core (red) visible in T2 (B), the active tumor structures (light blue) visible in T1Gd, surrounding the cystic/necrotic components of the core (green) (C). The segmentations are combined to generate the final labels of the tumor sub-regions (D): ED (yellow), NET (red), NCR cores (green), AT (blue). Figure taken from \cite{bratsTmiPaper}.}
        	       \label{fig:B18_subregions}
                \end{centering}
            \end{figure}

            There has been a growing body of literature on computational algorithms addressing this important task (Fig.~\ref{fig:B18_stats}). Unfortunately, open manually-annotated datasets for designing and testing these algorithms are not currently available, and private datasets differ so widely that it is hard to compare the different segmentation strategies that have been reported so far. Critical factors leading to these differences include, but are not limited to, i) the imaging modalities employed, ii) the type of the tumor (glioblastoma or lower grade glioma, primary or secondary tumors, solid or infiltratively growing), and iii) the state of disease (images may not only be acquired prior to treatment, but also post-operatively and therefore show radiotherapy effects and surgically-imposed cavities). Towards this end, BraTS is making available a large dataset of mpMRI \cite{bratsTmiPaper,NatSciDataPaper,TciaGbmPaper,TciaLggPaper}, with accompanying delineations of the relevant tumor sub-regions (Fig.~\ref{fig:B18_subregions}). The exact mpMRI data consists of a) a native T1-weighted scan (T1), b) a post-contrast T1-weighted scan (T1Gd), c) a native T2-weighted scan (T2), and d) a T2 Fluid Attenuated Inversion Recovery (T2-FLAIR) scan.

        \subsection{BraTS 2017 vs 2018}
            The last two instances of BraTS (i.e., 2017 and 2018) were focused on both the segmentation of tumor sub-structures, and the prediction of overall survival of patients diagnosed with primary de novo glioblastoma (GBM).

            For the segmentation of gliomas in pre-operative mpMRI scans, the participants were called to address this task by using the provided clinically-acquired training data to develop automated methods and produce segmentation labels of the different glioma sub-regions.

\begin{figure}[t!]
                \begin{centering}
                    \includegraphics[width=1\columnwidth]{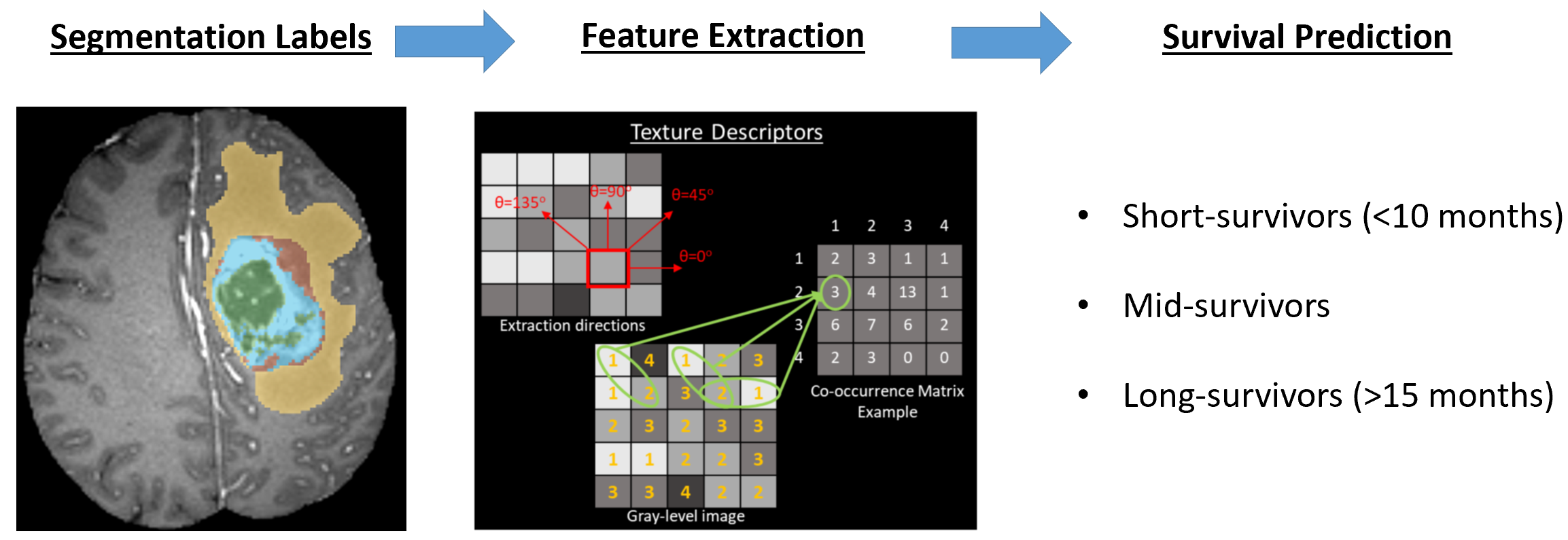}
                    \caption{Illustrative pipeline example for predicting patient overall survival.}
                    \label{fig:predictOS}
                \end{centering}
    	\end{figure}

            For the task of patient overall survival (OS) prediction from pre-operative mpMRI scans, once the participants produce their segmentation labels in the pre-operative scans, they were called to use these labels in combination with the provided mpMRI data to extract imaging/radiomic features that they consider appropriate \cite{IbsiPaper}, and analyze them through ML algorithms, to predict patient OS (Fig.~\ref{fig:predictOS}). The participants do not need to be limited to volumetric parameters, but can also consider intensity, morphologic, histogram-based, and textural features, as well as spatial information, and glioma diffusion properties extracted from glioma growth models.

    \section{Materials and Methods}
    \label{section:method}
	\noindent	
    	\subsection{BraTS Annotations and Structures}
            All the imaging datasets have been segmented manually, by one to four raters, following the same annotation protocol, and their ground truth annotations were approved by experienced neuro-radiologists. The tumor sub-regions considered for evaluation are: 1) the "active tumor" (AT), 2) the gross tumor, also known as the "tumor core" (TC), and 3) the complete tumor extent also referred to as the "whole tumor" (WT) (Fig.~\ref{fig:B18_subregions}). The AT is described by areas that show hyper-intensity in T1Gd when compared to T1, but also when compared to "healthy" white matter in T1Gd. The TC describes the bulk of the tumor, which is what is typically resected. The TC entails the AT, as well as the necrotic (fluid-filled) and the non-enhancing (solid) parts of the tumor. The appearance of the necrotic (NCR) and the non-enhancing (NET) tumor core is typically hypo-intense in T1-Gd when compared to T1. The WT describes the complete extent of the disease, as it entails the TC and the peritumoral edematous/invaded tissue (ED), which is typically depicted by hyper-intense signal in T2-FLAIR.

            The ground truth annotations were only approved by domain experts whereas they are actually created by multiple experts. Although a very specific annotation protocol (described below) was provided to each data contributing institution, slightly different annotation styles were noted for the various raters involved in the process. Therefore, all final labels included in the BraTS dataset were also further reviewed for consistency and compliance with the annotation protocol by a single board-certified neuro-radiologist with more than 15 years of experience.

        \subsection{Annotation Protocol}
            The BraTS dataset describes a collection of brain tumor MRI scans acquired from multiple different centers under standard clinical conditions, but with different equipment and imaging protocols, resulting in a vastly heterogeneous image quality reflecting diverse clinical practice across different institutions. However, we designed the following tumor annotation protocol, in order to make it possible to create similar ground truth delineations across various annotators.

            For the tasks related to BraTS, only structural MRI volumes were considered (T1, T1Gd, T2, T2-FLAIR), all of them co-registered to a common anatomical template (SRI \cite{sriPaper}) and resampled to 1$mm^3$. The details of the original scans are given in Table~\ref{tbl:brats_scansDetails}. Note that different native T1 scans exist, depending on whether they were 3D acquisitions, or 2D fast spin echo, or even just localizing images, and therefore not all T1 scans can be considered suitable for the task of segmentation. To our experience the T1Gd and the T2-FLAIR volumes have been the most useful to produce the ground truth segmentations.

            \begin{table}[!t]
                \centering
                \caption{Summarizing the original characteristics of the BraTS dataset.}
        	   	\begin{tabular}{|c|c|c|c|c|}
        	   		\hline
                    \textbf{Acronym} & \textbf{MRI Sequence} & \textbf{Property} & \textbf{Acquisition} & \textbf{Slice thickness}\\
        	   		\hline
                    \hline
        	   		T1 & T1-weighted & Native image & Sagittal or Axial & Variable (1-5mm)\\
        	   		\hline
        	   		T1Gd & T1-weighted & post-contrast enhancement (Gadolinium) & Axial 3D acquisition & Variable\\
        	   		\hline
        	   		T2 & T2-weighted & Native image & Axial 2D & Variable (2-4mm)\\
        	   		\hline
        	   		T2-FLAIR & T2-weighted & Native image & Axial or Coronal or Sagittal 2D & Variable\\
        	   		\hline
        	   	\end{tabular}
        	   	\label{tbl:brats_scansDetails}
            \end{table}

            We note that radiologic definition of tumor boundaries, especially in such infiltrative tumors as gliomas, is a well-known problem. In an attempt to offer a standardized approach to assess and evaluate various tumor sub-regions, the BraTS initiative, after consultation with internationally-recognized expert neuroradiologists, defined the following types of tumor sub-regions. However, we note that other criteria for delineation could be set, resulting in slightly different tumor sub-regions. The BraTS tumor sub-regions do not reflect strict biologic entities, but are rather image-based. For instance, the definition of the AT could simply be the regions with hyper-intense signal on T1Gd images. However, in high grade tumors, there are non-necrotic, non-cystic regions that do not enhance, but can be separable from the surrounding vasogenic edema, and represent non-enhancing infiltrative tumor. Another problem is the definition of tumor center in low-grade gliomas. In such cases, it is difficult to differentiate tumor from vasogenic edema, particularly in the absence of enhancement. It is also noteworthy that in order to produce the ground truth labels used in the provided data, we have recommended to start delineating the sub-regions of interest from the outside tumor boundaries, i.e., one should start from the manual delineation of the abnormal signal in the T2-weighted images, primarily defining the WT, then address the TC, and finally the enhancing and non-enhancing/necrotic core, possibly using semi-automatic tools.

            \subsubsection{BraTS 2012-2016 (Four tumor sub-regions)}
                BraTS 2012-2016 defined four tumor sub-regions, delineating the AT, NET, NCR, and ED.

                \begin{description}
                    \item[Label 1:] \textbf{NCR}. This sub-region describes the necrotic core, or necrocyst, that resides within the enhancing rim of high grade gliomas, and sometimes appears cystic.
                    \item[Label 2:] \textbf{ED}. This sub-region describes the peritumoral edematous and invaded tissue that is fairly easily defined on the T2-weighted images, as a hyperintense abnormal signal distribution, and hypo-intense signal on T1. This label primarily describes the tentacle-like shaped regions of edematous white matter into the subcortex of the gyri and, importantly, this is distinguished from cystic regions and the ventricles.
                    \item[Label 3:] \textbf{NET}. It is possible to identify such regions depicting the non-enhancing gross abnormality, by viewing the T2-weighted images. Some parts of the high-grade tumor do not enhance, but they are clearly distinguishable from the surrounding vasogenic edema on T2, as they have lower signal intensity and heterogeneous texture. Moreover, in low grade gliomas, this is the only category used for delineating the gross tumor.
                    \item[Label 4:] \textbf{AT}. This is a relatively easy definition, as it describes the enhancing regions within the gross tumor abnormality, but not the necrotic center. The threshold to exclude the necrotic center from the enhancing part should be set independently per subject. Note that vessels running in the neighboring regions and sulci are not included.
                \end{description}
                We cautiously note that the NET (i.e., "Label 3") can be overestimated by some annotators, and that oftentimes there is little evidence in the image data for this sub-region. Therefore, forcing the definition of this region could introduce an artifact, which could result in substantially different ground truth labels created from the annotators in different institutions. This case could potentially have implications in the ranking of the BraTS participants, i.e., a ranking bias towards the test cases ground truth annotator instead of ranking the actual algorithmic performance.

            \subsubsection{BraTS 2017-Present (Three tumor sub-regions)}
                In order to address the aforementioned issue, in BraTS 2017 the NET label ("Label 3") has been eliminated and combined with NCR ("Label 1"). Furthermore, contralateral and periventricular regions of T2-FLAIR hyper-intensity were excluded from the ED region, unless they were contiguous with peritumoral ED, as these areas are generally considered to represent chronic microvascular changes, or age-associated demyelination, rather than tumor infiltration \cite{Haller:pathologyAgeRelated}. The rationale for this is that contralateral and periventricular white matter hyper-intensities regions might be considered pre-existing conditions, related to small vessel ischemic disease, especially in older patients.

                \begin{description}
                    \item[WT:] \textbf{Segmenting the whole tumor extent (Union of all labels)}. One should start by loading the T2-FLAIR images and creating a new label for the WT. We recommend to start from the top of the brain (i.e., superiorly) and since this sub-region is usually the larger with a relatively smooth shape, it is sufficient to make manual delineations every third slice. Then morphological operations of dilation and erosion can be used to fill the in-between axial slices. Finally, smoothing with a Gaussian kernel ($\sigma=1$) can be used to smooth the jaggedness of the label on coronal and sagittal views.
                    \item[TC:] \textbf{Segmenting the gross tumor core outline (Union of labels 1, 3, and 4)}. For this sub-region, it is necessary to check whether there are non-enhancing tumor regions. The TC boundaries can be delineated on every other slice. Then, morphological operations of dilation and erosion can be used to fill the in-between axial slices, followed by a Gaussian smoothing filter to help with the non-continuous delineations on coronal view. Once the TC boundaries are defined, the remaining of the WT will correspond to the ED sub-region ("Label 2"), which is described by hyper-intense signal on the T2-FLAIR volumes.
                    \item[AT:] \textbf{Segmenting the active and the non-enhancing/necrotic tumor regions}. The active tumor (AT - i.e., enhancing rim) is described by areas that show hyper-intensity on T1-Gd when compared to T1, but also when compared to normal/healthy white matter (WM) in T1Gd. Biologically, AT is felt to represent regions where there is leakage of contrast through a disrupted blood-brain barrier that is commonly seen in high grade gliomas. The NET represents non-enhancing tumor regions, as well as transitional/prenecrotic and necrotic regions that belong to the non-enhancing part of the TC, and are typically resected in addition to the AT. The appearance of the NET is typically hypo-intense in T1-Gd when compared to T1, but also when compared to normal/healthy WM in T1-Gd.

                        To delineate the AT in gliomas, we suggest to use the T1Gd scans and the existing TC outline. One can then set an intensity threshold within this label to distinguish between the high intensity active/enhancing tumor and the low intensity non-enhancing/necrotic (and very tortuous) core regions. Note that the choroid plexus and areas of hemorrhage (when they can be identified by comparing to the native T1 scan), should not be labeled.
                    \item[LGG:] \textbf{Remarks on low grade gliomas}. For low grade gliomas (LGGs), we note that they do not exhibit much contrast enhancement, or ED. Biologically, LGGs may have less blood-brain barrier disruption (leading to less leak of contrast during the scan), and may grow at a rate slow enough to avoid significant edema formation, which results from rapid disruption, irritation, and infiltration of normal brain parenchyma by tumor cells. Specifically, after taking all the above into consideration, in scans of LGGs without an apparent ET area, we consider only the NET and vasogenic ED labels, by observing the texture or the intensity on T2-FLAIR images, whereas in LGG scans without ET and without obvious texture differences across modalities (e.g., small astrocytomas) we consider only the NET label, distinguishing between normal and abnormal brain tissue. The difficulty in estimating the accurate boundaries between tumor and healthy tissue in the operating room is reflected in the segmentation labels as well; there is high uncertainty among neurosurgeons, neuroradiologists, and imaging scientists in delineating these boundaries.
                \end{description}

        \subsection{The BraTS Data Since its Inception}
            The mpMRI scans made publicly available through the BraTS initiative, describe T1, T1Gd, T2, and T2-FLAIR volumes, acquired with different clinical protocols and various scanners from multiple institutions, mentioned as data contributors in the acknowledgements section. The provided data are distributed after their harmonization, following standardization pre-processing without affecting the apparent information in the images. Specifically, the pre-processing routines applied in all the BraTS mpMRI scans include co-registration to the same anatomical template \cite{sriPaper}, interpolation to a uniform isotropic resolution (1$mm^3$), and skull-stripping.

            \subsubsection{Continuously Growing Publicly Available Dataset}

                The BraTS dataset has evolved over the years (2012-2018) with a continuously increasing number of patient cases, as well as through an improvement of the data split used for algorithmic development and evaluation (Table~\ref{tbl:brats_data}).

                The first two instances of BraTS (2012-2013) comprised a training and a testing dataset of 35 and 15 mpMRI patient scans, respectively. The results and findings of these two first editions, were summarized in \cite{bratsTmiPaper}, which to date is the most popular and downloaded paper of the IEEE TMI journal since its publication, and reflects the interest of the scientific research community in the BraTS initiative as a publicly available dataset and a community benchmark.

                The subsequent three instances of BraTS (2014-2016) received a substantial dataset increase in two waves and also included longitudinal mpMRI scans. The first wave of increase came in during 2014-2015 primarily from contributions of The Cancer Imaging Archive (TCIA) repository \cite{TciaPaper} and then Heidelberg University, and the second wave of increase happened in 2016 with contributions from the Center for Biomedical Image Computing and Analytics (CBICA) at the University of Pennsylvania (UPenn). In addition, stemming from the analysis of the BraTS 2012-2013 results \cite{bratsTmiPaper}, BraTS 2014-2016 employed ground truth data created by label fusion of top-performing approaches.

                In 2017, thanks to additional contributions to the BraTS dataset, from CBICA@UPenn and the University of Alabama in Birmingham (UAB), a validation set was included to facilitate algorithm fine-tuning following a ML paradigm of training, validation, and testing datasets. Notably, in 2017 the number of cases was doubled with respect to the previous year, amounting to 477 cases, which was further increased in 2018 with 542 cases, thanks to contributions from MD Anderson Cancer Center in Texas, the Washington University School of Medicine in St. Louis, and the Tata Memorial Center in India.

            \subsubsection{Focus Beyond Segmentation}
                BraTS, as indicated by its acronym definition, has primarily focused on the segmentation to brain tumor sub-regions. However, after its first instances (2012-2013), its potential clinical relevance became apparent.

                BraTS was introduced with secondary tasks, where the results of the brain tumor segmentation algorithms are used towards promoting further analysis and accelerating discovery. From a clinical perspective these secondary tasks featured in the BraTS challenge can be crucial towards fostering the development of algorithms capable of addressing clinical requirements in a more reliable manner than the current clinical practice. Specifically, to pinpoint the clinical relevance of the segmentation task, in the BraTS instances of 2014-2016, longitudinal scans were included in the publicly available dataset, to evaluate the ability and potential of automated tumor volumetry in assessing disease progression. Along the same lines of research, in the last two instances of BraTS (2017-2018), clinical data of patient age, overall survival, and resection status were included, to facilitate the secondary task of predicting patient overall survival via integrative analyses of radiomic features and ML algorithms.

                \subsubsection{The Latest BraTS Data}
                The datasets used in BraTS 2017 and 2018 have been updated (since BraTS 2016), with more routine clinically-acquired 3T mpMRI scans and all the ground truth labels have been evaluated, and manually-revised when needed, by expert board-certified neuroradiologists. Ample multi-institutional (n=19) routine clinically-acquired pre-operative mpMRI scans of GBM/HGG and LGG, with pathologically confirmed diagnosis and available OS, were provided as the training, validation and testing data.

                The data provided since BraTS 2017 differs significantly from the data provided during the previous BraTS challenges (i.e., 2016 and backwards). Specifically, since BraTS 2017, expert neuroradiologists have radiologically assessed the complete original TCIA glioma collections (i.e., TCGA-GBM, n=262 \cite{TciaTcgaGbmPaper} and TCGA-LGG, n=199 \cite{TciaTcgaLggPaper}) and categorized each scan as pre-operative or post-operative. Subsequently, all the pre-operative TCIA scans (i.e., 135 GBM \cite{TciaGbmPaper} and 108 LGG \cite{TciaLggPaper}) were annotated by experts for the various sub-regions and included in the BraTS datasets \cite{NatSciDataPaper,TciaGbmPaper,TciaLggPaper}.

                \subsubsection{Data Availability}
                As one of the main objectives of the BraTS initiative is to provide an open source repository for continuous development of algorithms, the data of BraTS 2012-2016 has been made available through the Swiss Medical Image Repository (SMIR - www.smir.ch), and the data of BraTS 2017-2018 through the Image Processing Portal of the CBICA@UPenn (IPP - \url{ipp.cbica.upenn.edu}). Both platforms feature downloading of datasets, as well as the automatic evaluation of the results submitted by participants.

            	\begin{table}[!t]
                \centering
                    \caption{Summarizing the distribution of the BraTS data across the training, validation, and testing sets, since the inception the of BraTS initiative, together with the focused tasks of its BraTS instance.}
            	   	\begin{tabular}{|c|c|c|c|c|c|c|}
            	   		\hline
                        \multirow{2}{*}{\textbf{Year}} & \textbf{Total} & \textbf{Training} & \textbf{Validation} & \textbf{Testing} & \textbf{Tasks} & \textbf{Type of data}\\
                                        {} & \textbf{data}  & \textbf{data}     & \textbf{data}   & \textbf{data}    & {}    & {}\\
            	   		\hline
                        \hline
            	   		2012 & 50 & 35 & N/A & 15 & Segmentation & Pre-operative only\\
            	   		\hline
            	   		2013 & 60 & 35 & N/A & 25 & Segmentation & Pre-operative only\\
            	   		\hline
            	   		\multirow{2}{*} {2014} & {238} & {200} & {N/A} & {38} & Segmentation & {Longitudinal}\\
                                          {}   &   {}  &   {}  &   {}  &   {} & Disease progression & {}\\
            	   		\hline
            	   		\multirow{2}{*} {2015} & 253 & 200 & N/A & 53 & Segmentation & Longitudinal\\
                                          {}   &   {}  &   {}  &   {}  &   {} & Disease progression & {}\\
            	   		\hline
            	   		\multirow{2}{*} {2016} & 391 & 200 & N/A & 191 & Segmentation & Longitudinal\\
                                          {}   &   {}  &   {}  &   {}  &   {} & Disease progression & {}\\
            	   		\hline
            	   		\multirow{2}{*} {2017} & 477 & 285 & 46 & 146 & Segmentation & Pre-operative only\\
                                          {}   &   {}  &   {}  &   {}  &   {} & Survival prediction & {}\\
            	   		\hline
            	   		\multirow{2}{*} {2018} & 542 & 285 & 66 & 191 & Segmentation & Pre-operative only\\
                                          {}   &   {}  &   {}  &   {}  &   {} & Survival prediction & {}\\
            	   		\hline
            	   	\end{tabular}
            	   	\label{tbl:brats_data}
                \end{table}

            \subsubsection{The Ranking Scheme for the Segmentation Task (BraTS 2017-2018)}
                The ranking scheme followed during the BraTS 2017 and 2018 comprised the ranking of each team relative to its competitors for each of the testing subjects, for each evaluated region (i.e., AT, TC, WT), and for each measure (i.e., Dice and Hausdorff (95\%)). For example, in BraTS 2018, each team was ranked for 191 subjects, for 3 regions, and for 2 metrics, which resulted in 1146 individual rankings. The \emph{final ranking score} (FRS) for each team was then calculated by firstly averaging across all these individual rankings for each patient (i.e., \emph{Cumulative Rank}), and then averaging these cumulative ranks across all patients for each participating team. This ranking scheme has also been adopted in other challenges with satisfactory results, such as the Ischemic Stroke Lesion Segmentation (ISLES - \url{http://www.isles-challenge.org/}) challenge \cite{IslesPaper2015,IslesPaper2016}.

                We also conducted further permutation testing, to determine statistical significance of the relative rankings between each pair of teams. This permutation testing would reflect differences in performance that exceeded those that might be expected by chance. Specifically, for each team we started with a list of observed subject-level \emph{Cumulative Ranks}, i.e., the actual ranking described above. For each pair of teams, we repeatedly randomly permuted (i.e., 100,000 times) the \emph{Cumulative Ranks} for each subject. For each permutation, we calculated the difference in the FRS between this pair of teams. The proportion of times the difference in FRS calculated using randomly permuted data exceeded the observed difference in FRS (i.e., using the actual data) indicated the statistical significance of their relative rankings as a p-value. These values were reported in an upper triangular matrix.

            \subsubsection{Prediction of Patient Overall Survival (BraTS 2017-2018)}
                We identified 346 GBM patients with overall survival (OS), age, and resection status information. 164 of them had undergone surgery with gross total resection (GTR) status. The distributions of OS of GBM patients across the training, validation and testing datasets were matched (Table~\ref{tbl:B18_population_stats}). The patients were divided in three groups of survival comprising long-survivors (who survived more than 15 months), short-survivors (who survived less than 10 months), and mid-survivors (who survived between 10 and 15 months). These thresholds were derived after statistical consideration of the survival distributions across the complete dataset. Specifically, we chose these thresholds based on equal quantiles from the median OS (approximately 12.5 months) to avoid potential bias towards one of the survival groups (short- vs long- survivors) and while considering that discrimination of groups should be clinically meaningful. The median OS of the described cohorts is not significantly different from the median OS of GBM patients in several randomized Phase III trials, noting that our cohort consists of unselected patients rather than those eligible for such trials \cite{stuppLancet,tmzTrial}.

                The population of patients with available OS information was randomly and proportionally divided into the training, validation and testing sets. This process formed a) the training set, consisting of 163 cases (59 with GTR), b) the validation set, consisting of 53 cases (28 with GTR), and c) the testing set, consisting of 130 cases (77 with GTR). Table \ref{tbl:B18_population_stats} shows the distribution of patient cases for the task of the OS prediction.

                Participating teams were requested to submit OS prediction results in days for each patient with GTR. The evaluation system then automatically classified these into short-, intermediate-, and long-survivors.

                \begin{table}[!t]
                    \centering
                    \caption{The overall survival distribution of patients across the training, validation, and testing sets of BraTS 2017 and 2018.}
                    \begin{tabular}{c|ccc}
                        \textbf{} & \textbf{Training Data} & \textbf{Validation Data} & \textbf{Testing Data}\\
                        \hline
                        \hline
                        \multirow{2}{*}{\raisebox{-1.15\height}{\rotatebox{90}{\textbf{All subjects (2017)}}}} & n=93 & n=60 & n=174\\
                        \noindent
                        {}  & \begin{tikzpicture}
                                \node{\includegraphics[width=0.3\columnwidth]{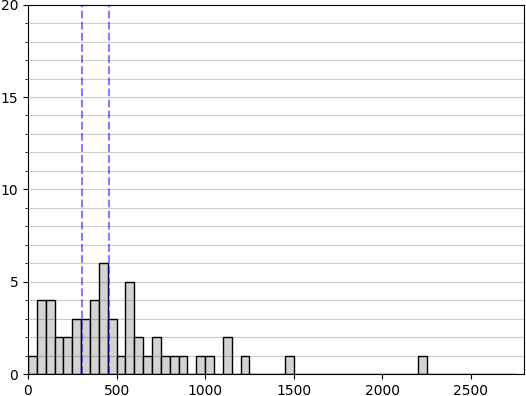}};
                                \end{tikzpicture}
                            & \begin{tikzpicture}
                                \node{\includegraphics[width=0.3\columnwidth]{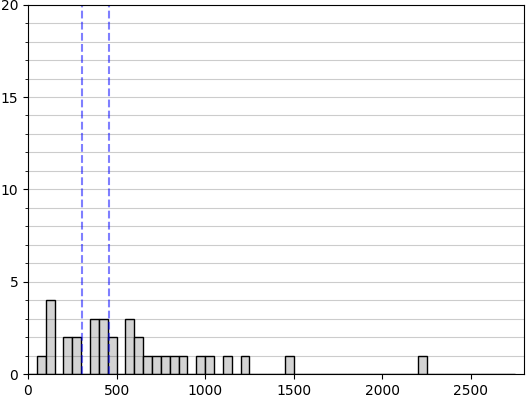}};
                                \end{tikzpicture}
                            & \begin{tikzpicture}
                                \node{\includegraphics[width=0.3\columnwidth]{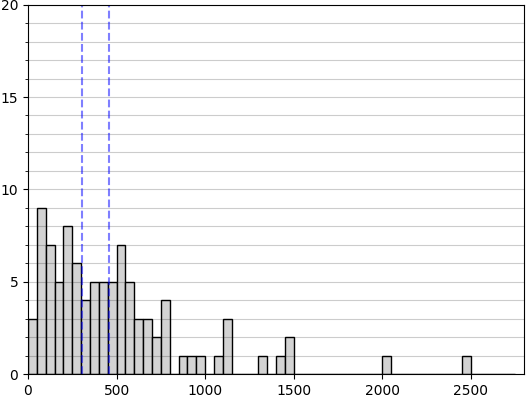}};
                                \end{tikzpicture}\\
                        \hline
                        \hline
                        \multirow{2}{*}{\raisebox{-1.15\height}{\rotatebox{90}{\textbf{All subjects (2018)}}}} & n=283 & n=81 & n=241\\
                        \noindent
                        {}  & \begin{tikzpicture}
                                \node{\includegraphics[width=0.3\columnwidth]{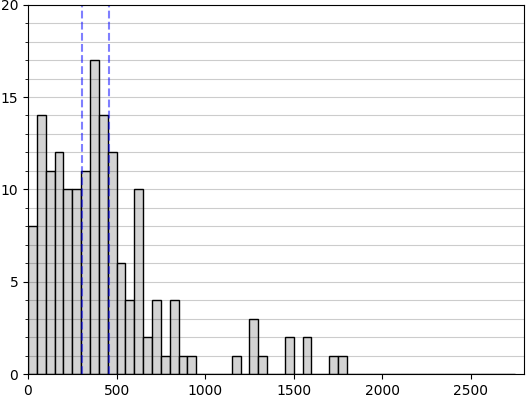}};
                                \end{tikzpicture}
                            & \begin{tikzpicture}
                                \node{\includegraphics[width=0.3\columnwidth]{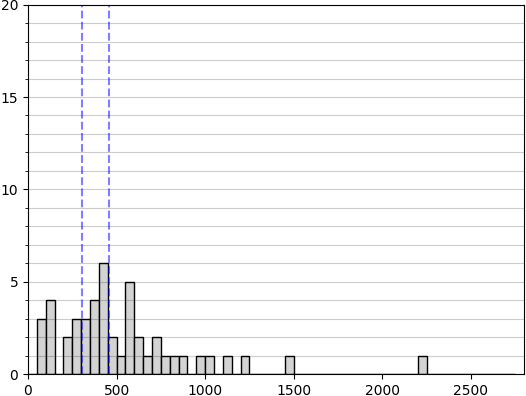}};
                                \end{tikzpicture}
                            & \begin{tikzpicture}
                                \node{\includegraphics[width=0.3\columnwidth]{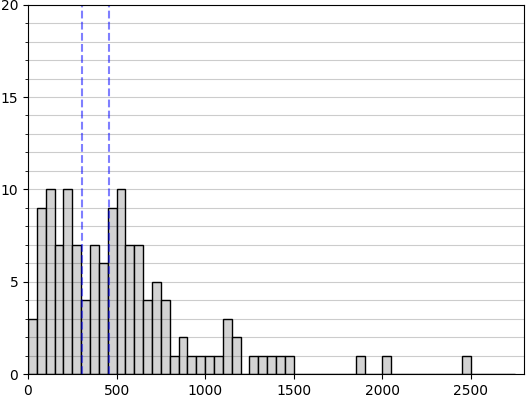}};
                                \end{tikzpicture}\\
                        \hline
                        \hline
                        \multirow{2}{*}{\raisebox{-0.5\height}{\rotatebox{90}{\textbf{Only GTR subjects (2018)}}}} & n=102 & n=47 & n=142\\
                        {} & \begin{tikzpicture}
                                \node{\includegraphics[width=0.3\columnwidth]{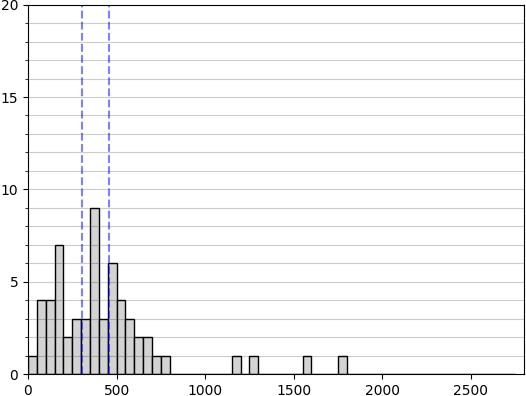}};
                                \end{tikzpicture}
                            & \begin{tikzpicture}
                                \node{\includegraphics[width=0.3\columnwidth]{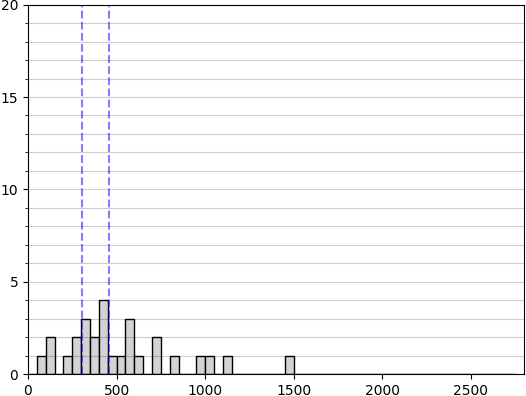}};
                                \end{tikzpicture}
                            & \begin{tikzpicture}
                                \node{\includegraphics[width=0.3\columnwidth]{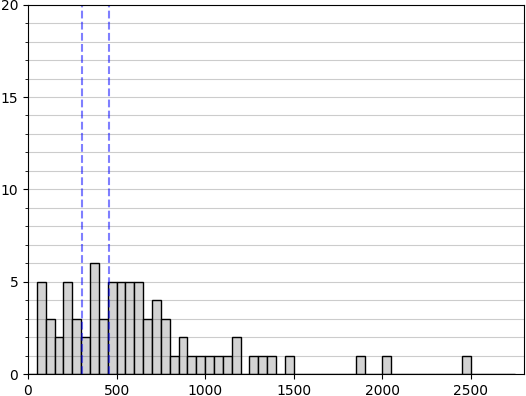}};
                                \end{tikzpicture}\\
                        \hline
                    \end{tabular}
                    \label{tbl:B18_population_stats}
                \end{table}

            \subsubsection{Evaluation Framework}
                For consistency purposes both in BraTS 2017 \& 2018 challenges, two reference standards were used for the two tasks of the challenge: 1) manual segmentation labels of tumor sub-regions, and 2) clinical data of OS.

                The introduction of the validation set since BraTS 2017 allows participants to obtain preliminary results in unseen data, in addition to their cross-validated results on the training data. The ground truth of the validation data was never provided to the participants. Finally, all participants were presented with the same test data, for a limited controlled time-window (48h), before the participants are required to submit their final results for quantitative evaluation and their ranking.

                For the segmentation task, and for consistency with the configuration of the previous BraTS challenges, the "Dice score" and the "Hausdorff distance" were used. Expanding upon this evaluation scheme, the metrics of "Sensitivity" and "Specificity" were also used, allowing to determine potential over- or under-segmentations of the tumor sub-regions by participating methods. Since the BraTS 2012-2013 were subsets of the BraTS 2018 test data, performance comparison on the 2012-2013 data will allow for a direct evaluation against the performances reported in \cite{bratsTmiPaper}.

                For the task of survival prediction, two evaluation schemes are considered. First, for ranking the participating teams, evaluation will be based on the classification of subjects as long-, intermediate-, and short-survivors. Predictions of the participating teams will be assessed based on classification accuracy (i.e. the number of correctly classified patients) with respect to this grouping. Note that participants are expected to provide predicted survival status only for subjects with resection status of GTR (i.e., Gross Total Resection). In addition, a pairwise error analysis between the predicted and actual survival in days was conducted and the results were shared with the participants, to allow the evaluation of their method for outliers. This analysis was done using the metrics of Mean-Square Error (MSE), median square error (medianSE), standard deviation of the square errors (stdSE), and the spearman correlation coefficient (spearmanR).

    \section{Results}
    \label{section:results}
        \subsection{BraTS 2012-2013}
\begin{figure}[t!]
                    \begin{centering}
                        \includegraphics[width=0.7\columnwidth]{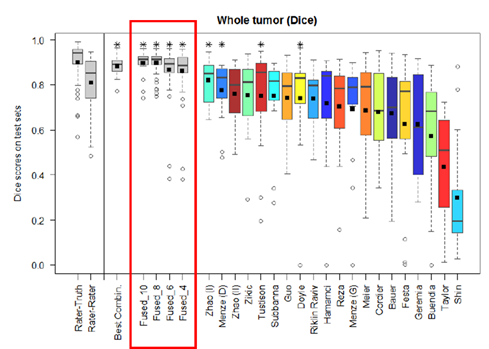}
                        \caption{Summary results of the BraTS 2012-2013. Label fusion (red outline) out-performs all individual methods and the inter-rater agreement. Figure adopted from \cite{bratsTmiPaper}.}
                        \label{fig:B18_2013_results}
                    \end{centering}
            	\end{figure}

                To emphasize the most interesting results of our previously published analyses summarizing BraTS 2012 and BraTS 2013 \cite{bratsTmiPaper}, we focus into two main points (Fig.~\ref{fig:B18_2013_results}). First, we note that even though most of the individual automated segmentation methods performed well, they did not outperform the inter-rater agreement, across expert clinicians, who have been trained for years to identify regions of infiltration and distinguish them from healthy brain. Secondly, the fusion of segmentation labels from top-ranked algorithms out-performed all individual methods and was comparable to inter-rater agreement. More specifically, while we observe that individual automated segmentation methods do not necessarily rank equally well in the different tumor segmentation tasks and under all metrics (i.e., when evaluating WT, TC, and AT segmentation, with respect to Dice score and Hausdorff distance), we note that the fused segmentation labels do consistently rank first in all tasks and both metrics. This suggests that ensembles of fused segmentation algorithms may be the favorable approach when translating tumor segmentation methods into clinical practice.

            \subsection{BraTS 2017 (Testing Phase)}
                During the testing phase of the BraTS 2017 challenge, we note participation of 48 independent teams \cite{brats17:lncs:MIRL,brats17:prepro:whatapain,brats17:prepro:cian,brats17:prepro:qtim,brats17:lncs:biomedia1BHN, brats17:prepro:drcubic,brats17:lncs:UPCDLMI,brats17:lncs:BCVUniandes,brats17:prepro:SCUTEE,brats17:lncs:CNEA, brats17:prepro:CISA,brats17:prepro:niftynet,brats17:prepro:xfeng,brats17:lncs:NPU,brats17:lncs:MICDKFZ, brats17:lncs:NUS_MPR,brats17:lncs:pvg,brats17:lncs:ISTB,brats17:lncs:biomedia1,brats17:prepro:BrICLab, brats17:lncs:Rocky,brats17:lncs:neuro.ml,brats17:lncs:OnePiece,brats17:prepro:ZejuLi:FDUBME,brats17:prepro:Mountain, brats17:prepro:SJTU,brats17:lncs:UCCS,brats17:prepro:PADAS,brats17:lncs:Bern,brats17:lncs:Alexander, brats17:lncs:MBI,brats17:lncs:Ashi,brats17:lncs:ROB,brats17:prepro:justdoit,brats17:lncs:HPI-Ultimate, brats17:lncs:SaraS,brats17:lncs:BRATZZ27,brats17:lncs:VisionLab,brats17:prepro:Dundee,brats17:lncs:LoVE, brats17:lncs:UCL-TIG,brats17:prepro:STH,brats17:prepro:CMR,brats17:lncs:Zhao,brats17:prepro:SegLZ, brats17:lncs:BIGS2,brats17:prepro:Zhouch,brats17:prepro:CamMIA}. Specifically, results for the segmentation task were submitted by 47 teams and for the survival prediction task by 16 teams (1 of which did not participate in the segmentation task).

\begin{figure}[t!]
                    \begin{centering}
                        \includegraphics[width=1\columnwidth]{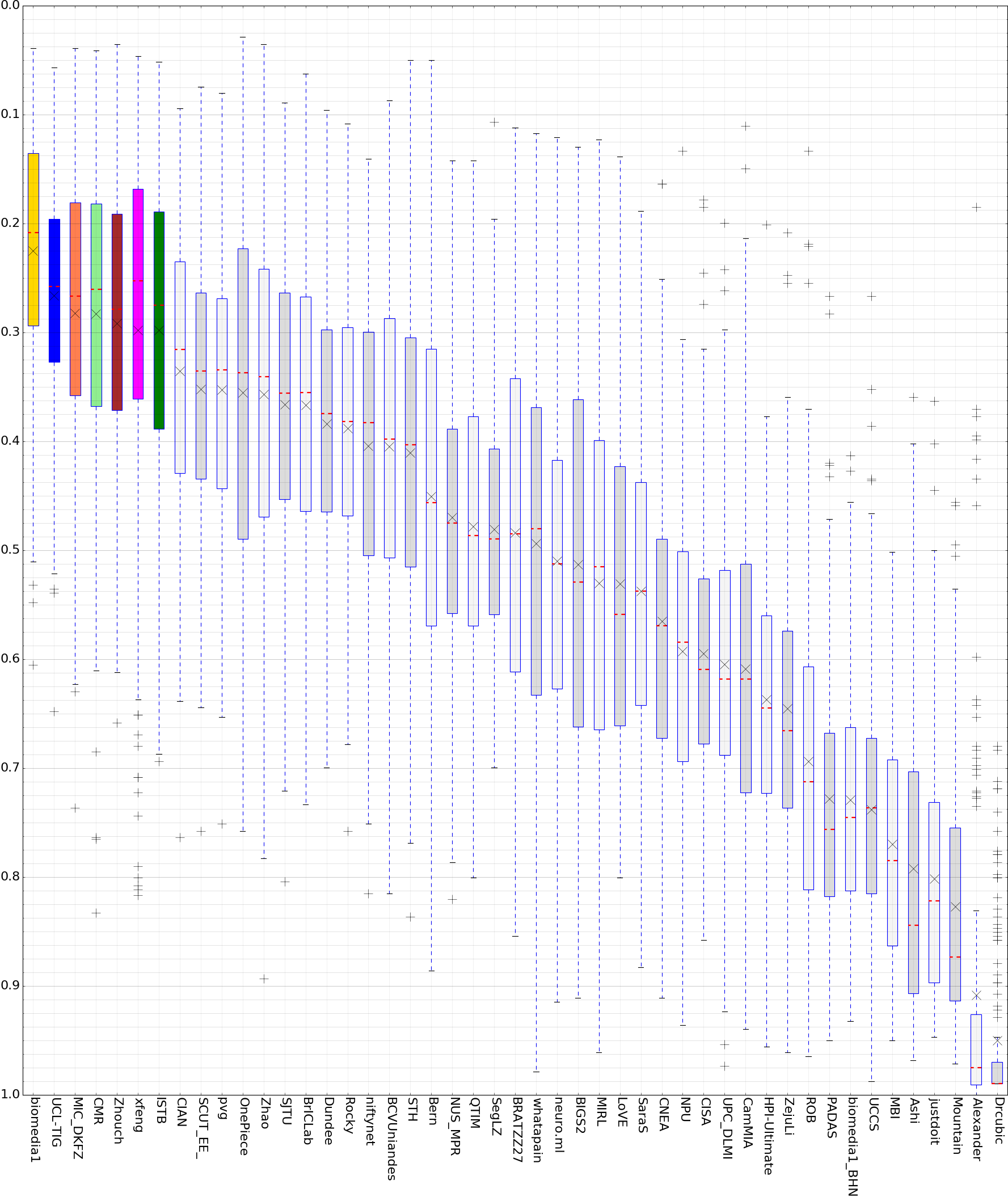}
                        \caption{BraTS 2017 Ranking of all Participating Teams in Segmentation Task. (smaller values are higher ranks)}
                        \label{fig:B17_2017_rankings}
                    \end{centering}
            	\end{figure}

\begin{figure}[t!]
                    \begin{centering}
                        \includegraphics[width=1\columnwidth]{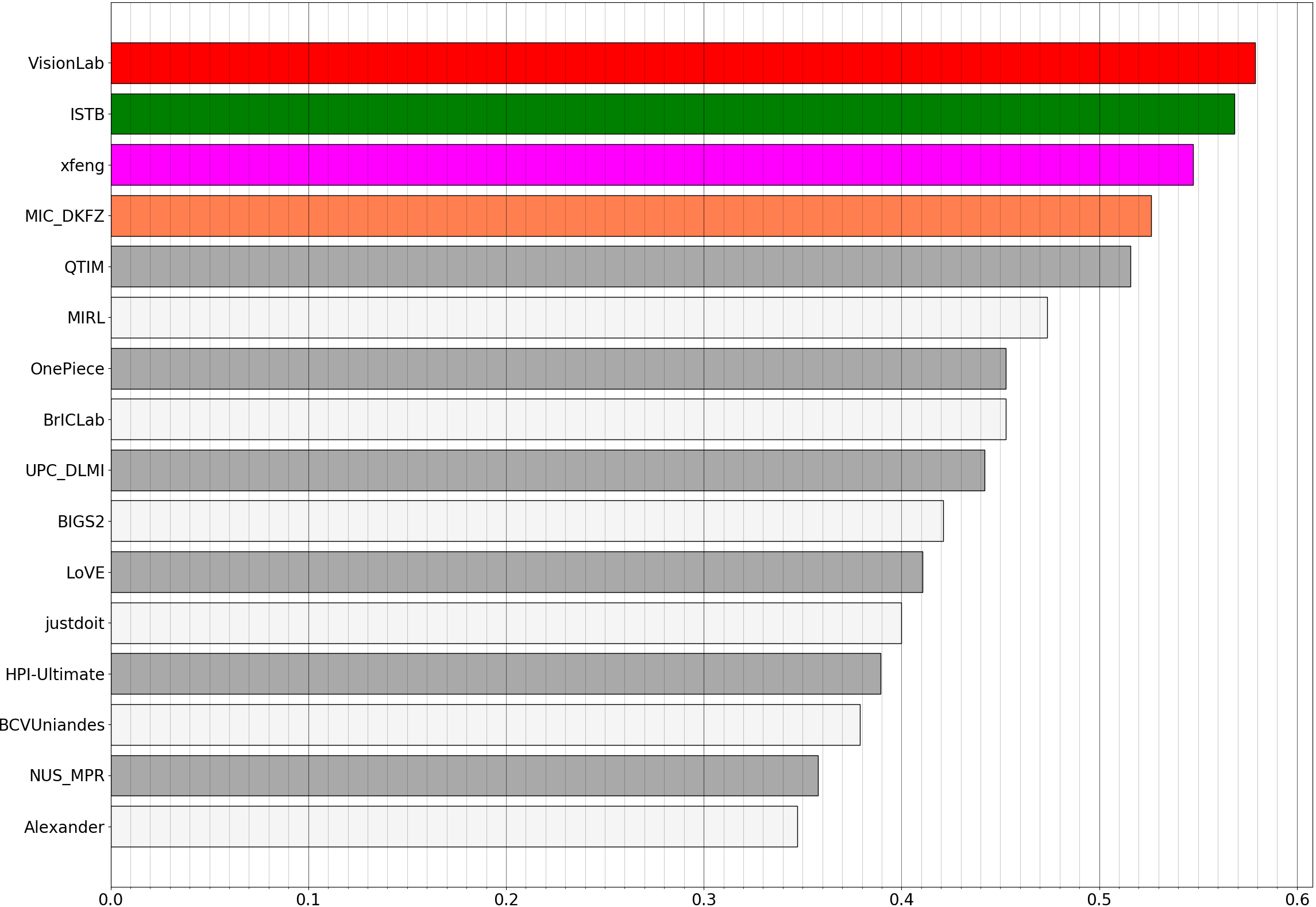}
                        \caption{BraTS 2017 Ranking of all Participating Teams in Survival Task. (larger values are better)}
                        \label{fig:B17_2017_rankings_survival}
                    \end{centering}
            	\end{figure}

                The ranking of the participating teams depicts a gradual improvement of the ranked approaches (Fig.~\ref{fig:B17_2017_rankings}-\ref{fig:B17_2017_rankings_survival}). We note that the variability of the ranked approaches (Fig.~\ref{fig:B17_2017_rankings}) does not dramatically change across any two sequentially ranked teams, indicating no particular dominance of a method over the other closely ranked methods. In order to assess potential statistically significant performance differences across teams, we also performed a pairwise comparison for significant differences based on 100,000 permutations. This allowed us to include a tie in the 3rd rank of the segmentation task (Table~\ref{tbl:2017_winners}). Specifically, the statistical evaluation of the top-ranked teams revealed that the first team was statistically better from the second (p-value<0.0003), whereas the second team was not statistically better than the third (p>0.1) and the fourth (p>0.14), but only from the fifth (p=0.01). This justified the decision of a tie in the third rank.

                \begin{table}[!t]
                    \centering
                    \caption{Top-ranked participating teams in BraTS 2017 for both the segmentation and the survival prediction tasks.}
                    \begin{tabular}{|c|c|c|c|p{7cm}|c|}
                        \hline
                        \textbf{Task} & \textbf{Rank} & \textbf{Team} & \textbf{First Author} & \textbf{Institution} & \textbf{Paper}\\
                        \hline
                        \hline
                        \multirow{4}{*}{\raisebox{-1.1\height}{\rotatebox{90}{Segmentation}}} & 1 & biomedia1 & Konstantinos Kamnitsas & Imperial College London, UK & \cite{brats17:lncs:biomedia1}\\
                        \cline{2-6}
                        {} & 2 & UCL-TIG & Guotai Wang & University College London (UCL), UK & \cite{brats17:lncs:UCL-TIG}\\
                        \cline{2-6}
                        {} & 3 (tie) & MIC\_DKFZ & Fabian Isensee & Division of Medical Image Computing, German Cancer Research Center (DKFZ), Heidelberg, Germany & \cite{brats17:lncs:MICDKFZ}\\
                        \cline{2-6}
                        {} & 3 (tie) & CMR & Tsai-Ling Yang & National Taiwan University of Science and Technology, Taipei, Taiwan & \cite{brats17:prepro:CMR}\\
                        \hline
                        \hline
                        \multirow{4}{*}{\raisebox{-1\height}{\rotatebox{90}{Survival}}} & 1 & VisionLab & Zeina Shboul & Old Dominion University, USA & \cite{brats17:lncs:VisionLab}\\
                        \cline{2-6}
                        {} & 2 & UBERN\_UCLM & Alain Jungo & University of Bern, Switzerland & \cite{brats17:lncs:ISTB}\\
                        \cline{2-6}
                        {} & 3 & xfeng & Xue Feng & Biomedical Engineering, University of Virginia, USA & \cite{brats17:prepro:xfeng}\\
        	   		\hline
        	   	\end{tabular}
        	   	\label{tbl:2017_winners}
        	   \end{table}

            \subsection{BraTS 2018 (Testing Phase)}
                During the testing phase of the BraTS 2018 challenge, we note participation of 63 independent teams \cite{brats18:1,brats18:rank3:surv:tata,brats18:2,brats18:3,brats18:prepro:cabezas,brats18:4,brats18paragios,brats18:prepro:chang,brats18:5,brats18:6, brats18:7,brats18:prepro:fang,brats18:rank1:surv:feng,brats18:prepro:fridman,brats18:8,brats18:9,brats18biros,brats18:10,brats18:prepro:huKong, brats18:12:bjoern,brats18:11,brats18:13,brats18:prepro:vivek,brats18:rank2:seg:fabian,brats18:14,brats18:15,brats18:16,brats18:17,brats18:18, brats18:19,brats18:prepro:XLi,brats18:prepro:Liu,brats18:20,brats18:21,brats18:rank3:seg:mckinley,brats18:22,brats18:prepro:Monteiro, brats18:rank1:seg:nvidia,brats18:23,brats18:prepro:Popli,brats18:24,brats18:rank2:surv:elodie,brats18:prepro:Ren,brats18:25, brats18:26,brats18:27:ODU,brats18:prepro:Deepnoid,brats18:28:stryker,brats18:rank2:surv:lisun,brats18:29:mauricio,brats18:prepro:Tseng, brats18:30,brats18:31:kcl,brats18:prepro:ChiatseWang,brats18:rank3:surv:leon,brats18:prepro:ShaochengWu,brats18:prepro:PeiyuanXu, brats18:prepro:XiaowenXu,brats18:32, brats18:33,brats18:34,brats18:prepro:XiaoyueZhang,brats18:rank3:seg:chenhong}. Specifically, results for the segmentation task were submitted by 61 teams and for the survival prediction task by 26 teams (2 of which did not participate in the segmentation task).

\begin{figure}
        \begin{centering}
            \includegraphics[width=1\columnwidth]{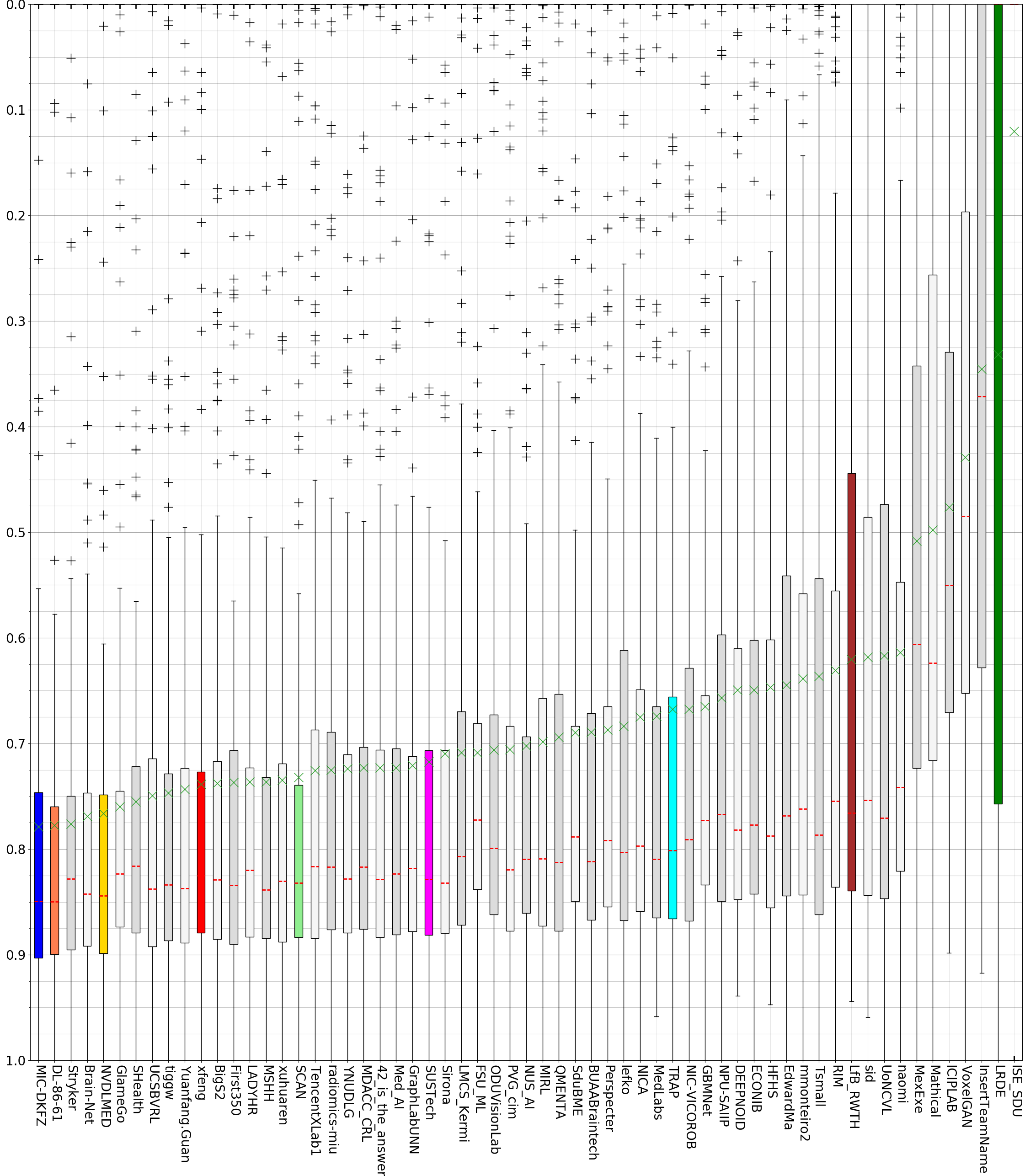}
            \caption{BraTS 2018 summarizing results (Dice) for the segmentation of the active tumor compartment.}
            \label{fig:sup:diceET}
        \end{centering}
    \end{figure}

\begin{figure}
        \begin{centering}
            \includegraphics[width=1\columnwidth]{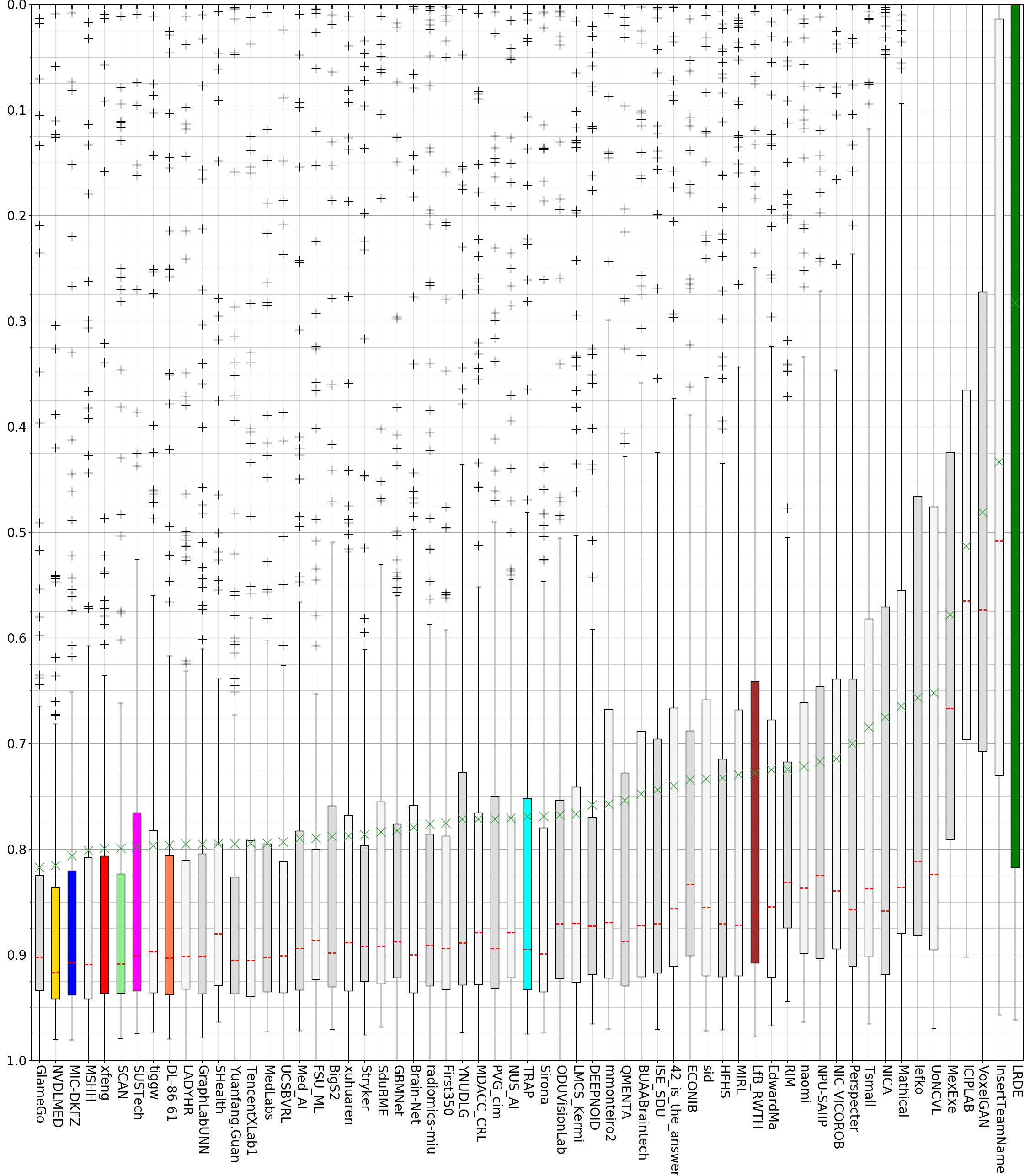}
            \caption{BraTS 2018 summarizing results (Dice) for the segmentation of the tumor core compartment.}
            \label{fig:sup:diceTC}
        \end{centering}
    \end{figure}

\begin{figure}
        \begin{centering}
            \includegraphics[width=1\columnwidth]{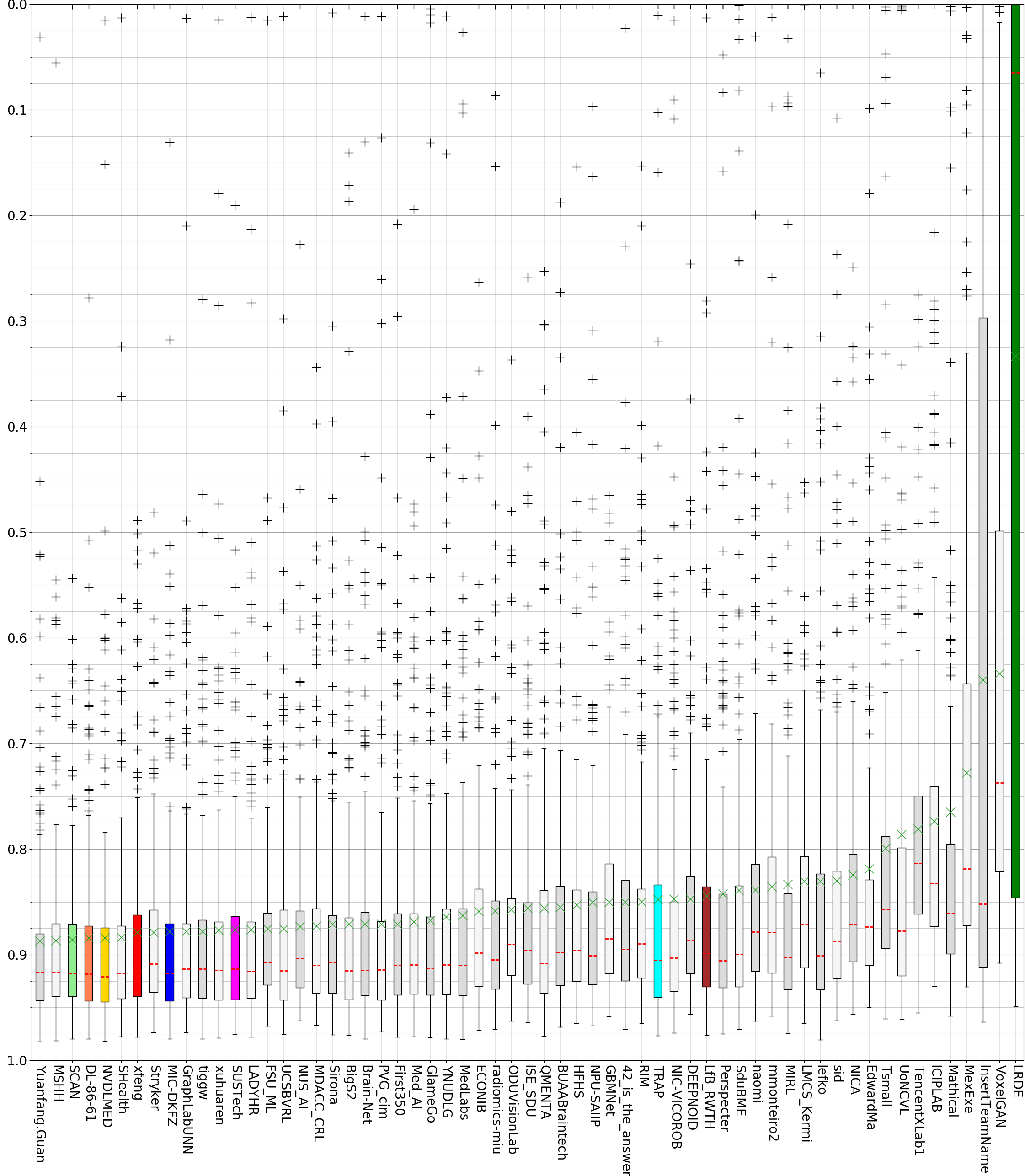}
            \caption{BraTS 2018 summarizing results (Dice) for the segmentation of the whole tumor compartment.}
            \label{fig:sup:diceWT}
        \end{centering}
    \end{figure}

                The BraTS 2018 results for the segmentation of the AT (Suppl.Fig.~\ref{fig:sup:diceET}) show a very marked skewness in the distribution of Dice metrics, as seen in the average and median values (crosses and vertical lines on each boxplot). These results illustrate the tendency of most methods to perform relatively well, in terms of median Dice (Median Dice for top 54/63 teams: [0.74-0.85]), but also the difference in levels of robustness as the average Dice is affected by increasing number of outliers in the results (Average Dice of same 54/63 teams: [0.61-0.77]). Segmentation results of the TC (Suppl.Fig.~\ref{fig:sup:diceTC}) presents a similar pattern, with the results of the AT, across teams. Similarly with observations from previous BraTS instances \cite{bratsTmiPaper}, the top positions are not systematically taken by the same teams, reflecting the added value of fusing segmentation labels from different approaches. In comparison to the AT, segmentation of the TC seems in general to be more robust (i.e., median inter-quantile range (IQR) for Dice of same 54/63 teams, for TC is 0.16, vs. 0.18 for the AT). It is worth mentioning though that the Dice metric is more sensitive to error of the AT, due to its typically much smaller volume. As also noted in previous instances of BraTS, the segmentation of the WT (Suppl.Fig.~\ref{fig:sup:diceWT}) represents the most robust and accurate segmentation results of the three evaluated tumor compartments (i.e., AT, TC, WT), with a median Dice coefficient of 0.9 for most of the participating teams.

\begin{figure}
        \begin{centering}
            \includegraphics[width=1\columnwidth]{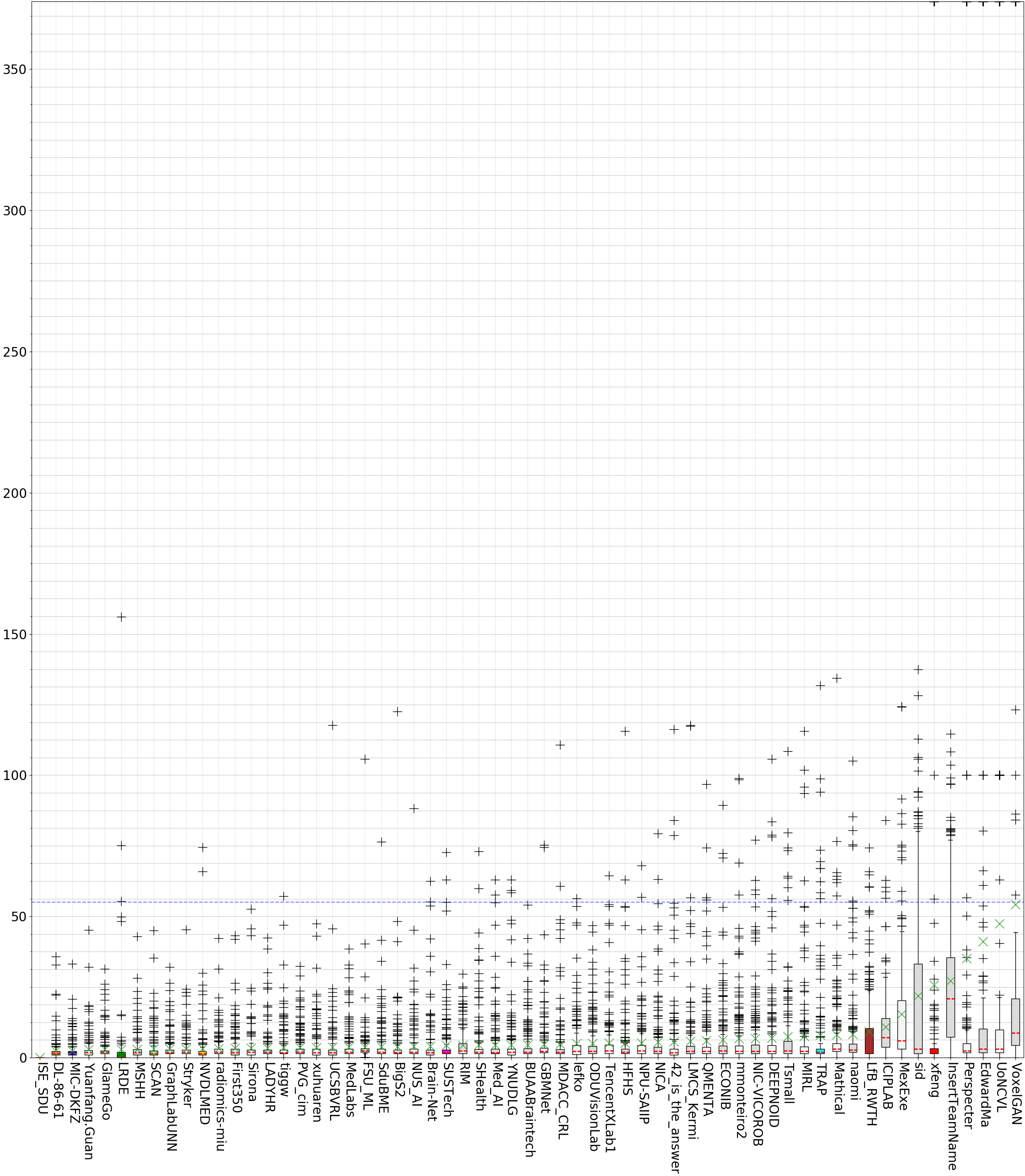}
            \caption{BraTS 2018 summarizing results (Hausdorff) for the segmentation of the active tumor compartment.}
            \label{fig:sup:hausdorffET}
        \end{centering}
    \end{figure}

\begin{figure}
        \begin{centering}
            \includegraphics[width=1\columnwidth]{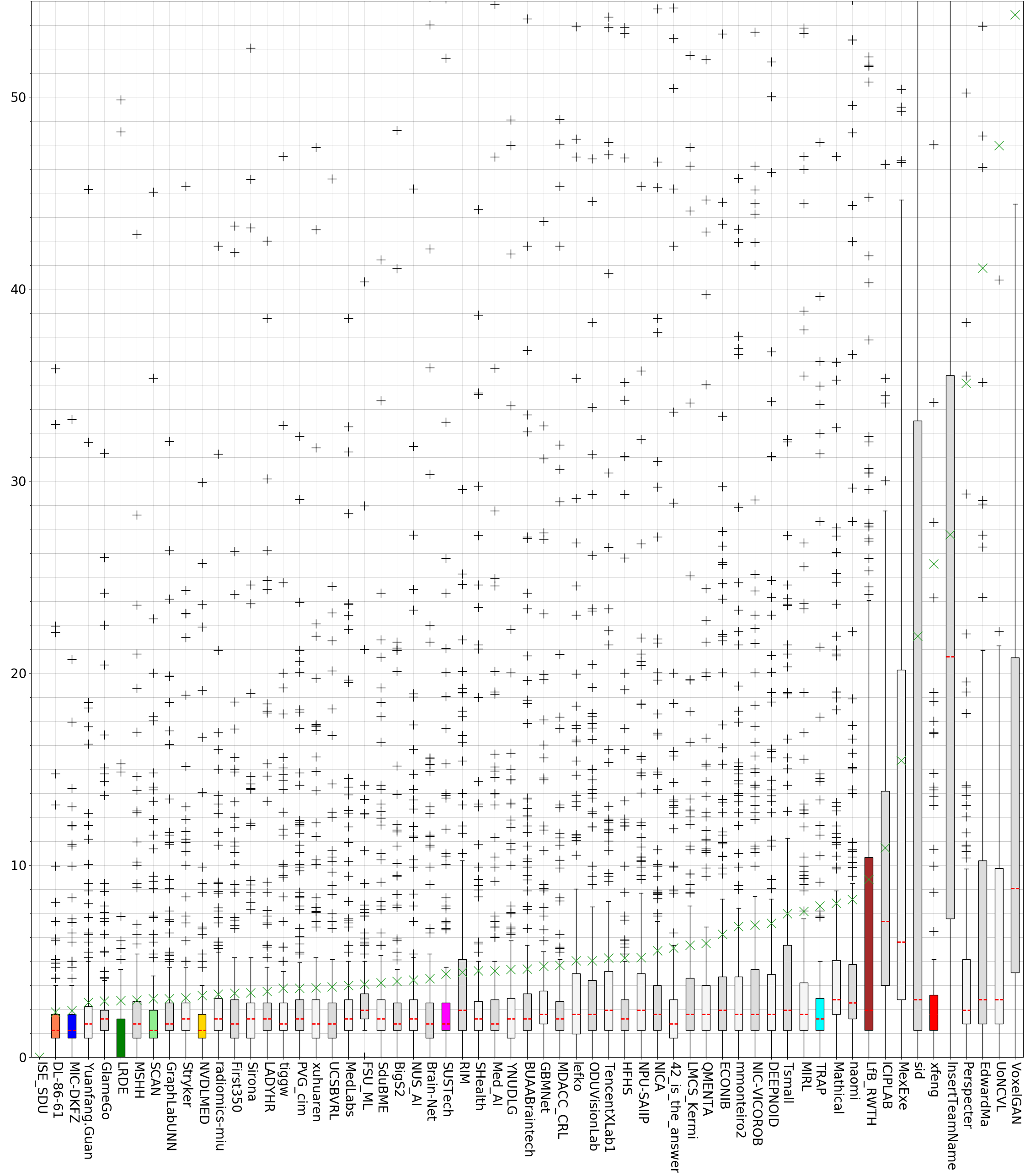}
            \caption{BraTS 2018 summarizing results (Hausdorff) for the segmentation of the active tumor compartment, with cutoff values for visualization purposes.}
            \label{fig:sup:hausdorffETCutOff}
        \end{centering}
    \end{figure}

\begin{figure}
        \begin{centering}
            \includegraphics[width=1\columnwidth]{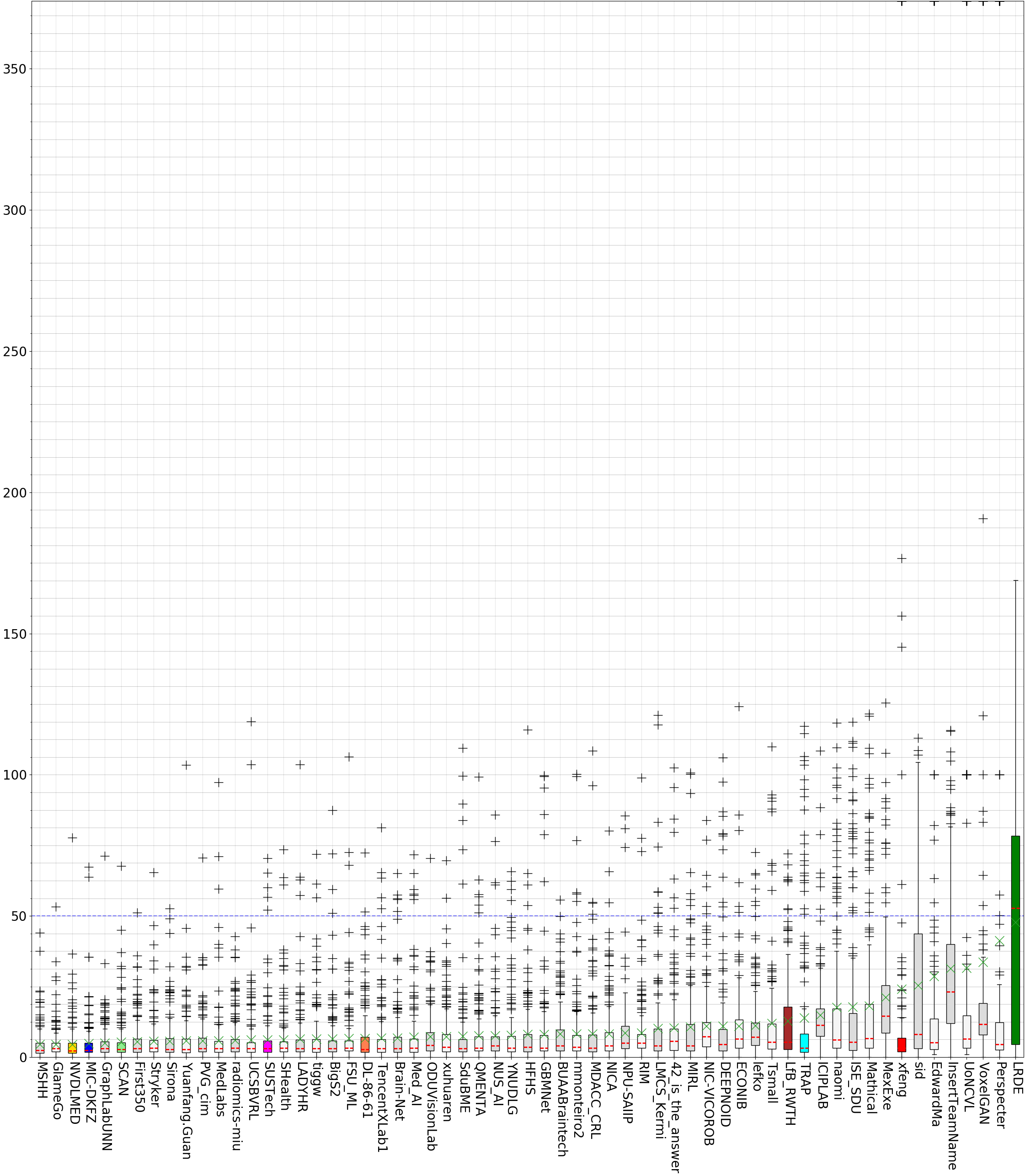}
            \caption{BraTS 2018 summarizing results (Hausdorff) for the segmentation of the tumor core compartment.}
            \label{fig:sup:hausdorffTC}
        \end{centering}
    \end{figure}

\begin{figure}
        \begin{centering}
            \includegraphics[width=1\columnwidth]{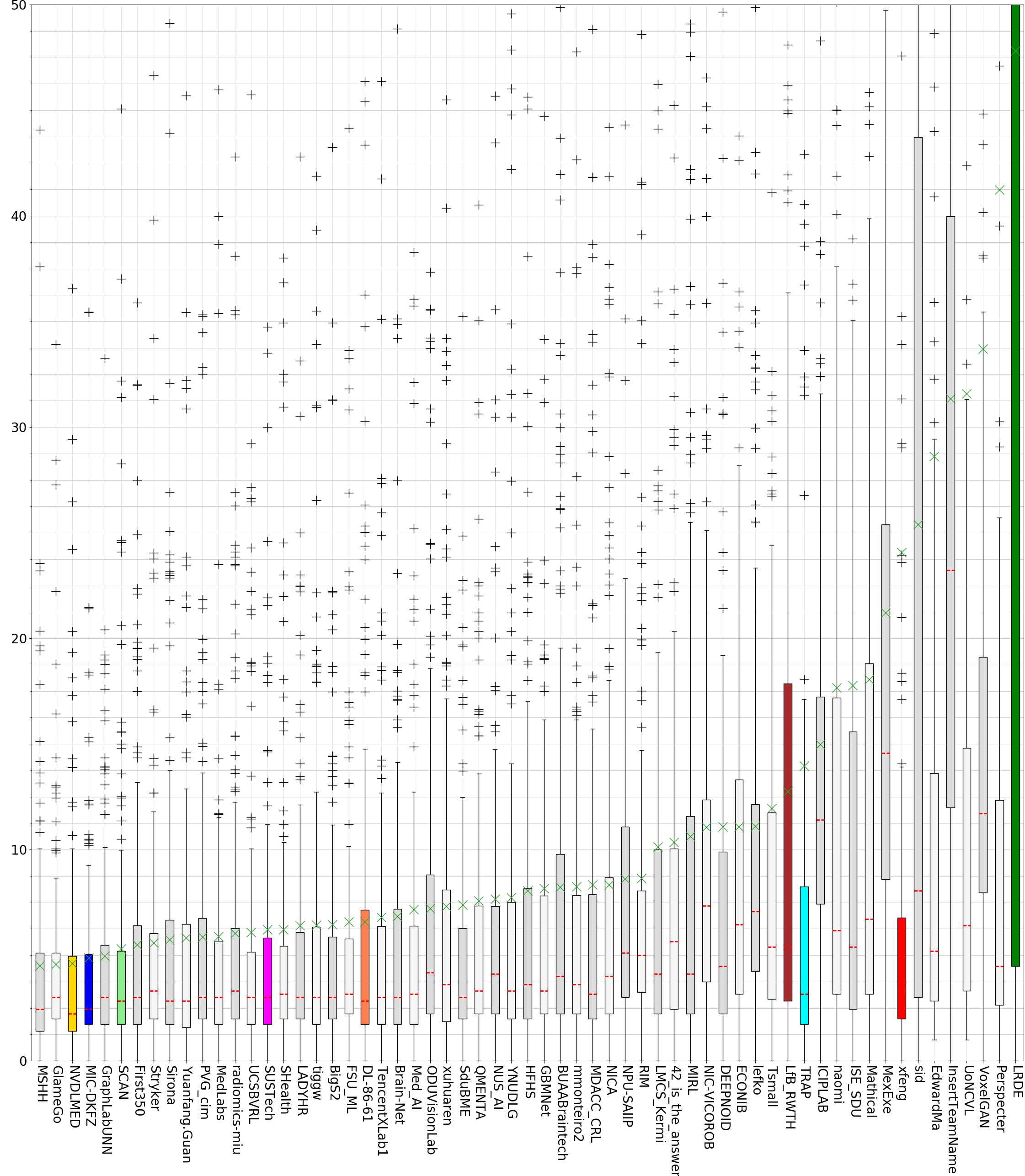}
            \caption{BraTS 2018 summarizing results (Hausdorff) for the segmentation of the tumor core compartment, with cutoff values for visualization purposes.}
            \label{fig:sup:hausdorffTCCutOff}
        \end{centering}
    \end{figure}

\begin{figure}
        \begin{centering}
            \includegraphics[width=1\columnwidth]{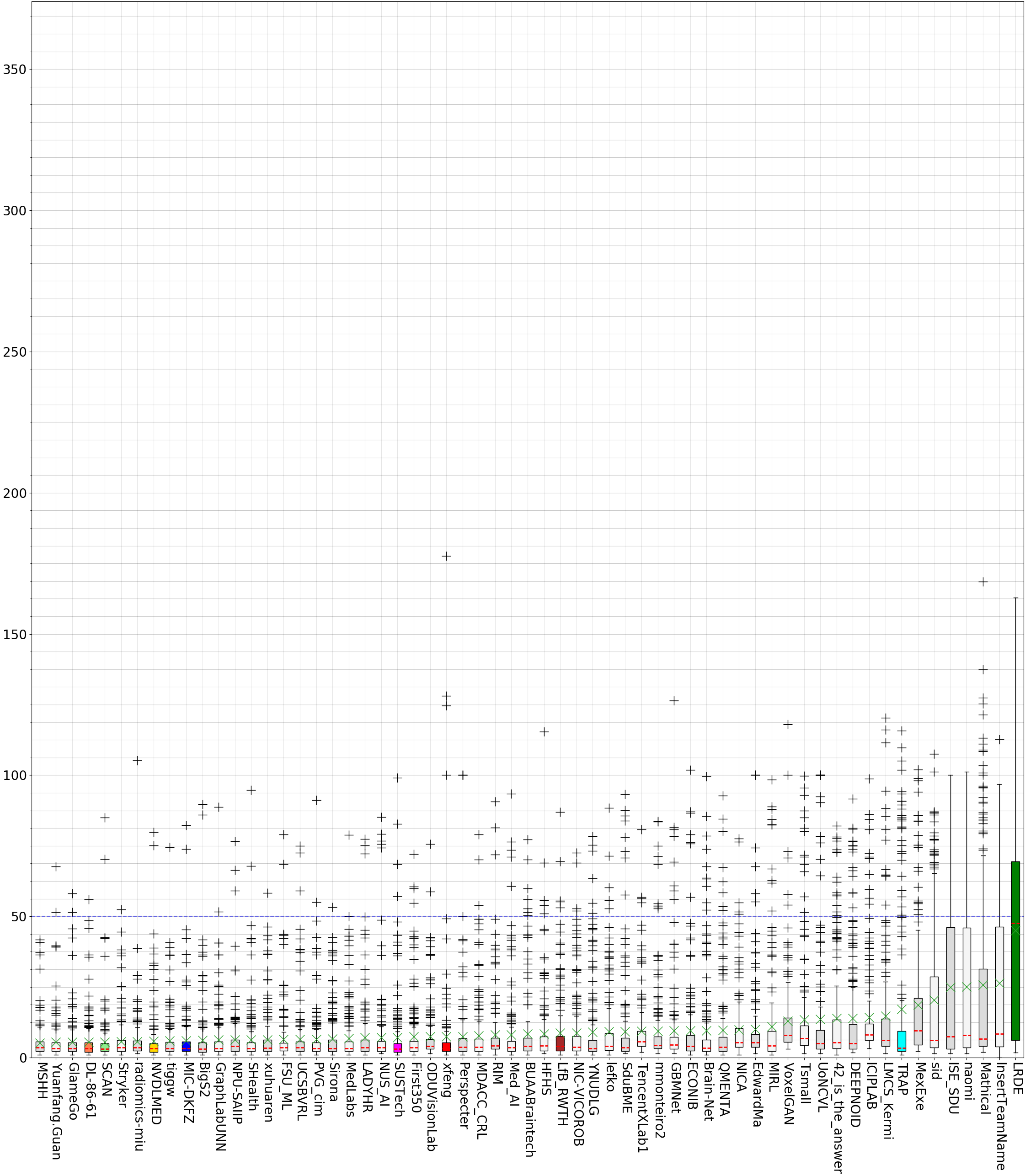}
            \caption{BraTS 2018 summarizing results (Hausdorff) for the segmentation of the whole tumor compartment.}
            \label{fig:sup:hausdorffWT}
        \end{centering}
    \end{figure}

\begin{figure}
        \begin{centering}
            \includegraphics[width=1\columnwidth]{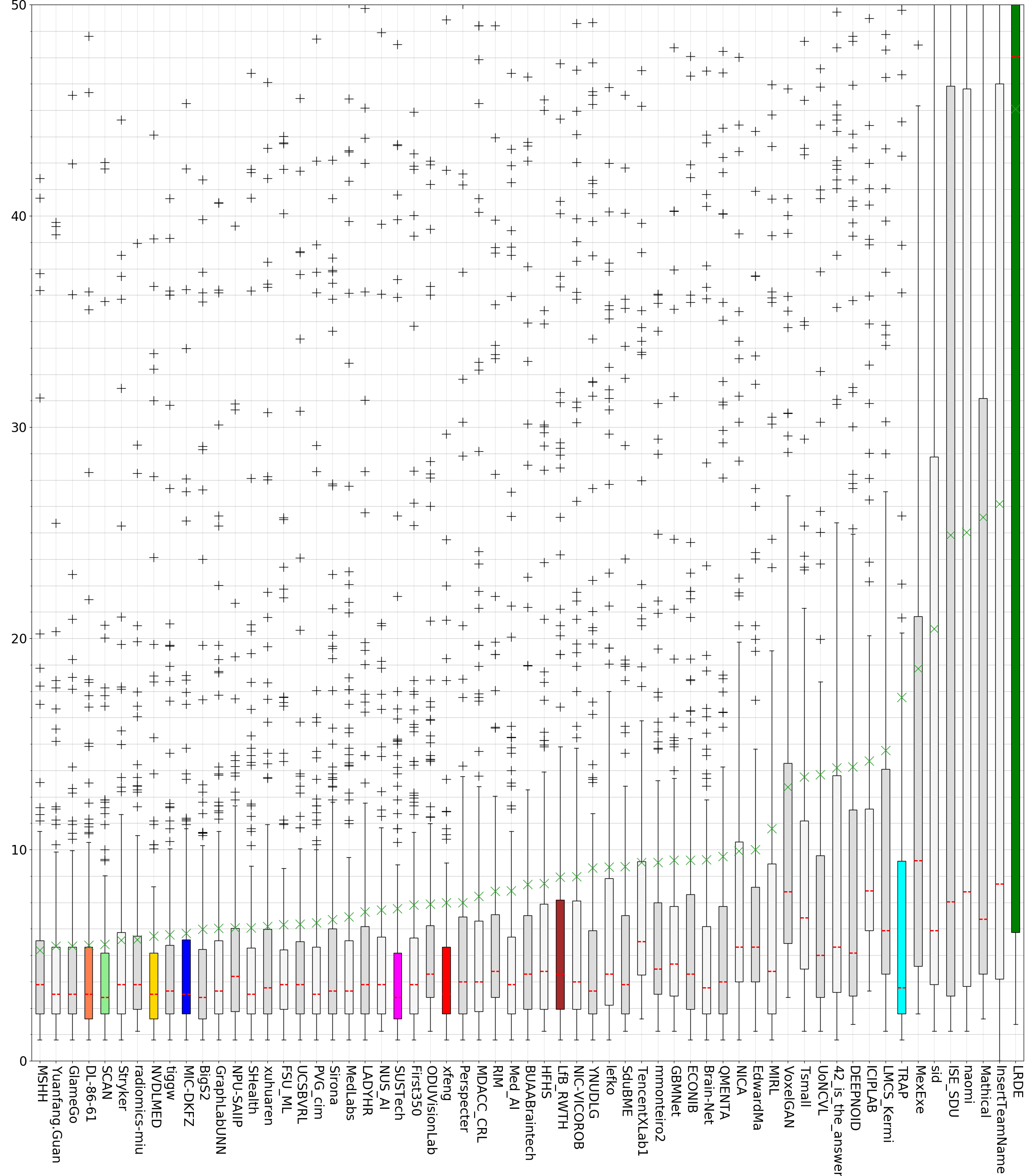}
            \caption{BraTS 2018 summarizing results (Hausdorff) for the segmentation of the whole tumor compartment, with cutoff values for visualization purposes.}
            \label{fig:sup:hausdorffWTCutOff}
        \end{centering}
    \end{figure}

                The 95\% Hausdorff distance metric is used to characterize the levels of robustness of the automated results. Supplementary Figures \ref{fig:sup:hausdorffET} through \ref{fig:sup:hausdorffWTCutOff} show the Hausdorff metric values for the three evaluated tumor compartments for all teams. Overall, the results for the AT seems to be the most robust for all three tumor labels (median IQR of 1.9 for the same 54/63 teams), followed by the results for the WT and that for the TC (IQR of 4.0 and 5.4 for the same 54/63 teams, respectively).

\begin{figure}[t!]
                    \begin{centering}
                        \includegraphics[width=1\columnwidth]{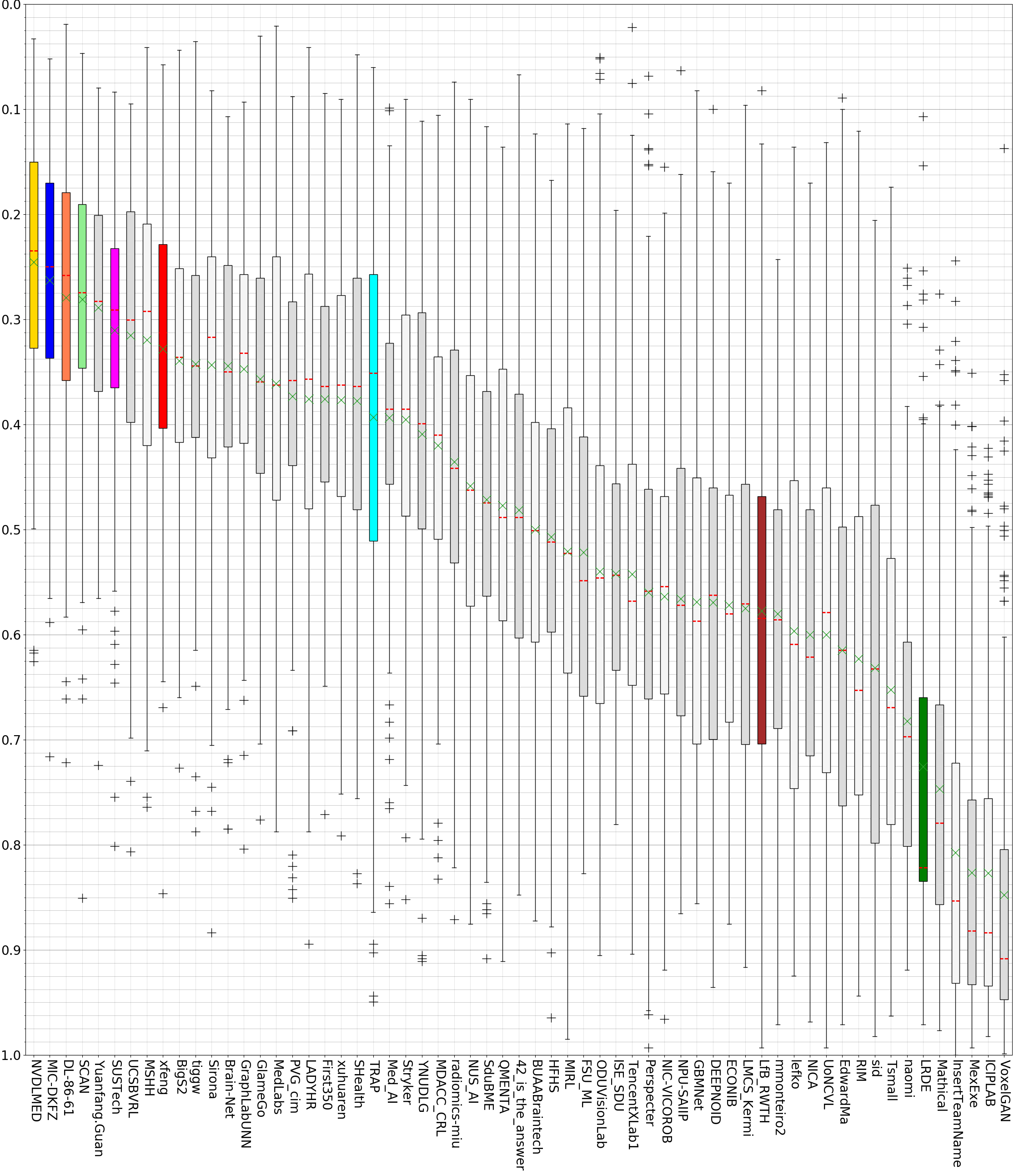}
                        \caption{BraTS 2018 Ranking of all Participating Teams in Segmentation Task. (smaller values are higher ranks)}
                        \label{fig:B18_2018_rankings}
                    \end{centering}
            	\end{figure}

                At the patient-wise ranking of the participating teams (Fig.~\ref{fig:B18_2018_rankings}) the distribution follows more closely a gradual improvement of the ranked approaches, similar to results from BraTS 2017. Worth noting is that the variability of the ranking of approaches at the case level does not dramatically change across teams, indicating no particular dominance of a method over the others. We also performed a pairwise comparison for significant differences based on 100,000 permutations that showed the statistically significant performance across teams. Specifically, the statistical evaluation of the top-ranked teams revealed that the first team was statistically better from the second (p-value=0.02), whereas the second team was not statistically better than the third (p=0.06) and the fourth (p=0.07), but only from the fifth (p=0.01). This justified the decision of a tie in the third rank.

\begin{figure}[t!]
                    \begin{centering}
                        \includegraphics[width=1\columnwidth]{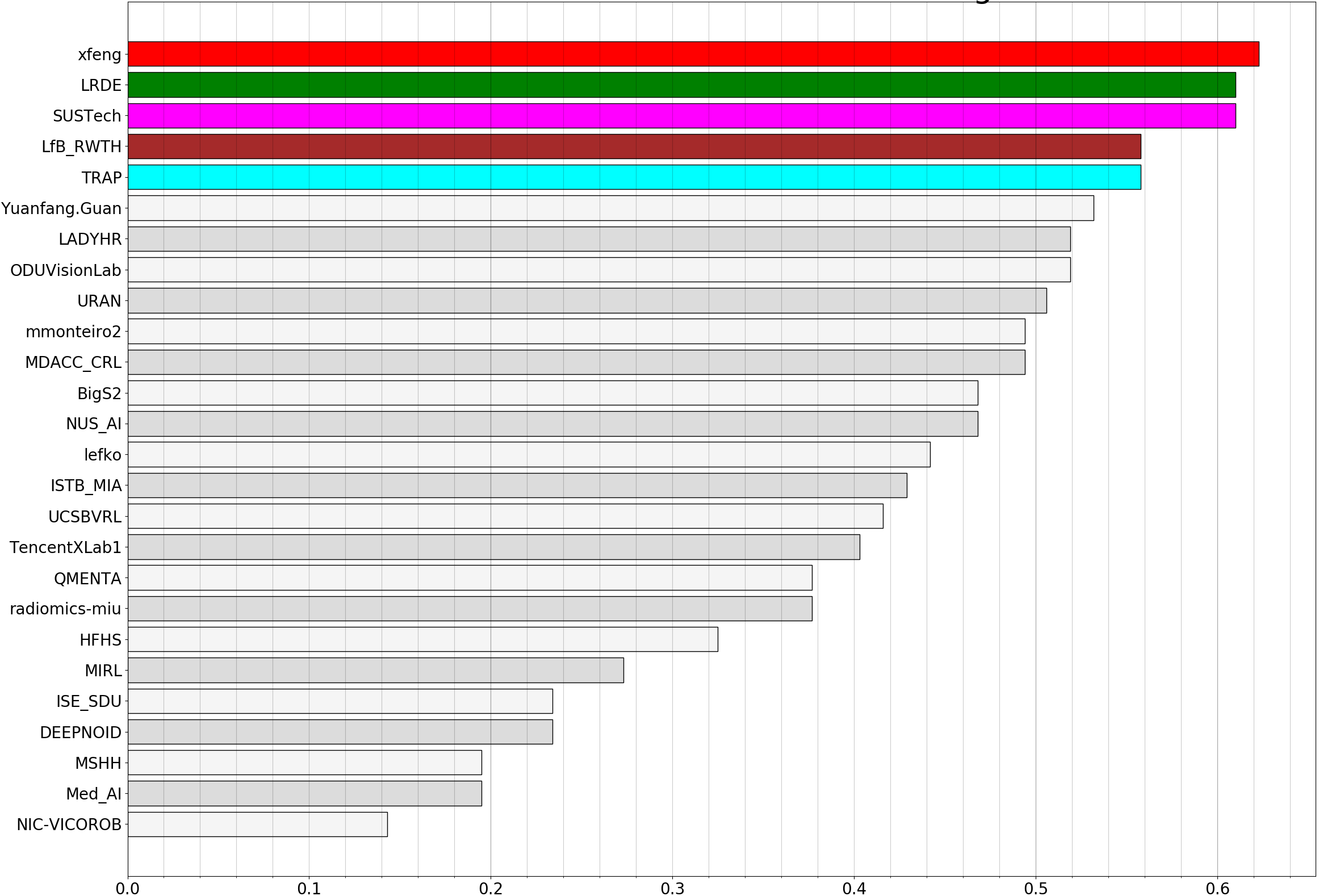}
                        \caption{BraTS 2018 Ranking of all Participating Teams in Survival Task. (larger values are better)}
                        \label{fig:B18_2018_rankings_survival}
                    \end{centering}
            	\end{figure}

                Results of the survival task are shown in Fig.~\ref{fig:B18_2018_rankings_survival}. Overall, the top-5 approaches obtained an accuracy around  0.6, while the rest of teams obtained an accuracy in the range of [0.15-0.55]. We should clarify that the random choice should be considered the 0.33 since this is a 3-class classification.

                The final top-performing participating teams positioned in ranks 1-3 are shown in Table \ref{tbl:2018_winners}.

                \begin{table}[!t]
                    \centering
                    \caption{Top-ranked participating teams in BraTS 2018 for both the segmentation and the survival prediction tasks.}
                    \begin{tabular}{|c|c|c|c|p{7cm}|c|}
                        \hline
                        \textbf{Task} & \textbf{Rank} & \textbf{Team} & \textbf{First Author} & \textbf{Institution} & \textbf{Paper}\\
                        \hline
                        \hline
                        \multirow{4}{*}{\raisebox{-1.1\height}{\rotatebox{90}{Segmentation}}} & 1 & NVDLMED & Andriy Myronenko & NVIDIA, Santa Clara, USA & \cite{brats18:rank1:seg:nvidia}\\
                        \cline{2-6}
                        {} & 2 & MIC-DKFZ & Fabian Isensee & Division of Medical Image Computing, German Cancer Research Center (DKFZ), Heidelberg, Germany & \cite{brats18:rank2:seg:fabian}\\
                        \cline{2-6}
                        {} & 3 (tie) & SCAN & Richard McKinley & Support Centre for Advanced Neuroimaging Inselspital, Bern University Hospital, Switzerland & \cite{brats18:rank3:seg:mckinley}\\
                        \cline{2-6}
                        {} & 3 (tie) & DL\_86\_81 & Chenhong Zhou & School of Electronic \& Information Engineering, South China University of Technology, China & \cite{brats18:rank3:seg:chenhong}\\
                        \hline
                        \hline
                        \multirow{4}{*}{\raisebox{-1.8\height}{\rotatebox{90}{Survival}}} & 1 & xfeng & Xue Feng & Biomedical Engineering, University of Virginia, USA & \cite{brats18:rank1:surv:feng}\\
                        \cline{2-6}
                        {} & 2 (tie) & LRDE & \'Elodie Puybareau & EPITA Research and Development Laboratory, France & \cite{brats18:rank2:surv:elodie}\\
                        \cline{2-6}
                        {} & 2 (tie) & SUSTech & Li Sun & Southern University of Science \& Technology, China & \cite{brats18:rank2:surv:lisun}\\
        	   		\cline{2-6}
        	   		{} & 3 (tie) & TRAP & Ujjwal Baid & Shri Guru Gobind Singhji Institute of Engineering and Technology, India & \cite{brats18:rank3:surv:tata}\\
        	   		\cline{2-6}
        	   		{} & 3 (tie) & LfB\_RWTH & Leon Weninger & Institute of Imaging \& Computer Vision, RWTH Aachen University, Germany & \cite{brats18:rank3:surv:leon}\\
    	   		\hline
    	   	\end{tabular}
    	   	\label{tbl:2018_winners}
    	   \end{table}

    \section{Discussion}
    \label{section:conclusion}
        \subsection{Performance of Automated Segmentation Methods}
            While the accuracy of individual automated segmentation methods has improved, we note that their level of robustness is still inferior to expert performance, i.e., inter-rater agreement. This robustness is expected to be continuously improving as the training set increases in size, in virtue of capturing and describing more diverse patient populations, along with improved training schemes and ML architectures. Beyond these speculative expectations, the results of our quantitative analyses support that the fusion of segmentation labels from various individual automated methods shows robustness superior to the ground truth inter-rater agreement (provided by clinical experts), in terms of both accuracy and consistency across subjects. However, proposed strategies to ensemble several models correspond to one practical way to reduce outliers and improve the precision of automated segmentation systems, by means of consensus segmentation across different models. We consider future research essential, in order to improve the robustness of individual approaches by increasing the ability of segmentation systems to handle confounding effects typically seen in images acquired using routine clinical workflows. Related to BraTS such effects include, but are not limited to, a) the presence of blood products, b) "air-pockets"/resection cavities in post-operative scans, c) better differentiation (or handling) of non-GBM entities, and d) improved performance for low-grade gliomas, featuring diffuse boundaries, especially while considering cases without AT sub-regions, and e) high sensitivity to effectively detect and assess their slow progression.

        \subsection{BraTS Ranking Schema}
            The BraTS challenge recently adopted a case-wise ranking schema, which enables a more clinically-relevant evaluation of participating teams, as it considers the complexity of patient cases that can vary significantly. Furthermore, the additional featured evaluation of statistical significance of differences across algorithmic results, also enables the evaluation of results across different instances of the BraTS challenge, which in turns enables a thorough analysis of the improvement attained over the last seven years of the BraTS initiative.

        \subsection{Beyond Segmentation}
            Importantly, two more clinically-relevant tasks/sub-challenges have been complementarily added in the BraTS initiative during these past seven years, aiming at emphasizing the clinical relevance of the brain tumor segmentation task. Both these clinically-relevant tasks promote the natural utilization of segmentation labels to answer clinical questions, address clinical requirements, and potentially support  the clinical decision-making process. The ultimate goal of these additions was to evaluate the potential usability and pave the way of automated segmentation methods towards their translation to routine clinical practice.

            \subsubsection{Assessment of Disease Progression}
                The inclusion of longitudinal (i.e., follow up) mpMRI scans took place during the BraTS 2014-2016 instances. In clinical practice, assessment of disease progression is to date performed through the Response Evaluation Criteria In Solid Tumours (RECIST) \cite{recist1,recist2,recist3,recist4} and the Response Assessment in Neuro-Oncology (RANO) criteria \cite{rano}, whose quantitative component is based on the relative change of tumor size (i.e., percentile changes) measured by the longest two axes of the assessed tumor. In this regard, we postulate that automated algorithms performing brain tumor volumetric segmentation (i.e., in three dimensions) should yield reliable comparable (if not better) estimates of volumetric tumor changes.

            \subsubsection{Prediction of Overall Survival}
                The inclusion of the OS prediction task took place during the BraTS 2017-2018 instances and has highlighted (or rather confirmed) the difficulties of Deep Learning (DL) approaches to handle small training sets, and the superiority of traditional ML approaches. While this finding clearly calls for larger training sets, it also identifies the need for potential synergies between DL and traditional ML approaches as we transition to larger training sets in the future, which can include more non-uniformly distributed clinical and/or molecular information. In other words, there is a need to develop advanced ML approaches able to handle the large existing heterogeneity of the patient-specific information available in the clinics, e.g., radiogenomics \cite{rgRutmanEjrLink,rgEllingsonGbm,rgGutmanJnrTcga,rgItakuraStmSubtypes,rgBakasCcrPhi,rgJayashreeCcrIdh,rgAkbariNeuroOncologyEgfrv3,rgBinderCancercellA289}, RIS reports.

        \subsection{Future Directions for the BraTS Initiative}
            The current trend over the years of the previous BraTS instances highlights (or rather confirms) a) the superiority of DL over traditional ML approaches in the segmentation task (and particularly in terms of Dice), and, in contrast, b) the struggle of DL and the superiority of traditional ML approaches, assessing more clinically-relevant problems, such as the prediction of clinical outcome (i.e., overall survival), where smaller training sets are typically available and need to be handled.

            Concentrating on the segmentation task, in terms of algorithmic design, the current general consensus seems to point in the direction of tackling the problem in a hierarchical/cascaded way, by first distinguishing between normal and abnormal/tumorous tissue, and then proceeding with the segmentation of the tumor sub-regions. Alternative research directions include the enhancement of the flexibility of DL systems that might lack a given set of input images \cite{havaeiBengio}, as a transition measure towards worldwide adoption of the standardization initiatives for GBM imaging \cite{elllingsonCoalitionTrials}.

            There are many clinical endpoints where the BraTS initiative can have a potential impact and these include, but not limited to: a) training systems for neuroradiology trainees,  b) differential diagnosis (e.g., metastases differentiation, disease progression assessment, radio-phenotyping), c) prognosis (e.g., prediction of overall survival, drug-response prediction), d) radiation therapy planning. However, for any of these to be potentially considered wider application of the developed methods needs to take place, which is why we created the BraTS algorithmic repository, and a closer collaboration with the clinical expertise is fundamental to tailor the design of the BraTS challenges towards an effective exploitation and translation of research findings into clinical practice.
            	
    \section{Acknowledgements}
    \label{section:conclusion}
        Importantly, we would like to express our gratitude to all the data contributing institutions that assisted in putting together the publicly-available multi-institutional mpMRI BraTS dataset, acquired with different clinical protocols and various scanners. Note that without these contributions the BraTS initiative would have never been feasible. These data contributors are: 1) Center for Biomedical Image Computing and Analytics (CBICA), University of Pennsylvania (UPenn), PA, USA, 2) University of Alabama at Birmingham, AL, USA, 3) Heidelberg University, Germany, 4) University Hospital of Bern, Switzerland, 5) University of Debrecen, Hungary, 6) Henry Ford Hospital, MI, USA, 7) University of California, CA, USA, 8) MD Anderson Cancer Center, TX, USA, 9) Emory University, GA, USA, 10) Mayo Clinic, MN, USA, 11) Thomas Jefferson University, PA, USA, 12) Duke University School of Medicine, NC, USA, 13) Saint Joseph Hospital and Medical Center, AZ, USA, 14) Case Western Reserve University, OH, USA, 15) University of North Carolina, NC, USA, 16) Fondazione IRCCS Instituto Neuroligico C. Besta, Italy, 17) MD Anderson Cancer Center, TX, USA, 18) Washington University School of Medicine in St. Louis, MO, USA, and 19) Tata Memorial Center, Mumbai, India. Note that data from institutions 6-16 are provided through The Cancer Imaging Archive (TCIA - http://www.cancerimagingarchive.net/), supported by the Cancer Imaging Program (CIP) of the National Cancer Institute (NCI) of the National Institutes of Health (NIH).

        We would also like to thank the sponsorship offered by the CBICA@UPenn for the plaques provided to the top-ranked participating teams of the challenge each year, as well as Intel AI for sponsoring the monetary prizes of total value of \$5,000, awarded to the three top-ranked participating teams of the BraTS 2018 challenge, who also shared publicly their containerized algorithm in the BraTS algorithmic repository: \url{github.com/BraTS/Instructions/blob/master/Repository_Links.md} \&  \url{hub.docker.com/u/brats/}.

        This work was supported in part by the 1) National Institute of Neurological Disorders and Stroke (NINDS) of the NIH R01 grant with award number R01-NS042645, 2) Informatics Technology for Cancer Research (ITCR) program of the NCI/NIH U24 grant with award number U24-CA189523, 3) Swiss Cancer League, under award number KFS-3979-08-2016, 4) Swiss National Science Foundation, under award number 169607. The content of this publication is solely the responsibility of the authors and does not necessarily represent the official views of NIH or any of the other funding bodies.

    \bibliographystyle{splncs}

\begin{thebibliography}{100}

\bibitem{bratsTmiPaper}
Menze, B.H., Jakab, A., Bauer, S., Kalpathy-Cramer, J., Farahani, K., Kirby,
  J., et~al.:
\newblock The multimodal brain tumor image segmentation benchmark (brats).
\newblock IEEE Transactions on Medical Imaging \textbf{34} (2015)  1993--2024

\bibitem{NatSciDataPaper}
Bakas, S., Akbari, H., Sotiras, A., Bilello, M., Rozycki, M., Kirby, J.S.,
  et~al.:
\newblock Advancing the cancer genome atlas glioma mri collections with expert
  segmentation labels and radiomic features.
\newblock Nature Scientific Data \textbf{4} (2017)  170117

\bibitem{TciaGbmPaper}
Bakas, S., Akbari, H., Sotiras, A., Bilello, M., Rozycki, M., Kirby, J.S.,
  et~al.:
\newblock Segmentation labels and radiomic features for the pre-operative scans
  of the tcga-gbm collection.
\newblock T. C. I. Archive (2017)

\bibitem{TciaLggPaper}
Bakas, S., Akbari, H., Sotiras, A., Bilello, M., Rozycki, M., Kirby, J.S.,
  et~al.:
\newblock Segmentation labels and radiomic features for the pre-operative scans
  of the tcga-lgg collection.
\newblock T. C. I. Archive (2015)

\bibitem{IbsiPaper}
Zwanenburg, A., Leger, S., Vallières, M., Löck, S., I.B.S.I:
\newblock Image biomarker standardisation initiative.
\newblock arXiv preprint arXiv:1612.07003 (2016)

\bibitem{sriPaper}
Rohlfing, T., Zahr, N.M., Sullivan, E.V., Pfefferbaum, A.:
\newblock The sri24 multi-channel atlas of normal adult human brain structure.
\newblock Human brain mapping \textbf{31} (2010)  798--819

\bibitem{Haller:pathologyAgeRelated}
Haller, S., Kovari, E., Herrmann, F.R., Cuvinciuc, V., Tomm, A.M., Zulian,
  G.B., Lovblad, K.O., Giannakopoulos, P., Bouras, C.:
\newblock Do brain t2/flair white matter hyperintensities correspond to myelin
  loss in normal aging. a radiologic-neuropathologic correlation study.
\newblock Acta Neuropathol Commun \textbf{1}(14) (2013)  1--7

\bibitem{TciaPaper}
Clark, K., Vendt, B., Smith, K., Freymann, J., Kirby, J., Koppel, P., et~al.:
\newblock The cancer imaging archive (tcia): Maintaining and operating a public
  information repository.
\newblock Journal of Digital Imaging \textbf{26} (2013)  1045--1057

\bibitem{TciaTcgaGbmPaper}
Scarpace, L., Mikkelsen, T., Cha, S., Rao, S., Tekchandani, S., Gutman, D.,
  et~al.:
\newblock Radiology data from the cancer genome atlas glioblastoma multiforme
  [tcga-gbm] collection.
\newblock The Cancer Imaging Archive (2016)

\bibitem{TciaTcgaLggPaper}
Pedano, N., Flanders, A.E., Scarpace, L., Mikkelsen, T., Eschbacher, J.M.,
  Hermes, B., et~al.:
\newblock Radiology data from the cancer genome atlas low grade glioma
  [tcga-lgg] collection.
\newblock The Cancer Imaging Archive (2016)

\bibitem{IslesPaper2015}
Maier, O., Menze, B.H., von~der Gablentz, J., Häni, L., Heinrich, M.P.,
  Liebrand, M., et~al.:
\newblock Isles 2015 - a public evaluation benchmark for ischemic stroke lesion
  segmentation from multispectral mri.
\newblock Medical Image Analysis \textbf{35} (2017)  250--269

\bibitem{IslesPaper2016}
Winzeck, S., Hakim, A., McKinley, R., Pinto, J., Alves, V., Silva, C., et~al.:
\newblock Isles 2016 and 2017-benchmarking ischemic stroke lesion outcome
  prediction based on multispectral mri.
\newblock Frontiers in neurology \textbf{9} (2018)  679--679

\bibitem{stuppLancet}
Stupp, R., Hegi, M.E., Mason, W.P., van~den Bent, M.J., Taphoorn, M.J.B.,
  Janzer, R.C., et~al.:
\newblock Effects of radiotherapy with concomitant and adjuvant temozolomide
  versus radiotherapy alone on survival in glioblastoma in a randomised phase
  iii study: 5-year analysis of the eortc-ncic trial.
\newblock The Lancet Oncology \textbf{10} (2009)  459--466

\bibitem{tmzTrial}
Gilbert, M.R., Wang, M., Aldape, K.D., Stupp, R., Hegi, M.E., Jaeckle, K.A.,
  et~al.:
\newblock Dose-dense temozolomide for newly diagnosed glioblastoma: A
  randomized phase iii clinical trial.
\newblock Journal of Clinical Oncology \textbf{31} (2013)  4085--4091

\bibitem{brats17:lncs:MIRL}
Alex, V., Safwan, M., Krishnamurthi, G.:
\newblock Automatic segmentation and overall survival prediction in gliomas
  using fully convolutional neural network and texture analysis.
\newblock BrainLes 2017, Springer LNCS \textbf{10670} (2018)  216--225

\bibitem{brats17:prepro:whatapain}
Amorim, P.H.A., Chagas, V.S., Escudero, G., Oliveira, D.D.C., Pereira, S.M.,
  Santos, H.M., Scussel, A.A.:
\newblock 3d u-nets for brain tumor segmentation in miccai 2017 brats
  challenge.
\newblock MICCAI BraTS 2017 Pre-proceedings -
  \url{https://www.cbica.upenn.edu/sbia/Spyridon.Bakas/MICCAI_BraTS/MICCAI_BraTS_2017_proceedings_shortPapers.pdf}
  (2017)  9--14

\bibitem{brats17:prepro:cian}
Andermatt, S., Pezold, S., Cattin, P.:
\newblock Multi-dimensional gated recurrent units for brain tumor segmentation.
\newblock MICCAI BraTS 2017 Pre-proceedings -
  \url{https://www.cbica.upenn.edu/sbia/Spyridon.Bakas/MICCAI_BraTS/MICCAI_BraTS_2017_proceedings_shortPapers.pdf}
  (2017)  15--19

\bibitem{brats17:prepro:qtim}
Beers, A., Chang, K., Brown, J., Sartor, E., Mammen, C., Gerstner, E., Rosen,
  B., Kalpathy-Cramer, J.:
\newblock Sequential 3d u-nets for brain tumor segmentation.
\newblock MICCAI BraTS 2017 Pre-proceedings -
  \url{https://www.cbica.upenn.edu/sbia/Spyridon.Bakas/MICCAI_BraTS/MICCAI_BraTS_2017_proceedings_shortPapers.pdf}
  (2017)  20--23

\bibitem{brats17:lncs:biomedia1BHN}
Bharath, H.N., Colleman, S., Sima, D., Huffel, S.V.:
\newblock Tumor segmentation from multimodal mri using random forest with
  superpixel and tensor based feature extraction.
\newblock BrainLes 2017, Springer LNCS \textbf{10670} (2018)  463--473

\bibitem{brats17:prepro:drcubic}
Cao, S., Qian, B., Yin, C., Li, X., Chang, S.:
\newblock 3d u-net for multimodal brain tumor segmentation.
\newblock MICCAI BraTS 2017 Pre-proceedings -
  \url{https://www.cbica.upenn.edu/sbia/Spyridon.Bakas/MICCAI_BraTS/MICCAI_BraTS_2017_proceedings_shortPapers.pdf}
  (2017)  30--33

\bibitem{brats17:lncs:UPCDLMI}
Casamitjana, A., Cat\'a, M., S\'anchez, I., Combalia, M., Vilaplana, V.:
\newblock Cascaded v-net using roi masks for brain tumor segmentation.
\newblock BrainLes 2017, Springer LNCS \textbf{10670} (2018)  381--391

\bibitem{brats17:lncs:BCVUniandes}
Castillo, L.S., Daza, L.A., Rivera, L.C., Arbel\'aez, P.:
\newblock Brain tumor segmentation and parsing on mris using multiresolution
  neural networks.
\newblock BrainLes 2017, Springer LNCS \textbf{10670} (2018)  332--343

\bibitem{brats17:prepro:SCUTEE}
Chen, S., Ding, C., Zhou, C.:
\newblock Brain tumor segmentation with label distribution learning and
  multi-level feature representation.
\newblock MICCAI BraTS 2017 Pre-proceedings -
  \url{https://www.cbica.upenn.edu/sbia/Spyridon.Bakas/MICCAI_BraTS/MICCAI_BraTS_2017_proceedings_shortPapers.pdf}
  (2017)  50--53

\bibitem{brats17:lncs:CNEA}
Colmeiro, R.G.R., Verrastro, C.A., Grosges, T.:
\newblock Multimodal brain tumor segmentation using 3d convolutional networks.
\newblock BrainLes 2017, Springer LNCS \textbf{10670} (2018)  226--240

\bibitem{brats17:prepro:CISA}
Dong, S.:
\newblock A separate 3d-segnet architecture for brain tumor segmentation.
\newblock MICCAI BraTS 2017 Pre-proceedings -
  \url{https://www.cbica.upenn.edu/sbia/Spyridon.Bakas/MICCAI_BraTS/MICCAI_BraTS_2017_proceedings_shortPapers.pdf}
  (2017)  54--60

\bibitem{brats17:prepro:niftynet}
Eaton-Rosen, Z., Li, W., Wang, G., Vercauteren, T., Bisdas, S., Ourselin, S.,
  Cardoso, M.J.:
\newblock Using niftynet to ensemble convolutional neural nets for the brats
  challenge.
\newblock MICCAI BraTS 2017 Pre-proceedings -
  \url{https://www.cbica.upenn.edu/sbia/Spyridon.Bakas/MICCAI_BraTS/MICCAI_BraTS_2017_proceedings_shortPapers.pdf}
  (2017)  61--66

\bibitem{brats17:prepro:xfeng}
Feng, X., Meyer, C.:
\newblock Patch-based 3d u-net for brain tumor segmentation.
\newblock MICCAI BraTS 2017 Pre-proceedings -
  \url{https://www.cbica.upenn.edu/sbia/Spyridon.Bakas/MICCAI_BraTS/MICCAI_BraTS_2017_proceedings_shortPapers.pdf}
  (2017)  67--72

\bibitem{brats17:lncs:NPU}
Hu, Y., Xia, Y.:
\newblock 3d deep neural network-based brain tumor segmentation using
  multimodality magnetic resonance sequences.
\newblock BrainLes 2017, Springer LNCS \textbf{10670} (2018)  423--434

\bibitem{brats17:lncs:MICDKFZ}
Isensee, F., Kickingereder, P., Wick, W., Bendszus, M., Maier-Hein, K.H.:
\newblock Brain tumor segmentation and radiomics survival prediction:
  Contribution to the brats 2017 challenge.
\newblock BrainLes 2017, Springer LNCS \textbf{10670} (2018)  287--297

\bibitem{brats17:lncs:NUS_MPR}
Islam, M., Ren, H.:
\newblock Multi-modal pixelnet for brain tumor segmentation.
\newblock BrainLes 2017, Springer LNCS \textbf{10670} (2018)  298--308

\bibitem{brats17:lncs:pvg}
Jesson, A., Arbel, T.:
\newblock Brain tumor segmentation using a 3d fcn with multi-scale loss.
\newblock BrainLes 2017, Springer LNCS \textbf{10670} (2018)  392--402

\bibitem{brats17:lncs:ISTB}
Jungo, A., McKinley, R., Meier, R., Knecht, U., Vera, L., P\'erez-Beteta, J.,
  Molina-Garc\'ia, D., P\'erez-Garc\'ia, V.M., Wiest, R., Reyes, M.:
\newblock Towards uncertainty-assisted brain tumor segmentation and survival
  prediction.
\newblock BrainLes 2017, Springer LNCS \textbf{10670} (2018)  474--485

\bibitem{brats17:lncs:biomedia1}
Kamnitsas, K., Bai, W., Ferrante, E., McDonagh, S., Sinclair, M., Pawlowski,
  N., Rajchl, M., Lee, M.C.H., Kainz, B., Rueckert, D., Glocker, B.:
\newblock Ensembles of multiple models and architectures for robust brain
  tumour segmentation.
\newblock BrainLes 2017, Springer LNCS \textbf{10670} (2018)  450--462

\bibitem{brats17:prepro:BrICLab}
Karnawat, A., Prasanna, P., Madabushi, A., Tiwari, P.:
\newblock Radiomics-based convolutional neural network (radcnn) for brain tumor
  segmentation on multi-parametric mri.
\newblock MICCAI BraTS 2017 Pre-proceedings -
  \url{https://www.cbica.upenn.edu/sbia/Spyridon.Bakas/MICCAI_BraTS/MICCAI_BraTS_2017_proceedings_shortPapers.pdf}
  (2017)  147--153

\bibitem{brats17:lncs:Rocky}
Kim, G.:
\newblock Brain tumor segmentation using deep fully convolutional neural
  networks.
\newblock BrainLes 2017, Springer LNCS \textbf{10670} (2018)  344--357

\bibitem{brats17:lncs:neuro.ml}
Krivov, E., Pisov, M., Belyaev, M.:
\newblock Mri augmentation via elastic registration for brain lesions
  segmentation.
\newblock BrainLes 2017, Springer LNCS \textbf{10670} (2018)  369--380

\bibitem{brats17:lncs:OnePiece}
Li, Y., Shen, L.:
\newblock Deep learning based multimodal brain tumor diagnosis.
\newblock BrainLes 2017, Springer LNCS \textbf{10670} (2018)  149--158

\bibitem{brats17:prepro:ZejuLi:FDUBME}
Li, Z., Wang, Y., Yu, J.:
\newblock Brain tumor segmentation using an adversarial network.
\newblock MICCAI BraTS 2017 Pre-proceedings -
  \url{https://www.cbica.upenn.edu/sbia/Spyridon.Bakas/MICCAI_BraTS/MICCAI_BraTS_2017_proceedings_shortPapers.pdf}
  (2017)  164--168

\bibitem{brats17:prepro:Mountain}
Li, X., Zhang, X., Luo, Z.:
\newblock Brain tumor segmentation via 3d fully dilated convolutional networks.
\newblock MICCAI BraTS 2017 Pre-proceedings -
  \url{https://www.cbica.upenn.edu/sbia/Spyridon.Bakas/MICCAI_BraTS/MICCAI_BraTS_2017_proceedings_shortPapers.pdf}
  (2017)  175--179

\bibitem{brats17:prepro:SJTU}
Liu, L., Nie, D., Wang, Q., Shen, D.:
\newblock A location sensitive brain tumor segmentation method.
\newblock MICCAI BraTS 2017 Pre-proceedings -
  \url{https://www.cbica.upenn.edu/sbia/Spyridon.Bakas/MICCAI_BraTS/MICCAI_BraTS_2017_proceedings_shortPapers.pdf}
  (2017)  180--187

\bibitem{brats17:lncs:UCCS}
Lopez, M.M., Ventura, J.:
\newblock Dilated convolutions for brain tumor segmentation in mri scans.
\newblock BrainLes 2017, Springer LNCS \textbf{10670} (2018)  253--262

\bibitem{brats17:prepro:PADAS}
Mang, A., Tharakan, S., Gholami, A., Himthani, N., Subramanian, S., Levitt, J.,
  Azmat, M., Scheufele, K., Mehl, M., Davatzikos, C., Barth, B., Biros, G.:
\newblock Sibia-gls: Scalable biophysics-based image analysis for glioma
  segmentation.
\newblock MICCAI BraTS 2017 Pre-proceedings -
  \url{https://www.cbica.upenn.edu/sbia/Spyridon.Bakas/MICCAI_BraTS/MICCAI_BraTS_2017_proceedings_shortPapers.pdf}
  (2017)  197--204

\bibitem{brats17:lncs:Bern}
McKinley, R., Jungo, A., Wiest, R., Reyes, M.:
\newblock Pooling-free fully convolutional networks with dense skip connections
  for semantic segmentation, with application to brain tumor segmentation.
\newblock BrainLes 2017, Springer LNCS \textbf{10670} (2018)  169--177

\bibitem{brats17:lncs:Alexander}
Osman, A.F.I.:
\newblock Automated brain tumor segmentation on magnetic resonance images and
  patient's overall survival prediction using support vector machines.
\newblock BrainLes 2017, Springer LNCS \textbf{10670} (2018)  435--449

\bibitem{brats17:lncs:MBI}
Pawar, K., Chen, Z., Shah, N.J., Egan, G.:
\newblock Residual encoder and convolutional decoder neural network for glioma
  segmentation.
\newblock BrainLes 2017, Springer LNCS \textbf{10670} (2018)  263--273

\bibitem{brats17:lncs:Ashi}
Phophalia, A., Maji, P.:
\newblock Multimodal brain tumor segmentation using ensemble of forest method.
\newblock BrainLes 2017, Springer LNCS \textbf{10670} (2018)  159--168

\bibitem{brats17:lncs:ROB}
Pourreza, R., Zhuge, Y., Ning, H., Miller, R.:
\newblock Brain tumor segmentation in mri scans using deeply-supervised neural
  networks.
\newblock BrainLes 2017, Springer LNCS \textbf{10670} (2018)  320--331

\bibitem{brats17:prepro:justdoit}
Revanuru, K., Shah, N.:
\newblock Fully automatic brain tumour segmentation using random forests and
  patient survival prediction using xgboost.
\newblock MICCAI BraTS 2017 Pre-proceedings -
  \url{https://www.cbica.upenn.edu/sbia/Spyridon.Bakas/MICCAI_BraTS/MICCAI_BraTS_2017_proceedings_shortPapers.pdf}
  (2017)  239--243

\bibitem{brats17:lncs:HPI-Ultimate}
Rezaei, M., Harmuth, K., Gierke, W., Kellermeier, T., Fischer, M., Yang, H.,
  Meinel, C.:
\newblock A conditional adversarial network for semantic segmentation of brain
  tumor.
\newblock BrainLes 2017, Springer LNCS \textbf{10670} (2018)  241--252

\bibitem{brats17:lncs:SaraS}
Sedlar, S.:
\newblock Brain tumor segmentation using a multi-path cnn based method.
\newblock BrainLes 2017, Springer LNCS \textbf{10670} (2018)  403--422

\bibitem{brats17:lncs:BRATZZ27}
Shaikh, M., Anand, G., Acharya, G., Amrutkar, A., Alex, V., Krishnamurthi, G.:
\newblock Brain tumor segmentation using dense fully convolutional neural
  network.
\newblock BrainLes 2017, Springer LNCS \textbf{10670} (2018)  309--319

\bibitem{brats17:lncs:VisionLab}
Shboul, Z.A., Vidyaratne, L., Alam, M., Iftekharuddin, K.M.:
\newblock Glioblastoma and survival prediction.
\newblock BrainLes 2017, Springer LNCS \textbf{10670} (2018)  358--368

\bibitem{brats17:prepro:Dundee}
Shen, H., Wang, R., Zhang, J., McKenna, S.:
\newblock Symmetry-driven fully convolutional network for brain tumor
  segmentation.
\newblock MICCAI BraTS 2017 Pre-proceedings -
  \url{https://www.cbica.upenn.edu/sbia/Spyridon.Bakas/MICCAI_BraTS/MICCAI_BraTS_2017_proceedings_shortPapers.pdf}
  (2017)  274--278

\bibitem{brats17:lncs:LoVE}
Soltaninejad, M., Zhang, L., Lambrou, T., Yang, G., Allinson, N., Ye, X.:
\newblock Mri brain tumor segmentation and patient survival prediction using
  random forests and fully convolutional networks.
\newblock BrainLes 2017, Springer LNCS \textbf{10670} (2018)  204--215

\bibitem{brats17:lncs:UCL-TIG}
Wang, G., Li, W., Ourselin, S., Vercauteren, T.:
\newblock Automatic brain tumor segmentation using cascaded anisotropic
  convolutional neural networks.
\newblock BrainLes 2017, Springer LNCS \textbf{10670} (2018)  178--190

\bibitem{brats17:prepro:STH}
Wang, C., Smedby, O.:
\newblock Automatic brain tumor segmentation using 2.5d u-nets.
\newblock MICCAI BraTS 2017 Pre-proceedings -
  \url{https://www.cbica.upenn.edu/sbia/Spyridon.Bakas/MICCAI_BraTS/MICCAI_BraTS_2017_proceedings_shortPapers.pdf}
  (2017)  292--296

\bibitem{brats17:prepro:CMR}
Yang, T.L., Ou, Y.N., Huang, T.Y.:
\newblock Automatic segmentation of brain tumor from mr images using segnet:
  selection of training data sets.
\newblock MICCAI BraTS 2017 Pre-proceedings -
  \url{https://www.cbica.upenn.edu/sbia/Spyridon.Bakas/MICCAI_BraTS/MICCAI_BraTS_2017_proceedings_shortPapers.pdf}
  (2017)  309--312

\bibitem{brats17:lncs:Zhao}
Zhao, X., Wu, Y., Song, G., Li, Z., Zhang, Y., Fan, Y.:
\newblock 3d brain tumor segmentation through integrating multiple 2d fcnns.
\newblock BrainLes 2017, Springer LNCS \textbf{10670} (2018)  191--203

\bibitem{brats17:prepro:SegLZ}
Zhao, L.:
\newblock Automatic brain tumor segmentation with 3d deconvolution network with
  dilated inception block.
\newblock MICCAI BraTS 2017 Pre-proceedings -
  \url{https://www.cbica.upenn.edu/sbia/Spyridon.Bakas/MICCAI_BraTS/MICCAI_BraTS_2017_proceedings_shortPapers.pdf}
  (2017)  316--320

\bibitem{brats17:lncs:BIGS2}
Zhou, F., Li, T., Li, H., Zhu, H.:
\newblock Tpcnn: Two-phase patch-based convolutional neural network for
  automatic brain tumor segmentation and survival prediction.
\newblock BrainLes 2017, Springer LNCS \textbf{10670} (2018)  274--286

\bibitem{brats17:prepro:Zhouch}
Zhou, C., Ding, C., Lu, Z., Zhang, T.:
\newblock Brain tumor segmentation with cascaded convolutional neural networks.
\newblock MICCAI BraTS 2017 Pre-proceedings -
  \url{https://www.cbica.upenn.edu/sbia/Spyridon.Bakas/MICCAI_BraTS/MICCAI_BraTS_2017_proceedings_shortPapers.pdf}
  (2017)  328--333

\bibitem{brats17:prepro:CamMIA}
Zhu, J., Wang, D., Teng, Z., Li\'o, P.:
\newblock A multi-pathway 3d dilated convolutional neural network for brain
  tumor segmentation.
\newblock MICCAI BraTS 2017 Pre-proceedings -
  \url{https://www.cbica.upenn.edu/sbia/Spyridon.Bakas/MICCAI_BraTS/MICCAI_BraTS_2017_proceedings_shortPapers.pdf}
  (2017)  342--347

\bibitem{brats18:1}
Albiol, A., Albiol, A., Albiol, F.:
\newblock Extending 2d deep learning architectures to 3d image segmentation
  problems.
\newblock BrainLes 2018, Springer LNCS \textbf{11384} (2019)  73--82

\bibitem{brats18:rank3:surv:tata}
Baid, U., Talbar, S., Rane, S., Gupta, S., Thakur, M.H., Moiyadi, A., Thakur,
  S., Mahajan, A.:
\newblock Deep learning radiomics algorithm for gliomas (drag) model: A novel
  approach using 3d unet based deep convolutional neural network for predicting
  survival in gliomas.
\newblock BrainLes 2018, Springer LNCS \textbf{11384} (2019)  369--379

\bibitem{brats18:2}
Banerjee, S., Mitra, S., Shankar, B.U.:
\newblock Multi-planar spatial-convnet for segmentation and survival prediction
  in brain cancer.
\newblock BrainLes 2018, Springer LNCS \textbf{11384} (2019)  94--104

\bibitem{brats18:3}
Benson, E., Pound, M.P., French, A.P., Jackson, A.S., Pridmore, T.P.:
\newblock Deep hourglass for brain tumor segmentation.
\newblock BrainLes 2018, Springer LNCS \textbf{11384} (2019)  419--428

\bibitem{brats18:prepro:cabezas}
Cabezas, M., Valverde, S., Gonz\'alez-Vill\'a, S., C\'eérigues, A., Salem, M.,
  Kushibar, K., Bernal, J., Oliver, A., Salvi, J., Llad\'o, X.:
\newblock Survival prediction using ensemble tumor segmentation and transfer
  learning.
\newblock MICCAI BraTS 2018 Pre-proceedings -
  \url{https://www.cbica.upenn.edu/sbia/Spyridon.Bakas/MICCAI_BraTS/MICCAI_BraTS_2018_proceedings_shortPapers.pdf}
  (2018)  54--62

\bibitem{brats18:4}
Carver, E., Liu, C., Zong, W., Dai, Z., Snyder, J.M., Lee, J., Wen, N.:
\newblock Automatic brain tumor segmentation and overall survival prediction
  using machine learning algorithms.
\newblock BrainLes 2018, Springer LNCS \textbf{11384} (2019)  406--418

\bibitem{brats18paragios}
Chandra, S., Vakalopoulou, M., Fidon, L., Battistella, E., Estienne, T., Sun,
  R., Robert, C., Deutsch, E., Paragios, N.:
\newblock Context aware 3d cnns for brain tumor segmentation.
\newblock BrainLes 2018, Springer LNCS \textbf{11384} (2019)  299--310

\bibitem{brats18:prepro:chang}
Chang, Y.J., Lin, Z.S., Yang, T.L., Huang, T.Y.:
\newblock Automatic segmentation of brain tumor from 3d mr images using a 2d
  convolutional neural network.
\newblock MICCAI BraTS 2018 Pre-proceedings -
  \url{https://www.cbica.upenn.edu/sbia/Spyridon.Bakas/MICCAI_BraTS/MICCAI_BraTS_2018_proceedings_shortPapers.pdf}
  (2018)  83--90

\bibitem{brats18:5}
Chen, W., Liu, B., Peng, S., Sun, J., Qiao, X.:
\newblock S3d-unet: Separable 3d u-net for brain tumor segmentation.
\newblock BrainLes 2018, Springer LNCS \textbf{11384} (2019)  358--368

\bibitem{brats18:6}
Choudhury, A.R., Vanguri, R., Jambawalikar, S.R., Kumar, P.:
\newblock Segmentation of brain tumors using deeplabv3+.
\newblock BrainLes 2018, Springer LNCS \textbf{11384} (2019)  154--167

\bibitem{brats18:7}
Dai, L., Li, T., Shu, H., Zhong, L., Shen, H., Zhu, H.:
\newblock Automatic brain tumor segmentation with domain adaptation.
\newblock BrainLes 2018, Springer LNCS \textbf{11384} (2019)  380--392

\bibitem{brats18:prepro:fang}
Fang, L., He, H.:
\newblock Three pathways u-net for brain tumor segmentation.
\newblock MICCAI BraTS 2018 Pre-proceedings -
  \url{https://www.cbica.upenn.edu/sbia/Spyridon.Bakas/MICCAI_BraTS/MICCAI_BraTS_2018_proceedings_shortPapers.pdf}
  (2018)  119--126

\bibitem{brats18:rank1:surv:feng}
Feng, X., Tustison, N., Meyer, C.:
\newblock Brain tumor segmentation using an ensemble of 3d u-nets and overall
  survival prediction using radiomic features.
\newblock BrainLes 2018, Springer LNCS \textbf{11384} (2019)  279--288

\bibitem{brats18:prepro:fridman}
Fridman, N.:
\newblock Brain tumor detection and segmentation using deep learning u-net on
  multi modal mri.
\newblock MICCAI BraTS 2018 Pre-proceedings -
  \url{https://www.cbica.upenn.edu/sbia/Spyridon.Bakas/MICCAI_BraTS/MICCAI_BraTS_2018_proceedings_shortPapers.pdf}
  (2018)  135--143

\bibitem{brats18:8}
Gates, E., Pauloski, J.G., Schellingerhout, D., Fuentes, D.:
\newblock Glioma segmentation and a simple accurate model for overall survival
  prediction.
\newblock BrainLes 2018, Springer LNCS \textbf{11384} (2019)  476--484

\bibitem{brats18:9}
Gering, D., Sun, K., Avery, A., Chylla, R., Vivekanandan, A., Kohli, L., Knapp,
  H., Paschke, B., Young-Moxon, B., King, N., Mackie, T.:
\newblock Semi-automatic brain tumor segmentation by drawing long axes on
  multi-plane reformat.
\newblock BrainLes 2018, Springer LNCS \textbf{11384} (2019)  441--455

\bibitem{brats18biros}
Gholami, A., Subramanian, S., Shenoy, V., Himthani, N., Yue, X., Zhao, S., Jin,
  P., Biros, G., Keutzer, K.:
\newblock A novel domain adaptation framework for medical image segmentation.
\newblock BrainLes 2018, Springer LNCS \textbf{11384} (2019)  289--298

\bibitem{brats18:10}
Han, W.S., Han, I.S.:
\newblock Neuromorphic neural network for multimodal brain image segmentation
  and overall survival analysis.
\newblock BrainLes 2018, Springer LNCS \textbf{11384} (2019)  178--188

\bibitem{brats18:prepro:huKong}
Hu, X., Huang, W., Kong, D., Guo, S., Scott, M.R.:
\newblock Brainnet: 3d local refinement network for brain tumor segmentation.
\newblock MICCAI BraTS 2018 Pre-proceedings -
  \url{https://www.cbica.upenn.edu/sbia/Spyridon.Bakas/MICCAI_BraTS/MICCAI_BraTS_2018_proceedings_shortPapers.pdf}
  (2018)  179--187

\bibitem{brats18:12:bjoern}
Hu, X., Li, H., Zhao, Y., Dong, C., Menze, B.H., Piraud, M.:
\newblock Hierarchical multi-class segmentation of glioma images using networks
  with multi-level activation function.
\newblock BrainLes 2018, Springer LNCS \textbf{11384} (2019)  116--127

\bibitem{brats18:11}
Hu, Y., Liu, X., Wen, X., Niu, C., Xia, Y.:
\newblock Brain tumor segmentation on multimodal mr imaging using multi-level
  upsampling in decoder.
\newblock BrainLes 2018, Springer LNCS \textbf{11384} (2019)  168--177

\bibitem{brats18:13}
Hua, R., Huo, Q., Gao, Y., Sun, Y., Shi, F.:
\newblock Multimodal brain tumor segmentation using cascaded v-nets.
\newblock BrainLes 2018, Springer LNCS \textbf{11384} (2019)  49--60

\bibitem{brats18:prepro:vivek}
HV, V.:
\newblock Pre and post processing techniques for brain tumor segmentation.
\newblock MICCAI BraTS 2018 Pre-proceedings -
  \url{https://www.cbica.upenn.edu/sbia/Spyridon.Bakas/MICCAI_BraTS/MICCAI_BraTS_2018_proceedings_shortPapers.pdf}
  (2018)  213--221

\bibitem{brats18:rank2:seg:fabian}
Isensee, F., Kickingereder, P., Wick, W., Bendszus, M., Maier-Hein, K.H.:
\newblock No new-net.
\newblock BrainLes 2018, Springer LNCS \textbf{11384} (2019)  234--244

\bibitem{brats18:14}
Islam, M., Jose, V.J.M., Ren, H.:
\newblock Glioma prognosis: Segmentation of the tumor and survival prediction
  using shape, geometric and clinical information.
\newblock BrainLes 2018, Springer LNCS \textbf{11384} (2019)  142--153

\bibitem{brats18:15}
Kao, P.Y., Ngo, T., Zhang, A., Chen, J.W., Manjunath, B.S.:
\newblock Brain tumor segmentation and tractographic feature extraction from
  structural mr images for overall survival prediction.
\newblock BrainLes 2018, Springer LNCS \textbf{11384} (2019)  128--141

\bibitem{brats18:16}
Kermi, A., Mahmoudi, I., Khadir, M.T.:
\newblock Deep convolutional neural networks using u-net for automatic brain
  tumor segmentation in multimodal mri volumes.
\newblock BrainLes 2018, Springer LNCS \textbf{11384} (2019)  37--48

\bibitem{brats18:17}
Kori, A., Soni, M., Pranjal, B., Khened, M., Alex, V., Krishnamurthi, G.:
\newblock Ensemble of fully convolutional neural network for brain tumor
  segmentation from magnetic resonance images.
\newblock BrainLes 2018, Springer LNCS \textbf{11384} (2019)  485--496

\bibitem{brats18:18}
Lachinov, D., Vasiliev, E., Turlapov, V.:
\newblock Glioma segmentation with cascaded unet.
\newblock BrainLes 2018, Springer LNCS \textbf{11384} (2019)  189--198

\bibitem{brats18:19}
Lefkovits, S., Szil\'agyi, L., Lefkovits, L.:
\newblock Brain tumor segmentation and survival prediction using a cascade of
  random forests.
\newblock BrainLes 2018, Springer LNCS \textbf{11384} (2019)  334--345

\bibitem{brats18:prepro:XLi}
Li, X.:
\newblock Fused u-net for brain tumor segmentation based on multimodal mr
  images.
\newblock MICCAI BraTS 2018 Pre-proceedings -
  \url{https://www.cbica.upenn.edu/sbia/Spyridon.Bakas/MICCAI_BraTS/MICCAI_BraTS_2018_proceedings_shortPapers.pdf}
  (2018)  290--297

\bibitem{brats18:prepro:Liu}
Liu, M.:
\newblock Coarse-to-fine deep convolutional neural networks for multi-modality
  brain tumor semantic segmentation.
\newblock MICCAI BraTS 2018 Pre-proceedings -
  \url{https://www.cbica.upenn.edu/sbia/Spyridon.Bakas/MICCAI_BraTS/MICCAI_BraTS_2018_proceedings_shortPapers.pdf}
  (2018)  298--305

\bibitem{brats18:20}
Ma, J., Yang, X.:
\newblock Automatic brain tumor segmentation by exploring the multi-modality
  complementary information and cascaded 3d lightweight cnns.
\newblock BrainLes 2018, Springer LNCS \textbf{11384} (2019)  25--36

\bibitem{brats18:21}
Marcinkiewicz, M., Nalepa, J., Lorenzo, P.R., Dudzik, W., Mrukwa, G.:
\newblock Segmenting brain tumors from mri using cascaded multi-modal u-nets.
\newblock BrainLes 2018, Springer LNCS \textbf{11384} (2019)  13--24

\bibitem{brats18:rank3:seg:mckinley}
McKinley, R., Meier, R., Wiest, R.:
\newblock Ensembles of densely-connected cnns with label-uncertainty for brain
  tumor segmentation.
\newblock BrainLes 2018, Springer LNCS \textbf{11384} (2019)  456--465

\bibitem{brats18:22}
Mehta, R., Arbel, T.:
\newblock 3d u-net for brain tumour segmentation.
\newblock BrainLes 2018, Springer LNCS \textbf{11384} (2019)  254--266

\bibitem{brats18:prepro:Monteiro}
Monteiro, M., Oliveira, A.L.:
\newblock Ensemble of fully convolutional neural networks for brain tumour
  semantic segmentation.
\newblock MICCAI BraTS 2018 Pre-proceedings -
  \url{https://www.cbica.upenn.edu/sbia/Spyridon.Bakas/MICCAI_BraTS/MICCAI_BraTS_2018_proceedings_shortPapers.pdf}
  (2018)  341--348

\bibitem{brats18:rank1:seg:nvidia}
Myronenko, A.:
\newblock 3d mri brain tumor segmentation using autoencoder regularization.
\newblock BrainLes 2018, Springer LNCS \textbf{11384} (2019)  311--320

\bibitem{brats18:23}
Nuechterlein, N., Mehta, S.:
\newblock 3d-espnet with pyramidal refinement for volumetric brain tumor image
  segmentation.
\newblock BrainLes 2018, Springer LNCS \textbf{11384} (2019)  245--253

\bibitem{brats18:prepro:Popli}
Popli, A., Agarwal, M., Pillai, G.:
\newblock Automatic brain tumor segmentation using u-net based 3d fully
  convolutional network.
\newblock MICCAI BraTS 2018 Pre-proceedings -
  \url{https://www.cbica.upenn.edu/sbia/Spyridon.Bakas/MICCAI_BraTS/MICCAI_BraTS_2018_proceedings_shortPapers.pdf}
  (2018)  374--382

\bibitem{brats18:24}
Puch, S., S\'anchez, I., Hern\'andez, A., Piella, G., Pr\'ckovska, V.:
\newblock Global planar convolutions for improved context aggregation in brain
  tumor segmentation.
\newblock BrainLes 2018, Springer LNCS \textbf{11384} (2019)  393--405

\bibitem{brats18:rank2:surv:elodie}
Puybareau, E., Tochon, G., Chazalon, J., Fabrizio, J.:
\newblock Segmentation of gliomas and prediction of patient overall survival: A
  simple and fast procedure.
\newblock BrainLes 2018, Springer LNCS \textbf{11384} (2019)  199--209

\bibitem{brats18:prepro:Ren}
Ren, X., Zhang, L., Shen, D., Wang, Q.:
\newblock Ensembles of multiple scales, losses and models for brain tumor
  segmentation and overall survival time prediction task.
\newblock MICCAI BraTS 2018 Pre-proceedings -
  \url{https://www.cbica.upenn.edu/sbia/Spyridon.Bakas/MICCAI_BraTS/MICCAI_BraTS_2018_proceedings_shortPapers.pdf}
  (2018)  402--410

\bibitem{brats18:25}
Rezaei, M., Yang, H., Meinel, C.:
\newblock voxel-gan: Adversarial framework for learning imbalanced brain tumor
  segmentation.
\newblock BrainLes 2018, Springer LNCS \textbf{11384} (2019)  321--333

\bibitem{brats18:26}
Serrano-Rubio, J.P., Everson, R.:
\newblock Brain tumour segmentation method based on supervoxels and sparse
  dictionaries.
\newblock BrainLes 2018, Springer LNCS \textbf{11384} (2019)  210--221

\bibitem{brats18:27:ODU}
Shboul, Z.A., Alam, M., Vidyaratne, L., Pei, L., Iftekharuddin, K.M.:
\newblock Glioblastoma survival prediction.
\newblock BrainLes 2018, Springer LNCS \textbf{11384} (2019)  508--515

\bibitem{brats18:prepro:Deepnoid}
Shin, H.E., Park, M.S.:
\newblock Brain tumor segmentation using 2d u-net.
\newblock MICCAI BraTS 2018 Pre-proceedings -
  \url{https://www.cbica.upenn.edu/sbia/Spyridon.Bakas/MICCAI_BraTS/MICCAI_BraTS_2018_proceedings_shortPapers.pdf}
  (2018)  428--437

\bibitem{brats18:28:stryker}
Stawiaski, J.:
\newblock A pretrained densenet encoder for brain tumor segmentation.
\newblock BrainLes 2018, Springer LNCS \textbf{11384} (2019)  105--115

\bibitem{brats18:rank2:surv:lisun}
Sun, L., Zhang, S., Luo, L.:
\newblock Tumor segmentation and survival prediction in glioma with deep
  learning.
\newblock BrainLes 2018, Springer LNCS \textbf{11384} (2019)  83--93

\bibitem{brats18:29:mauricio}
Suter, Y., Jungo, A., Rebsamen, M., Knecht, U., Herrmann, E., Wiest, R., Reyes,
  M.:
\newblock Deep learning versus classical regression for brain tumor patient
  survival prediction.
\newblock BrainLes 2018, Springer LNCS \textbf{11384} (2019)  429--440

\bibitem{brats18:prepro:Tseng}
Tseng, K.L., Hsu, W.:
\newblock End-to-end cascade network for 3d brain tumor segmentation in miccai
  2018 brats challenge.
\newblock MICCAI BraTS 2018 Pre-proceedings -
  \url{https://www.cbica.upenn.edu/sbia/Spyridon.Bakas/MICCAI_BraTS/MICCAI_BraTS_2018_proceedings_shortPapers.pdf}
  (2018)  466--473

\bibitem{brats18:30}
Tuan, T.A., Tuan, T.A., Bao, P.T.:
\newblock Brain tumor segmentation using bit-plane and unet.
\newblock BrainLes 2018, Springer LNCS \textbf{11384} (2019)  466--475

\bibitem{brats18:31:kcl}
Wang, G., Li, W., Ourselin, S., Vercauteren, T.:
\newblock Automatic brain tumor segmentation using convolutional neural
  networks with test-time augmentation.
\newblock BrainLes 2018, Springer LNCS \textbf{11384} (2019)  61--72

\bibitem{brats18:prepro:ChiatseWang}
Wang, C.J., Tsai, Y.M., Lee, C., Lee, Y., Costa, A., Hsu, C., Oermann, E.,
  Wang, W.:
\newblock Brain tumor segmentation with capsule networks versus fully
  convolutional neural networks.
\newblock MICCAI BraTS 2018 Pre-proceedings -
  \url{https://www.cbica.upenn.edu/sbia/Spyridon.Bakas/MICCAI_BraTS/MICCAI_BraTS_2018_proceedings_shortPapers.pdf}
  (2018)  482--491

\bibitem{brats18:rank3:surv:leon}
Weninger, L., Rippel, O., Koppers, S., Merhof, D.:
\newblock Segmentation of brain tumors and patient survival prediction: Methods
  for the brats 2018 challenge.
\newblock BrainLes 2018, Springer LNCS \textbf{11384} (2019)  3--12

\bibitem{brats18:prepro:ShaochengWu}
Wu, S., Li, H., Guan, Y.:
\newblock Multimodal brain tumor segmentation using u-net.
\newblock MICCAI BraTS 2018 Pre-proceedings -
  \url{https://www.cbica.upenn.edu/sbia/Spyridon.Bakas/MICCAI_BraTS/MICCAI_BraTS_2018_proceedings_shortPapers.pdf}
  (2018)  508--515

\bibitem{brats18:prepro:PeiyuanXu}
Xu, P., Hu, Y., Ma, K., Zheng, Y.:
\newblock A two-step cascaded strategy for automatic brain tumor segmentation
  in miccai 2018 brats challenge.
\newblock MICCAI BraTS 2018 Pre-proceedings -
  \url{https://www.cbica.upenn.edu/sbia/Spyridon.Bakas/MICCAI_BraTS/MICCAI_BraTS_2018_proceedings_shortPapers.pdf}
  (2018)  516--524

\bibitem{brats18:prepro:XiaowenXu}
Xu, X., Kong, X., Sun, G., Lin, F., Cui, X., Sun, S., Wu, Q., Liu, J.:
\newblock Brain tumor segmentation and survival prediction based on extended
  u-net model and xgboost.
\newblock MICCAI BraTS 2018 Pre-proceedings -
  \url{https://www.cbica.upenn.edu/sbia/Spyridon.Bakas/MICCAI_BraTS/MICCAI_BraTS_2018_proceedings_shortPapers.pdf}
  (2018)  525--533

\bibitem{brats18:32}
Xu, Y., Gong, M., Fu, H., Tao, D., Zhang, K., Batmanghelich, K.:
\newblock Multi-scale masked 3-d u-net for brain tumor segmentation.
\newblock BrainLes 2018, Springer LNCS \textbf{11384} (2019)  222--233

\bibitem{brats18:33}
Yang, H.Y., Yang, J.:
\newblock Automatic brain tumor segmentation with contour aware residual
  network and adversarial training.
\newblock BrainLes 2018, Springer LNCS \textbf{11384} (2019)  267--278

\bibitem{brats18:34}
Yao, H., Zhou, X., Zhang, X.:
\newblock Automatic segmentation of brain tumor using 3d se-inception networks
  with residual connections.
\newblock BrainLes 2018, Springer LNCS \textbf{11384} (2019)  346--357

\bibitem{brats18:prepro:XiaoyueZhang}
Zhang, X., Jian, W., Cheng, K.:
\newblock 3d dense u-nets for brain tumor segmentation.
\newblock MICCAI BraTS 2018 Pre-proceedings -
  \url{https://www.cbica.upenn.edu/sbia/Spyridon.Bakas/MICCAI_BraTS/MICCAI_BraTS_2018_proceedings_shortPapers.pdf}
  (2018)  562--570

\bibitem{brats18:rank3:seg:chenhong}
Zhou, C., Chen, S., Ding, C., Tao, D.:
\newblock Learning contextual and attentive information for brain tumor
  segmentation.
\newblock BrainLes 2018, Springer LNCS \textbf{11384} (2019)  497--507

\bibitem{recist1}
Tsuchida, Y., Therasse, P.:
\newblock Response evaluation criteria in solid tumors (recist): New
  guidelines.
\newblock Medical and Pediatric Oncology \textbf{37} (2001)  1--3

\bibitem{recist2}
Eisenhauer, E.A., Therasse, P., Bogaerts, J., Schwartz, L.H., Sargent, D.,
  Ford, R., et~al.:
\newblock New response evaluation criteria in solid tumours: revised recist
  guideline (version 1.1).
\newblock European journal of cancer \textbf{45} (2009)  228--247

\bibitem{recist3}
v.~P.~van Meerten, E.L., Gelderblom, H., Bloem, J.L.:
\newblock Recist revised: implications for the radiologist. a review article on
  the modified recist guideline.
\newblock European radiology \textbf{20} (2010)  1456--1467

\bibitem{recist4}
Ellingson, B.M., Wen, P.Y., Cloughesy, T.F.:
\newblock Modified criteria for radiographic response assessment in
  glioblastoma clinical trials.
\newblock Neurotherapeutics : the journal of the American Society for
  Experimental NeuroTherapeutics \textbf{14} (2017)  307--320

\bibitem{rano}
Wen, P.Y., Macdonald, D.R., Reardon, D.A., Cloughesy, T.F., Sorensen, A.G.,
  Galanis, E., et~al.:
\newblock Updated response assessment criteria for high-grade gliomas: Response
  assessment in neuro-oncology working group.
\newblock Journal of Clinical Oncology \textbf{28} (2010)  1963--1972

\bibitem{rgRutmanEjrLink}
Rutman, A.M., Kuo, M.D.:
\newblock Radiogenomics: Creating a link between molecular diagnostics and
  diagnostic imaging.
\newblock European Journal of Radiology \textbf{70} (2009)  232--241

\bibitem{rgEllingsonGbm}
Ellingson, B.M.:
\newblock Radiogenomics and imaging phenotypes in glioblastoma: novel
  observations and correlation with molecular characteristics.
\newblock Curr Neurol Neurosci Rep \textbf{15} (2015)  506

\bibitem{rgGutmanJnrTcga}
Nicolasjilwan, M., Hu, Y., Yan, C., Meerzaman, D., Holder, C.A., Gutman, D.,
  et~al.:
\newblock Addition of mr imaging features and genetic biomarkers strengthens
  glioblastoma survival prediction in tcga patients.
\newblock Journal of Neuroradiology \textbf{42} (2015)  212--221

\bibitem{rgItakuraStmSubtypes}
Itakura, H., Achrol, A.S., Mitchell, L.A., Loya, J.J., Liu, T., Westbroek,
  E.M., et~al.:
\newblock Magnetic resonance image features identify glioblastoma phenotypic
  subtypes with distinct molecular pathway activities.
\newblock Science Translational Medicine \textbf{7} (2015)  303ra128--303ra128

\bibitem{rgBakasCcrPhi}
Bakas, S., Akbari, H., Pisapia, J., Martinez-Lage, M., Rozycki, M., Rathore,
  S., et~al.:
\newblock In vivo detection of egfrviii in glioblastoma via perfusion magnetic
  resonance imaging signature consistent with deep peritumoral infiltration:
  the $\phi$-index.
\newblock Clinical Cancer Research \textbf{23} (2017)  4724--4734

\bibitem{rgJayashreeCcrIdh}
Chang, K., Bai, H.X., Zhou, H., Su, C., Bi, W.L., Agbodza, E., et~al.:
\newblock Residual convolutional neural network for the determination of
  <em>idh</em> status in low- and high-grade gliomas from mr imaging.
\newblock Clinical Cancer Research \textbf{24} (2018)  1073--1081

\bibitem{rgAkbariNeuroOncologyEgfrv3}
Akbari, H., Bakas, S., Pisapia, J.M., Nasrallah, M.P., Rozycki, M.,
  Martinez-Lage, M., et~al.:
\newblock In vivo evaluation of egfrviii mutation in primary glioblastoma
  patients via complex multiparametric mri signature.
\newblock Neuro-Oncology \textbf{20} (2018)  1068--1079

\bibitem{rgBinderCancercellA289}
Binder, Z.A., Thorne, A.H., Bakas, S., Wileyto, E.P., Bilello, M., Akbari, H.,
  et~al.:
\newblock Epidermal growth factor receptor extracellular domain mutations in
  glioblastoma present opportunities for clinical imaging and therapeutic
  development.
\newblock Cancer Cell \textbf{34} (2018)  163--177

\bibitem{havaeiBengio}
Havaei, M., Guizard, N., Chapados, N., Bengio, Y.:
\newblock Hemis: Hetero-modal image segmentation.
\newblock Cham \textbf{2016} (2016)  469--477

\bibitem{elllingsonCoalitionTrials}
Ellingson, B.M., Bendszus, M., Boxerman, J., et~al.:
\newblock Consensus recommendations for a standardized brain tumor imaging
  protocol in clinical trials.
\newblock Neuro-Oncology \textbf{17} (2015)  1188--1198

\end{thebibliography}

    \newpage
    \section{Supplementary Material}
    \label{section:supplement}
\subsection{BraTS 2018 Detailed Evaluation}

\newpage
\subsection{BraTS 2017 Detailed Evaluation}

    \begin{figure}
        \begin{centering}
            \includegraphics[width=1\columnwidth]{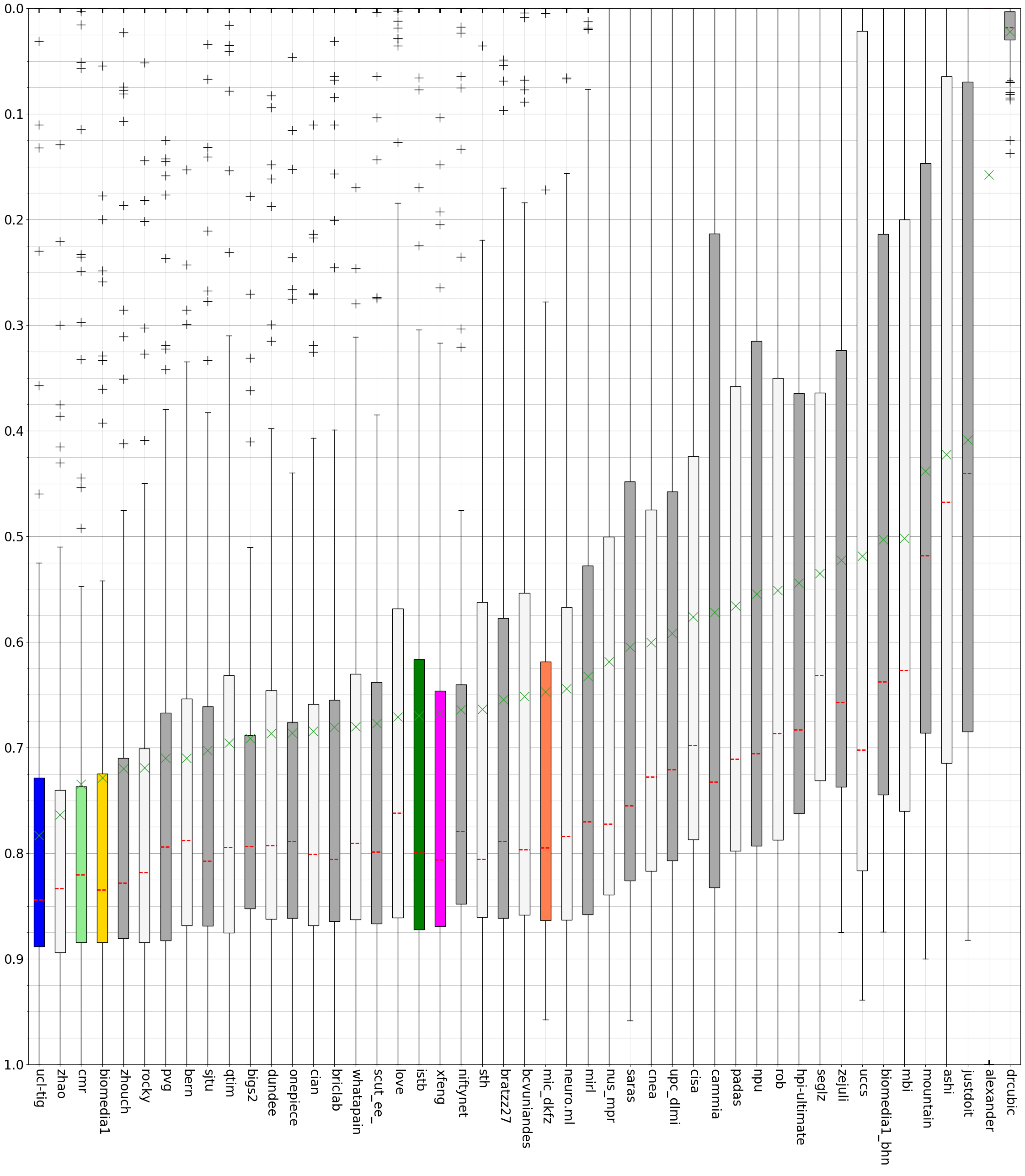}
            \caption{BraTS 2017 summarizing results (Dice) for the segmentation of the active tumor compartment.}
            \label{fig:sup2017:diceET}
        \end{centering}
    \end{figure}

    \begin{figure}
        \begin{centering}
            \includegraphics[width=1\columnwidth]{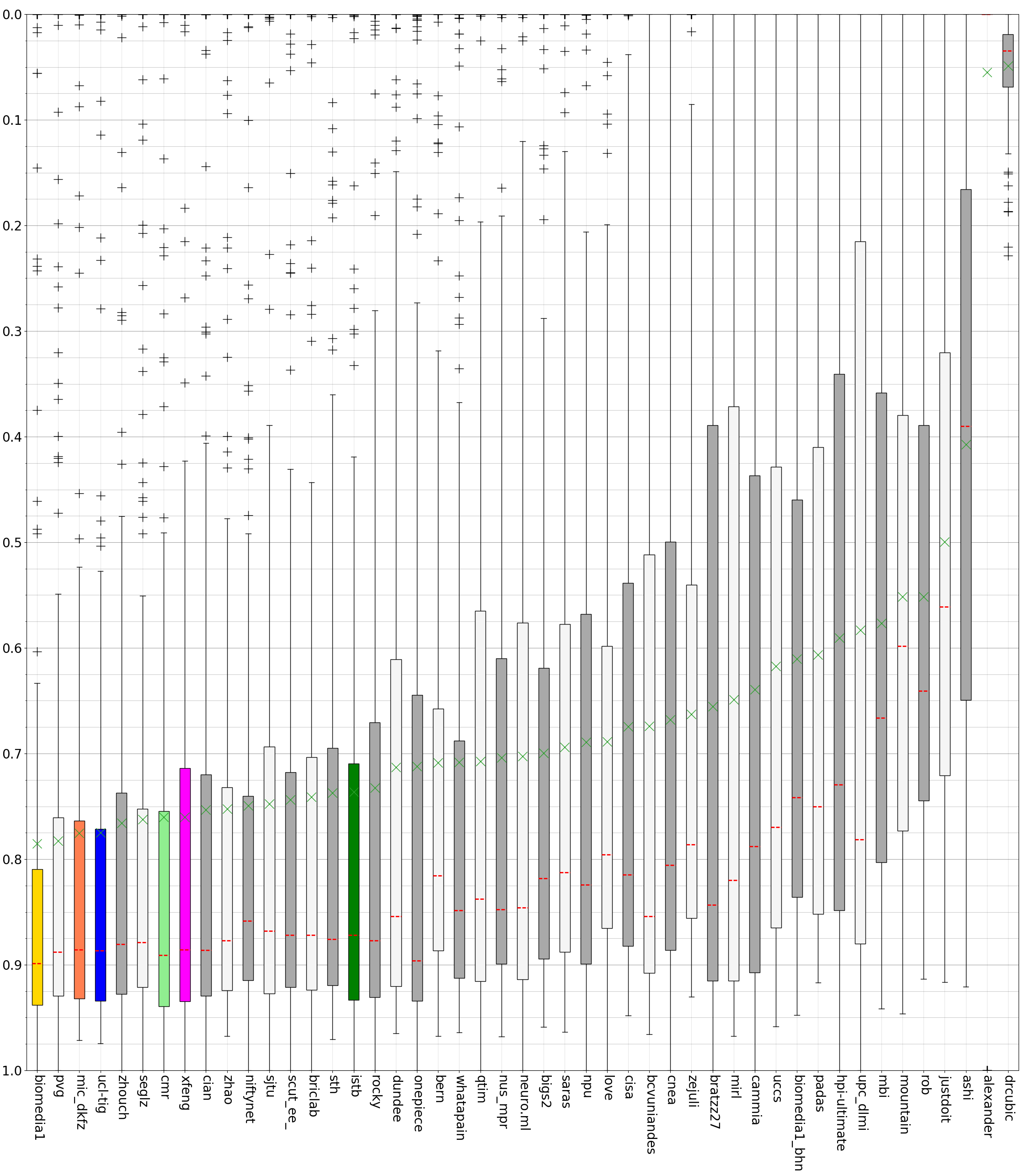}
            \caption{BraTS 2017 summarizing results (Dice) for the segmentation of the tumor core compartment.}
            \label{fig:sup2017:diceTC}
        \end{centering}
    \end{figure}

    \begin{figure}
        \begin{centering}
            \includegraphics[width=1\columnwidth]{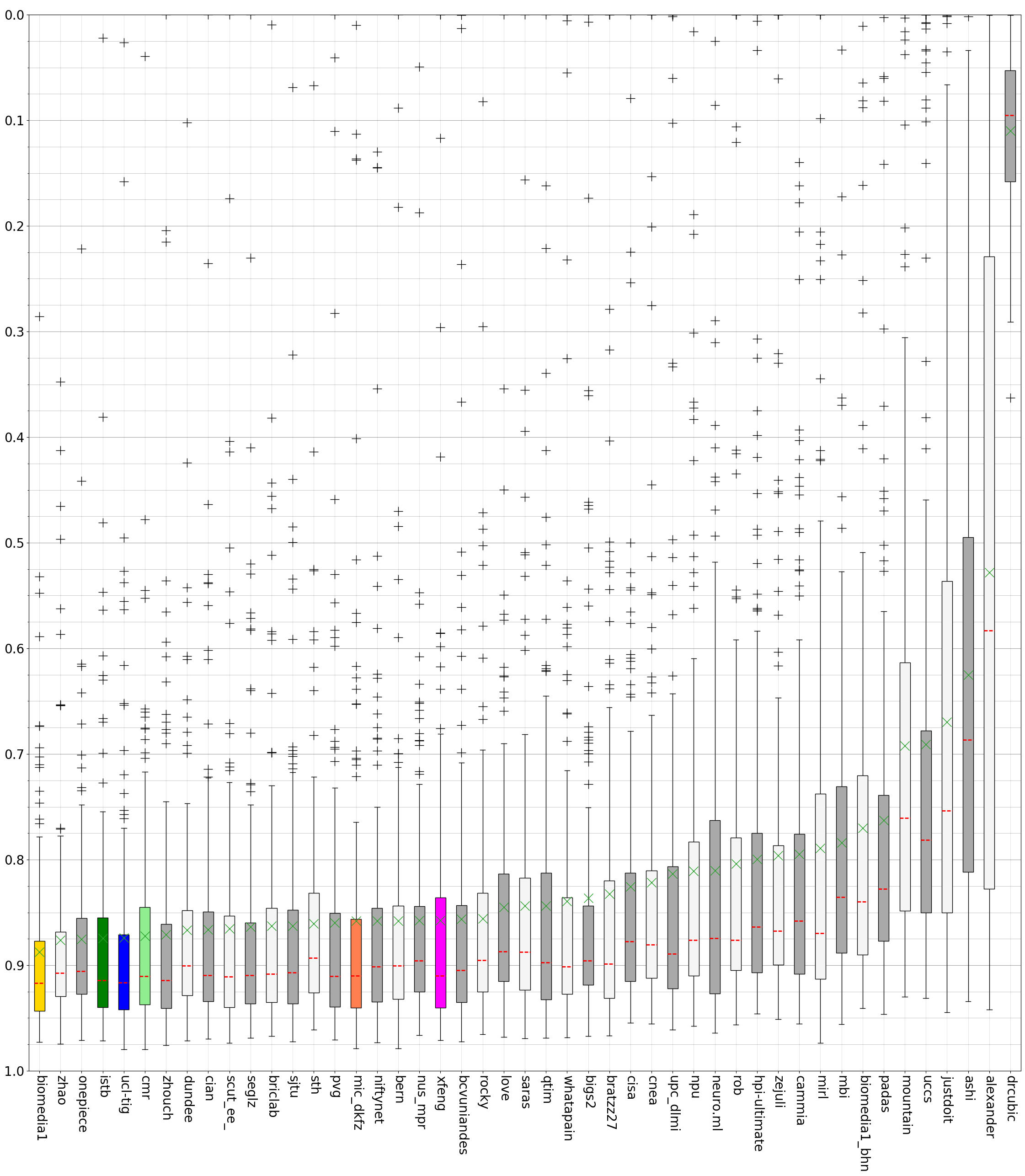}
            \caption{BraTS 2017 summarizing results (Dice) for the segmentation of the whole tumor compartment.}
            \label{fig:sup2017:diceWT}
        \end{centering}
    \end{figure}

    \begin{figure}
        \begin{centering}
            \includegraphics[width=1\columnwidth]{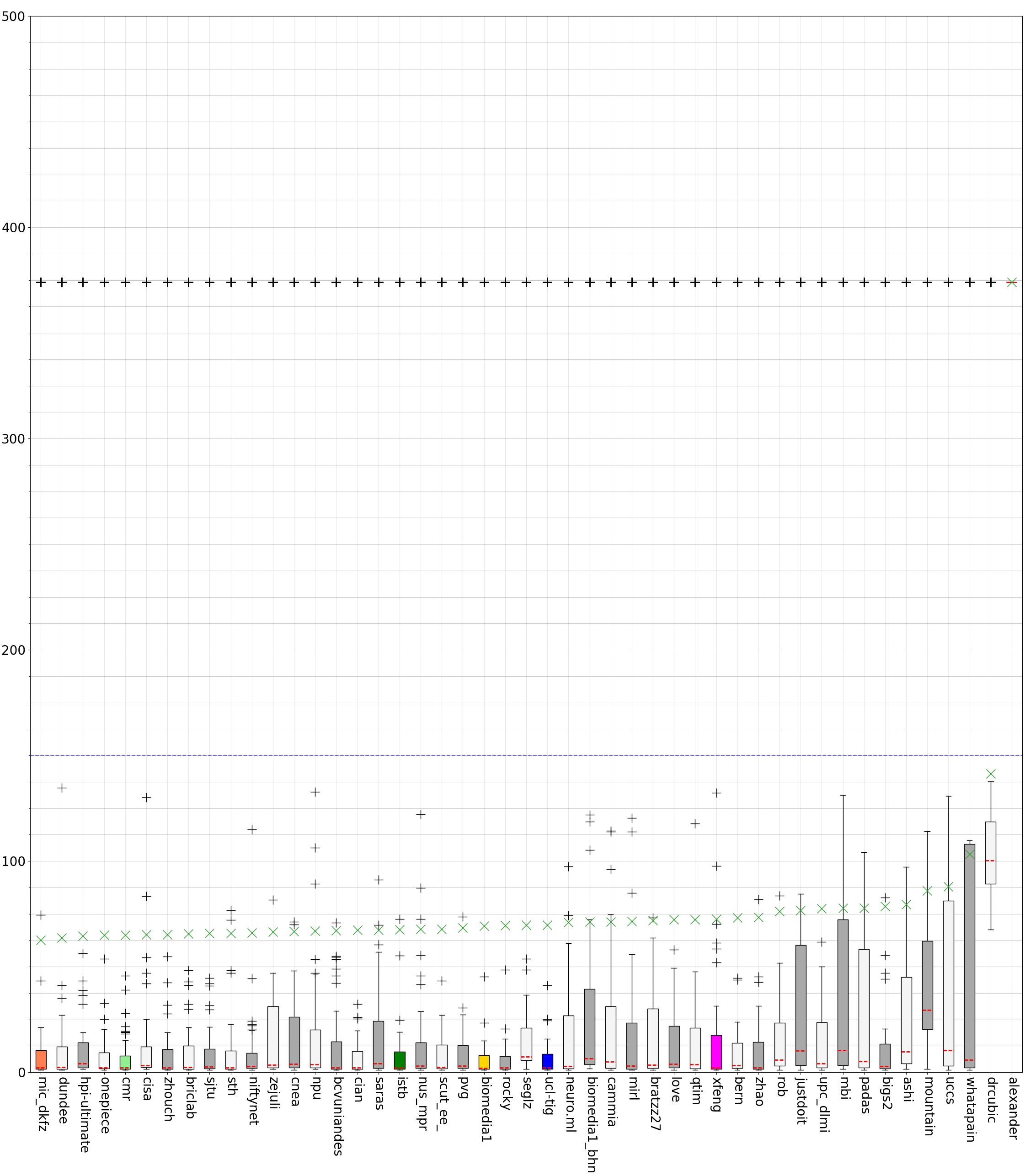}
            \caption{BraTS 2017 summarizing results (Hausdorff) for the segmentation of the active tumor compartment.}
            \label{fig:sup2017:hausdorffET}
        \end{centering}
    \end{figure}

    \begin{figure}
        \begin{centering}
            \includegraphics[width=1\columnwidth]{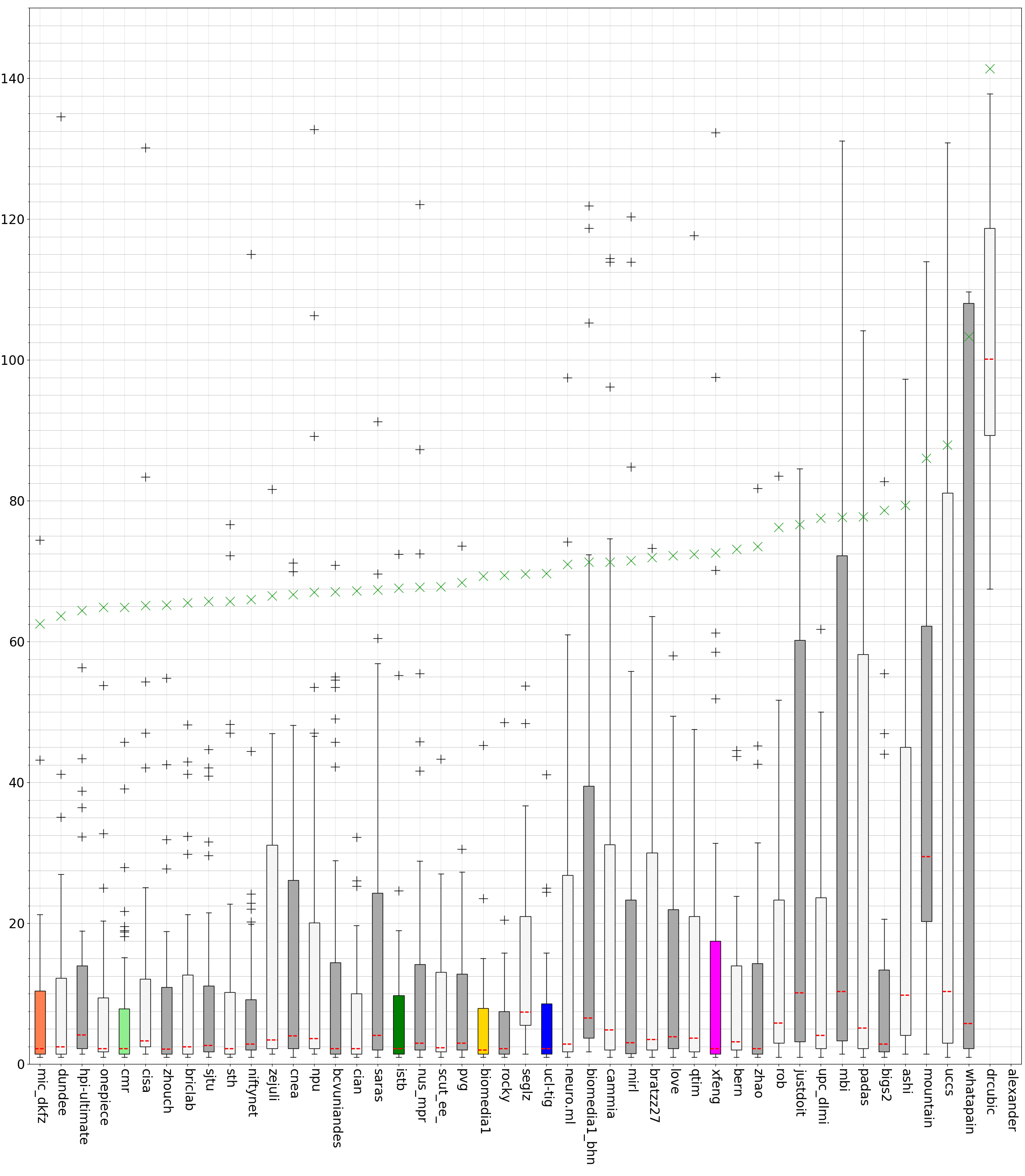}
            \caption{BraTS 2017 summarizing results (Hausdorff) for the segmentation of the active tumor compartment, with cutoff values for visualization purposes.}
            \label{fig:sup2017:hausdorffETCutOff}
        \end{centering}
    \end{figure}

    \begin{figure}
        \begin{centering}
            \includegraphics[width=1\columnwidth]{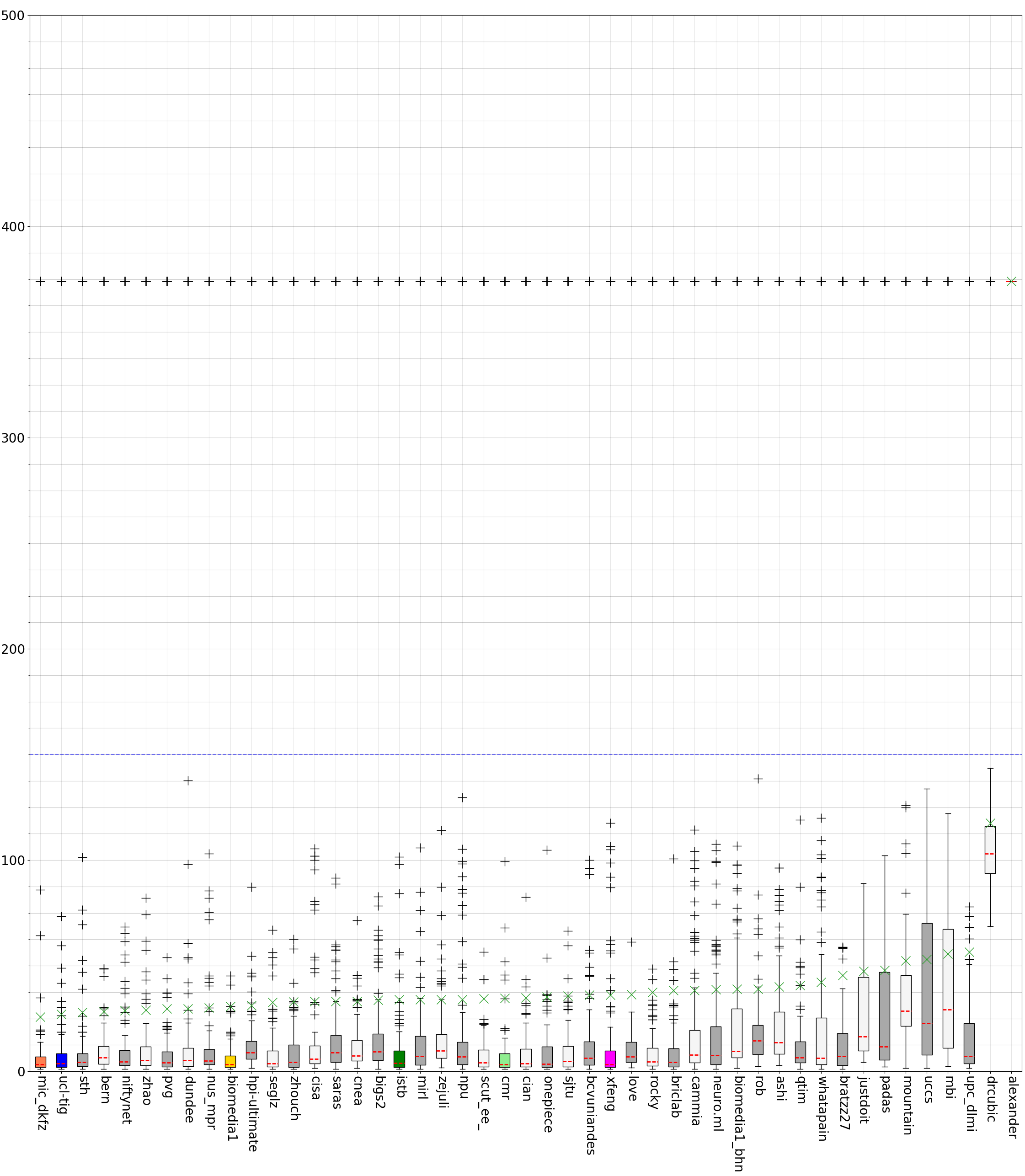}
            \caption{BraTS 2017 summarizing results (Hausdorff) for the segmentation of the tumor core compartment.}
            \label{fig:sup2017:hausdorffTC}
        \end{centering}
    \end{figure}

    \begin{figure}
        \begin{centering}
            \includegraphics[width=1\columnwidth]{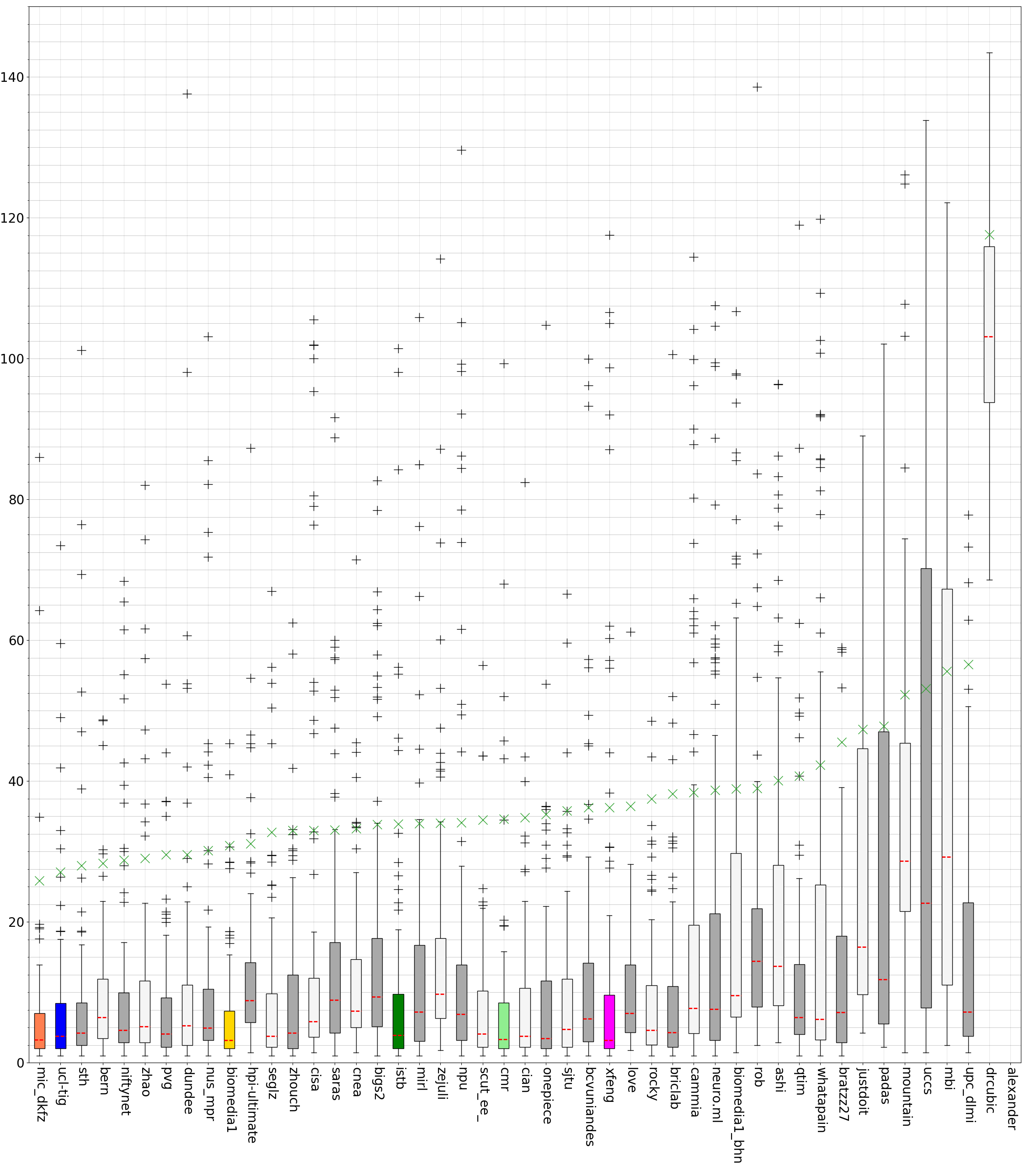}
            \caption{BraTS 2017 summarizing results (Hausdorff) for the segmentation of the tumor core compartment, with cutoff values for visualization purposes.}
            \label{fig:sup2017:hausdorffTCCutOff}
        \end{centering}
    \end{figure}

    \begin{figure}
        \begin{centering}
            \includegraphics[width=1\columnwidth]{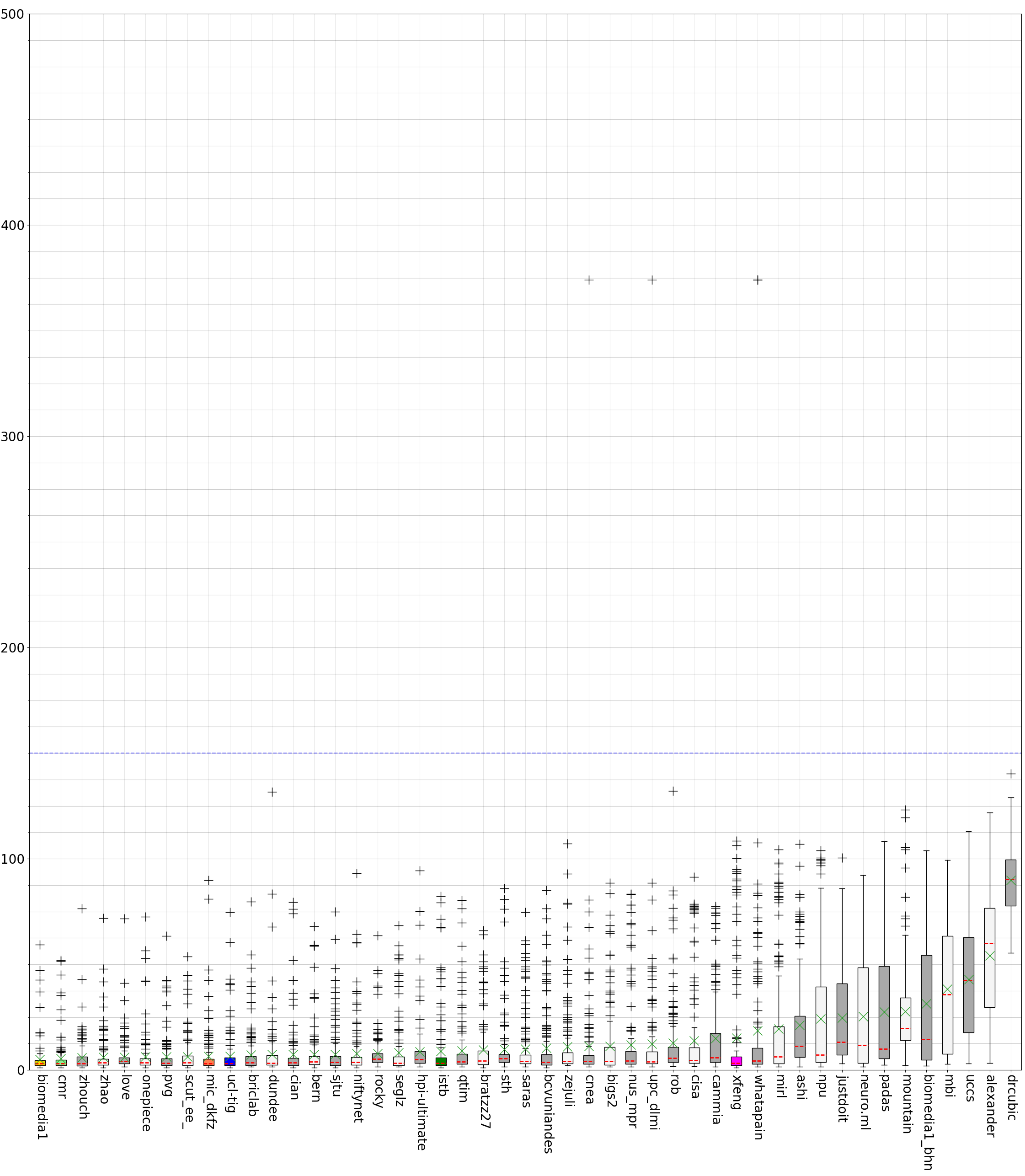}
            \caption{BraTS 2017 summarizing results (Hausdorff) for the segmentation of the whole tumor compartment.}
            \label{fig:sup2017:hausdorffWT}
        \end{centering}
    \end{figure}

    \begin{figure}
        \begin{centering}
            \includegraphics[width=1\columnwidth]{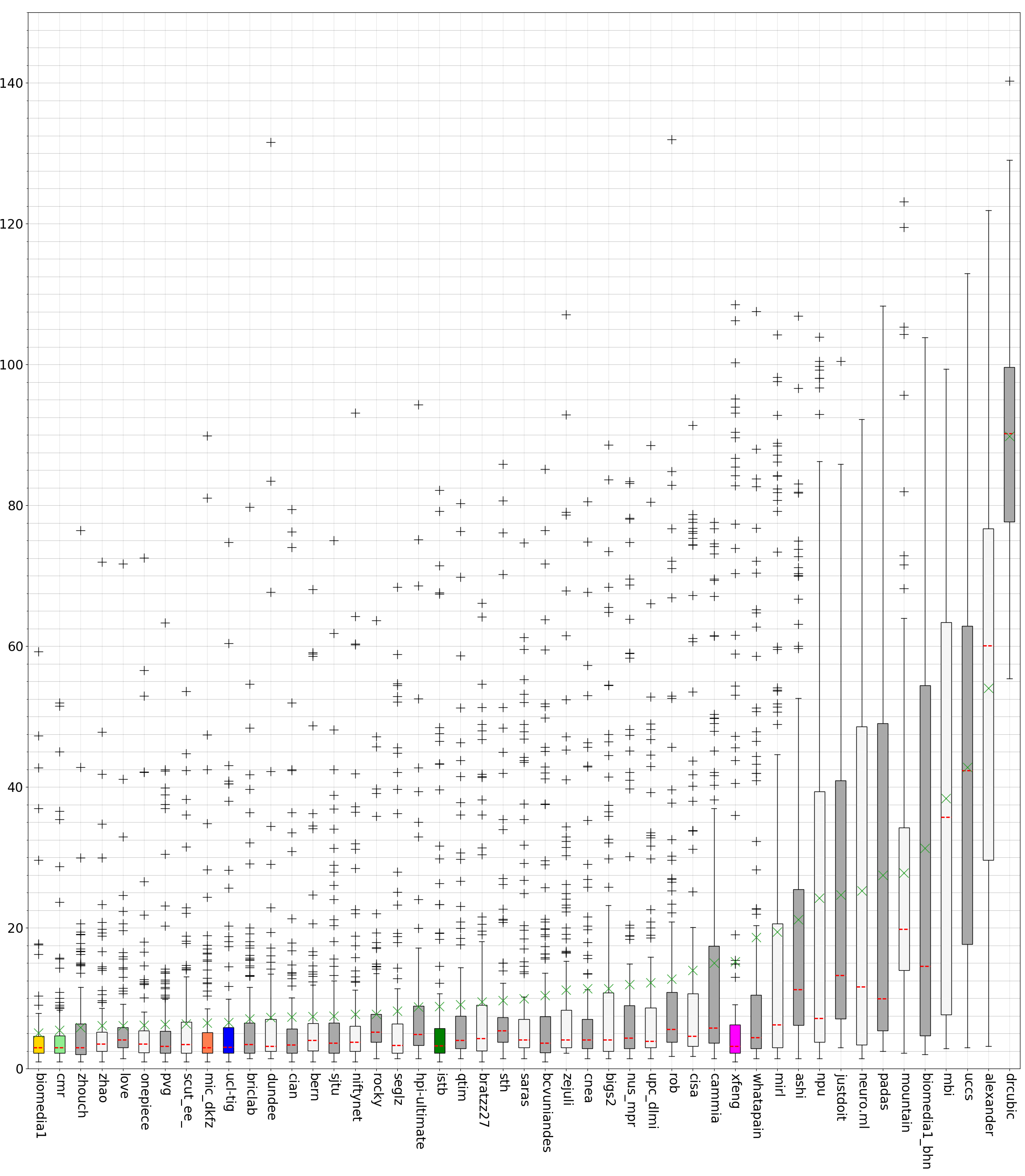}
            \caption{BraTS 2017 summarizing results (Hausdorff) for the segmentation of the whole tumor compartment, with cutoff values for visualization purposes.}
            \label{fig:sup2017:hausdorffWTCutOff}
        \end{centering}
    \end{figure}

\end{document}